\documentclass[lettersize,journal]{IEEEtran}
\usepackage{amsmath,amsfonts}
\usepackage{algorithmic}
\usepackage{algorithm}
\usepackage{array}
\usepackage[caption=false,font=normalsize,labelfont=sf,textfont=sf]{subfig}
\usepackage{textcomp}
\usepackage{stfloats}
\usepackage{url}
\usepackage{verbatim}
\usepackage{graphicx}
\usepackage{cite}
\usepackage[OT1]{fontenc}
\usepackage[dvipsnames]{xcolor}
\usepackage{multicol}
\usepackage{multirow}
\usepackage{siunitx}
\usepackage{amssymb}
\usepackage{amsmath}
\usepackage[nolist]{acronym} 
\usepackage{tikz}
\usetikzlibrary{positioning}
\usepackage{verbatim}
\usepackage{pgfplots}
\usepackage{array}
\usepackage{booktabs}

\usetikzlibrary{patterns}

\usepackage{array, booktabs, multirow}

\usepackage{rotating}

\usepackage{pifont} 
\newcommand{\cmark}{\ding{51}}%
\newcommand{\xmark}{\ding{55}}%

\pgfplotsset{compat=1.17}
\definecolorseries{plotcolor}{rgb}{last}{blue!0}{blue!100}
\resetcolorseries[100]{plotcolor}
\pgfplotsset{
    my area legend/.style={
        draw=none,
        inner ysep=0pt,
        inner xsep=3pt,
        row sep=-.5\pgflinewidth,
        nodes={inner ysep=0pt},
    },
}

\begin{acronym}
\acro{mde}[MDE]{Mean Distance Error}
\acro{caffe}[CaFFe]{``CAlving Fronts and where to Find thEm''}
\acro{SAR}[SAR]{Synthetic Aperture Radar}
\acro{SAM}[SAM]{Segment Anything Model}
\acro{NA}[NA]{``no information available''}
\acro{ViT}[ViT]{Vision Transformer}
\acro{CNN}[CNN]{Convolutional Neural Network}
\acro{AMD-HookNet}[AMD-HookNet]{attention-multihooking-deep-supervision HookNet}
\acro{crf}[CRF]{Conditional Random Field}
\acro{gee}[GEE]{Google Earth Engine}
\acro{sam}[SAM]{Segment Anything Model}
\acro{tsx}[TSX]{TerraSAR-X}
\acro{tdx}[TDX]{TanDEM-X}
\acro{wmo}[WMO]{World Meteorological Organization}
\acro{ecv}[ECV]{Essential Climate Variable}
\acro{gla-st}[GLA-ST]{global-local attention Swin-Transformer block}
\acro{gla}[GLA-STDeepLab]{global-local attention Swin-Transformer-based DeepLabv3+}
\acro{mtl}[MTL]{multi-task learning}
\acro{aspp}[ASPP]{Atrous Spatial Pyramid Pooling}
\acro{dl}[DL]{Deep Learning}
\acro{mcc}[MCC]{Matthew's Correlation Coefficient}
\acro{iou}[IoU]{Intersection over Union}
\end{acronym}

\begin{document}

\title{Comparison Study: Glacier Calving Front Delineation in Synthetic Aperture Radar Images With Deep Learning}

\author{Nora Gourmelon$^{*,1}$, Konrad  Heidler$^{2}$, Erik Loebel$^{3}$, Daniel Cheng$^{4}$, Julian Klink$^{1}$, \\ Anda Dong$^{1}$, Fei Wu$^{1}$, Noah Maul$^{1}$, Moritz Koch$^{5}$, Marcel Dreier$^{1}$, Dakota Pyles$^{5}$, \\ Thorsten Seehaus$^{5}$, Matthias Braun$^{5}$, Andreas Maier$^{1}$, Vincent Christlein$^{1}$ \\
\thanks{$^{1}$Department of Computer Science, Friedrich-Alexander-Universität Erlangen-Nürnberg, Erlangen, Germany}
\thanks{$^{2}$School of Engineering and Design, Technische Universität München, Munich, Germany}
\thanks{$^{3}$Institut für Planetare Geodäsie, Technische Universität Dresden, Dresden, Germany}
\thanks{$^{4}$Jet Propulsion Laboratory, California Institute of Technology, USA}
\thanks{$^{5}$Institut für Geographie, Friedrich-Alexander-Universität Erlangen-Nürnberg, Erlangen, Germany}
\thanks{$^{*}$Corresponding author, \texttt{{nora.gourmelon@fau.de}}}
}

\markboth{IEEE Transactions on Pattern Analysis and Machine Intelligence,~Vol.~XX, No.~X, Month~Year}%
{Gourmelon \MakeLowercase{\textit{et al.}}: Title}

\maketitle

\begin{abstract}
Continuous monitoring of glacier calving fronts is essential for sea level rise projections. This study benchmarks Deep Learning systems for front delineation in Synthetic Aperture Radar imagery. While Deep Learning systems exhibit errors up to 221\,m, human annotators deviate by only 38\,m, underscoring the need for further research.
\end{abstract}


\section{Introduction}
\IEEEPARstart{C}{limate} change is altering our world. 
    One significant change is the recession of glaciers \cite{Otosaka.2023, Hugonnet.2021}.
    For marine-terminating glaciers, major ice mass loss occurs not only due to increasing meltwater runoff but also due to changes in ice dynamics~\cite{Khan.2015, Sheperd.2018}. 
    Glacier calving and changes in the calving front position are two of the main mechanisms controlling these dynamic changes.
    Hence, the calving front position is an essential indicator of glacier dynamics and stability of any marine- or lacustrine-terminating glacier.
    Frontal positions of glacier termini are required to quantify frontal ablation and, thus, quantify their mass change.
    Neglecting the frontal ablation and calving front dynamics can lead to an underestimation of the ice thickness of up to 30~\% \cite{Recinos_2019} and a reduction of glacier contribution to mean sea level rise by $2~\%$ for all temperature change scenarios from 2015 to 2100~\cite{Rounce.2023}.
    Numerical glacier models utilize calving front positions to calibrate and validate their performance or to readjust the model by data assimilation~\cite{Bondzio.2017, Vieli.2011}. 
    Meanwhile, modern satellite systems provide weekly to sub-daily observing capabilities depending on the region whereby the positions of the calving fronts can be localized in the acquired images. 
    \acf{SAR} imagery provides the advantage of continuous monitoring capabilities since the radar signals are illumination and cloud-independent, in contrast to optical imagery. 
    Especially since the launch of the Sentinel-1 mission, the amount of publicly accessible \ac{SAR} imagery has increased substantially.
    The vast amount of data poses a new challenge: manual detection of the front in the individual images becomes infeasible. 
    In addition, there is a large archive of \ac{SAR} imagery from previous missions ranging back to the 1990s.
    Therefore, algorithms for automated analysis of large data quantities are required. 

    Classical approaches are mostly based on preprocessing strategies like denoising and edge enhancement combined with local thresholding and edge detection or contour models~\cite{Liu.2004, Mason.1996, Modava.2017, Tello.2011, Javed.2013, Sohn.1999, Krieger.2017}. These approaches have only been applied to a limited set of scenes and do not guarantee generalizability to new unseen data~\cite{Baumhoer.2019}.
    Since 2019, several studies have applied \ac{dl} techniques to delineate the calving front of marine-terminating glaciers or the coastline of entire ice shelves in satellite imagery.
    The first studies~\cite{Baumhoer.2019, Davari_Islam.2021, Davari_Baller.2021, Gourmelon_2022, Gourmelon.2023, Hartmann.2021, Heidler.2021, Herrmann.2023, Holzmann.2021, Loebel.2022, Mohajerani.2019, Periyasamy.2022, Wu.2023_1, Zhang.2019} are all based on the U-Net architecture \cite{Ronneberger.2015}. 
    Later studies~\cite{Cheng.2021, Heidler.2023, Marochov.2021, Zhang.2021, Zhang.2023} employ networks such as DeepLabv3+~\cite{Chen_2018_ECCV}, Xception~\cite{Chollet_2017_CVPR}, and VGG16~\cite{Simonyan_2015_ICLR}.
    Zhu et\,al.~\cite{Zhu.2023} explore the combination of \acp{CNN} and \acfp{ViT}~\cite{Dosovitskiy.2020}. Currently,
    only one study~\cite{Wu.2023_2} relies on a fully \ac{ViT}-based network~\cite{Dosovitskiy.2020}.
    As different datasets and metrics were used to train and evaluate these algorithms, the results are not comparable.
    
    This study compares these algorithms in terms of their ability to delineate the calving front of marine-terminating glaciers, using \ac{SAR} imagery. 
    In total, we assess the performance of 22 \acf{dl} systems by adapting, re-training, and evaluating every single system with a common benchmark dataset, which was published in prior work~\cite{Gourmelon_2022}.
    We address the questions of whether a particular neural network architecture is better suited for localizing the calving front than others, what influence the label used for training has on performance, and whether more global-scale semantic information in the input is beneficial.
    The in-depth analysis of the assessment offers potential avenues for future research.
    To put the \ac{dl} performance in perspective, we conduct a multi-annotator study.
    Ten anonymous annotators manually labeled each \ac{SAR} image, allowing us to assess the variance between human annotators and check whether automatic front extraction has already reached the quality of manually labeled calving front products.
    
    This study aims to inspire further research on applying deep learning models to calving front delineation in \ac{SAR} imagery by highlighting the unique challenges of the task, providing analytical insights into model performance, and outlining promising future directions with broad implications for climate science and the society.

\section{Methodology}
    For the comparison, the multi-mission \ac{caffe} benchmark dataset~\cite{Gourmelon_2022} is chosen as the basis. It is the largest, manually annotated, and publicly available \ac{SAR} calving front dataset that provides both \ac{SAR} images and corresponding labels.
    Each \ac{SAR} image in the dataset has two manually annotated labels with the same geolocation.
    One label shows the calving front as a binary segmentation mask, where each pixel in the mask belongs to either the front or the background. 
    The other label displays a multi-class segmentation into landscape zones, including ocean and ice mélange (a combination of sea ice and icebergs), rock outcrop, glacier, and a \ac{NA} area that comprises \ac{SAR} shadows and regions outside the radar scene.
    For the zone labels, the calving front is extracted during post-processing.
    The test set of \ac{caffe} includes images of two glaciers, the Columbia and the Mapple Glacier. These glaciers were not seen during training.
    
    In this study, 22 \ac{dl} systems are re-optimized, trained, and evaluated on \ac{caffe}.
    Depending on the type of label used in the original publication, each \ac{dl} system is either trained using the binary front labels, the zone labels, or both labels together.
    The performance is assessed using the \ac{mde} and the number of images with no predicted front, which were introduced together with \ac{caffe}~\cite{Gourmelon_2022}.

    \ac{SAR} imagery is not easy to interpret.
    Ice mélange, for example, exhibits similar characteristics as glacial ice and is therefore easily confused as part of the glacier (see Fig.~\ref{fig:Mapple_bad} and~\ref{fig:Col_bad} as examples).
    Hence, we conducted a multi-annotator study for \ac{caffe}'s test set to visualize and quantify the differences in human annotations.
    The multi-annotator study entailed additional automated post-processing to achieve standard calving front products like the ones commonly provided and used in the community. This allows us to compare the \ac{dl} models to the quality of a standard product. 
    To assess the inter-annotator variance, we calculate the \ac{mde} between each annotator and the combination of the remaining annotator.

    To examine whether \ac{dl} has already reached the quality of standard calving front products based on \ac{SAR} images, the best-performing \ac{dl} system is compared to a ground truth that is built by combining the annotations of all annotators.
    To ensure a fair comparison, the additional automated post-processing applied to the annotations is also applied to the predictions of the \ac{dl} system.

\section{Results}
\subsection{Influences on calving front delineation performance of Deep Learning systems}
\label{subsec:comparison_ai_systems}
    \begin{figure*}[tbp]
        \input{gourmelon-fig1.tex}
        \caption{Overview of \acp{mde} with confidence intervals alongside the number of images with no predicted front for all 22 \ac{dl} systems and the comparisons of the multi-annotator study. The number of images with no predicted front is log-scale \fboxsep=1pt\colorbox{blue!100}{\color{white}intensity encoded} (blue)
        from zero to the number of images in the test set. The multi-annotator study is on the right side of the 
        \fboxsep=1pt\colorbox{violet!100}{\color{white}violet} dashed line. 
        All comparisons in the multi-annotator study were performed with additionally post-processed calving fronts. The asterisk~(*) indicates that the outputs were not compared with \ac{caffe}'s test set but with combined annotations from the multi-annotator study.} 
        \label{fig:ai_vs_ai_vs_human_vs_ai}
    \end{figure*}

\begin{figure*}
    \centering
        \includegraphics[width=1.0\textwidth]{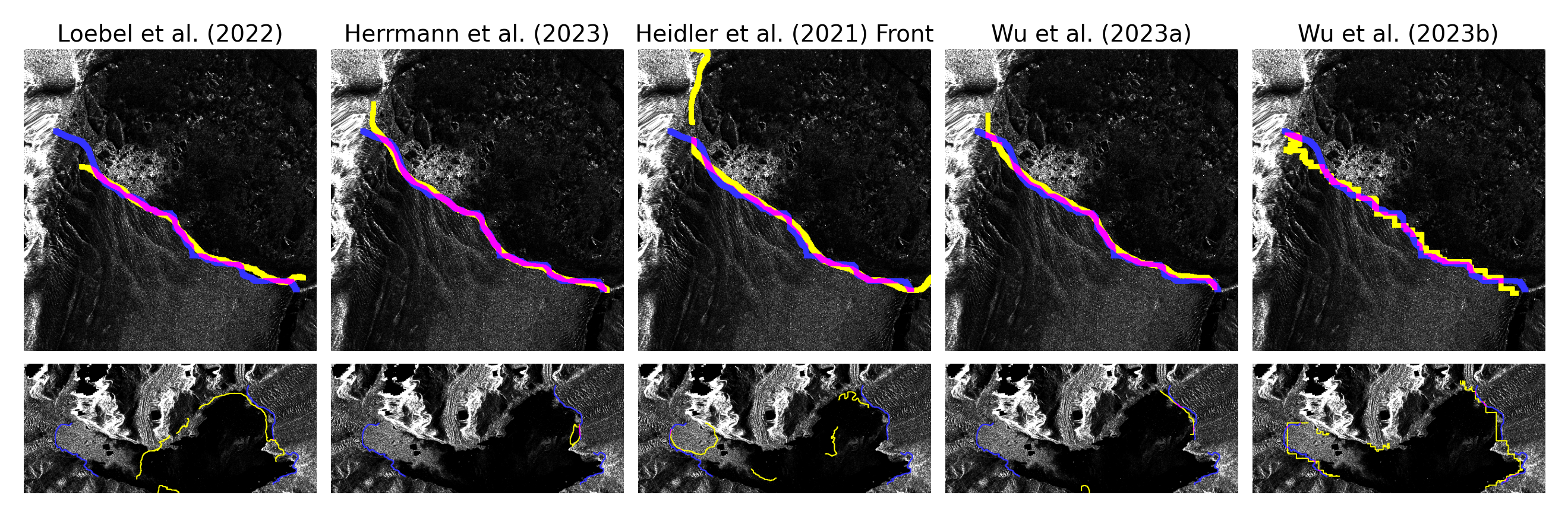}
        \caption{Predicted calving fronts of the five best-performing \ac{dl} systems for an image of the Mapple Glacier taken on 24th October 2008 by the TerraSAR-X satellite and an image of the Columbia Glacier taken on 8th September 2017 by the Sentinel-1 satellite. \fboxsep=1pt\colorbox{yellow!100}{Yellow} depicts the prediction, 
        \fboxsep=1pt\colorbox{blue!100}{\color{white}blue} is used for the ground truth front from CaFFe's test set, and 
        \fboxsep=1pt\colorbox{magenta!100}{\color{white}pink} signifies a perfect match between prediction and ground truth. \ac{SAR} imagery is provided by DLR, ESA, and ASF. Best viewed zoomed in.}
        \label{fig:five_best_models_Mapple_COL}
    \end{figure*}

    \noindent The number of images with no predicted front varies strongly between the 22 systems.
    One system fails to detect a calving front in 100 of the 122 images in the test set, while two systems detect fronts in all 122 images.
    The \acp{mde} of the systems range between \SIrange{338}{4712}{\meter} (Fig.~\ref{fig:ai_vs_ai_vs_human_vs_ai}).
    We provide a visual comparison between the predictions of the five \ac{dl} systems with the lowest \ac{mde} for sample images of the Columbia and Mapple glaciers in fig.~\ref{fig:five_best_models_Mapple_COL}.
    In an attempt to explain the significant differences in performance between the systems, we sort the \ac{dl} systems according to certain characteristics and check whether there is a link with performance.
    The statistical methods used and the numerical results can be found in  the supplementary material.
    The first feature we examine is the basic architecture, i.\,e., the underlying neural network composition upon which the individual model is built. 
    \acp{ViT}~\cite{Dosovitskiy.2020} significantly outperform other architectures such as DeepLabv3+~\cite{Chen_2018_ECCV} or U-Net~\cite{Ronneberger.2015}.
    Further analyses suggest that the inclusion of global-scale semantic information through larger input sizes and strategies for the targeted use of this information, e.\,g., deeper U-Net~\cite{Ronneberger.2015} architectures, appear to be crucial factors for the performance of \ac{dl} systems.
    The integration of other strategies for utilizing global and multi-scale information, such as \ac{aspp}~\cite{Chen.2018}, the HookNet architecture~\cite{vanRijthoven.2021}, or attention mechanisms as in \ac{ViT}~\cite{Dosovitskiy.2020}, is also beneficial.
    Employing additional information with regards to the training labels offers another advantage; both \ac{mtl} approaches~\cite{Cheng.2021, Heidler.2021, Herrmann.2023} and systems trained only on \ac{caffe}'s zone label~\cite{Gourmelon_2022, Gourmelon.2023, Hartmann.2021, Kirillov.2023, Loebel.2022, Marochov.2021, Periyasamy.2022, Wu.2023_1, Wu.2023_2, Zhang.2023, Zhang.2021, Zhu.2023} outperform systems trained only on \ac{caffe}'s binary front~\cite{Davari_Islam.2021, Davari_Baller.2021, Gourmelon_2022, Holzmann.2021, Mohajerani.2019}. 
    
    Nonetheless, several \ac{dl} systems show difficulties in segmenting images of the Columbia Glacier taken by Sentinel-1.
    One of the \ac{dl} systems in our comparison is a foundation model~\cite{Kirillov.2023}.
    For \ac{caffe}, the usage of this foundation model~\cite{Kirillov.2023} in the advertised zero-shot way resulted in a higher \ac{mde} than the \ac{mde} of the model~\cite{Wu.2023_2} used to generate input prompts.

    The \ac{dl} system with the lowest \ac{mde} is the HookFormer~\cite{Wu.2023_2}, a \ac{ViT} that has two connected branches with different resolution levels and was trained on \ac{caffe}'s zone labels.
    One of the branches receives a down-scaled image showing the greater global surroundings, while the other branch takes in the current high-resolution region of interest.
    This mimics the human approach of first mapping the surroundings and then zooming into the area of the calving front once the overarching formation is recognized.
    Although the HookFormer achieves the lowest \ac{mde}, it still encounters issues with some predictions.
    In certain images from the test set, the system incorrectly identifies ice mélange as part of the glacier, erroneously shifting the calving front towards the ocean. 
	This misclassification reduces the system's performance during the winter months.    
	In other images, rocky coastline is misidentified as part of the calving front.
    Moreover, the HookFormer, like the other \ac{dl} systems, exhibits a decreased delineation performance for Sentinel-1 images of the Columbia Glacier.
    Additionally, the outputs of the HookFormer show slight patching artifacts.    
    Since the complete images provided in \ac{caffe} are too large to be fed unchanged into the neural network, the images must be divided into patches, which, in this case, sometimes leads to completely straight edges between the predicted classes.     
    Lastly, HookFormer's delineated calving fronts seem to be jagged and unsmooth, which could be fixed during post-processing.
    
\subsection{Variations between manual annotations}
    \noindent In most cases, the labeled calving fronts of the multi-annotator study do not differ much between the different annotators.
    The averaged \ac{mde} of all annotators for the complete test set is \SI{38}{\meter} with a standard deviation of \SI{15}{\meter}.
    The \acp{mde} for each annotator and for different subsets of the test set are given in Table~\ref{tab:distance_errors_annotators}.
	The labeling of images from the Mapple Glacier was more ambiguous compared to those of the Columbia Glacier. 
    Annotations for Envisat and ALOS PALSAR satellite images of the Mapple Glacier exhibit higher \acp{mde} compared to other sensors.
    On average, the \ac{mde} for Sentinel-1 images is higher than for ERS-1/2 and TanDEM-X.
    
    \begin{table*}[htbp]
        \centering
        \caption{Mean Distance Errors (MDEs) in meters to the joint annotations for the post-processed outputs of the ten annotators and the five trained HookFormers (Run). 
        The HookFormer runs are trained on annotator number ten and compared with the combined annotations of all ten annotators, while the single annotators are compared to the combined annotations of the remaining nine annotators.
        The MDE is calculated for different subsets: complete test set (All), summer (Sum.), winter (Win.), Mapple Glacier (Map.), Columbia Glacier (Col.), Sentinel-1 (S1), Envisat (Envi.), ERS-1/2 (ERS), ALOS PALSAR (PAL.), TerraSAR-X and TanDEM-X (TSX), a resolution of \SI{20}{\meter} (20), a resolution of \SI{17}{\meter} (17), and a resolution of \SI{7}{\meter} (7). The number of images with no predicted front is not given, as it is 0 for all runs and annotators.}
        \label{tab:distance_errors_annotators}
        \begin{tabular}{p{0.05\textwidth} p{0.05\textwidth} p{0.04\textwidth} |p{0.04\textwidth} p{0.04\textwidth} |p{0.04\textwidth} p{0.04\textwidth} |p{0.04\textwidth} p{0.04\textwidth} p{0.04\textwidth} p{0.04\textwidth} p{0.04\textwidth}| p{0.04\textwidth} p{0.04\textwidth} p{0.04\textwidth}} 
            \toprule
            &&& \multicolumn{2}{c}{\textit{Season}} & \multicolumn{2}{c}{\textit{Glacier}} & \multicolumn{5}{c}{\textit{Sensor}} & \multicolumn{3}{c}{\textit{Resolution}}\\
            
            & & \textit{All} & \textit{Sum.} & \textit{Win.} & \textit{Map.} & \textit{Col.} & \textit{S1} & \textit{Envi.} & \textit{ERS} & \textit{PAL.} & \textit{TSX} & \textit{20} & \textit{17} & \textit{7}\\
            \midrule
            \multirow{10}{*}{Anno.} 
            & \# 1 & $81$ & $75$ & $87$ & $238$ & $44$ & $149$ & $816$ & $21$ & $651$ & $30$ & $240$ & $651$ & $30$ \\
            & \# 2 & $30$ & $24$ & $36$ & $32$ & $30$ & $86$ & $36$ & $22$ & $61$ & $19$ & $79$ & $61$ & $19$ \\
            & \# 3 & $45$ & $44$ & $47$ & $56$ & $43$ & $135$ & $60$ & $24$ & $226$ & $23$ & $125$ & $226$ & $23$ \\
            & \# 4 & $38$ & $31$ & $46$ & $58$ & $34$ & $117$ & $239$ & $12$ & $36$ & $18$ & $129$ & $36$ & $18$ \\
            & \# 5 & $35$ & $28$ & $42$ & $34$ & $35$ & $103$ & $78$ & $29$ & $47$ & $20$ & $99$ & $47$ & $20$ \\
            & \# 6 & $23$ & $18$ & $28$ & $28$ & $22$ & $66$ & $78$ & $22$ & $41$ & $13$ & $67$ & $41$ & $13$ \\
            & \# 7 & $33$ & $27$ & $40$ & $27$ & $35$ & $95$ & $57$ & $32$ & $58$ & $21$ & $89$ & $58$ & $21$ \\
            & \# 8 & $30$ & $24$ & $35$ & $29$ & $30$ & $92$ & $51$ & $15$ & $56$ & $17$ & $86$ & $56$ & $17$ \\
            & \# 9 & $28$ & $26$ & $30$ & $37$ & $26$ & $74$ & $65$ & $16$ & $47$ & $18$ & $72$ & $47$ & $18$ \\
            & \# 10 & $41$ & $39$ & $44$ & $28$ & $44$ & $73$ & $34$ & $26$ & $68$ & $35$ & $68$ & $68$ & $35$ \\
            & Mean & $\mathbf{38}$ & $\mathbf{34}$ & $\mathbf{43}$ & $\mathbf{57}$ & $\mathbf{34}$ & $\mathbf{99}$ & $\mathbf{151}$ & $\mathbf{22}$ & $\mathbf{129}$ & $\mathbf{21}$ & $\mathbf{105}$ & $\mathbf{129}$ & $\mathbf{21}$ \\
            \midrule
            \multirow{5}{*}{Run} 
            & \# 1 & $207$ & $161$ & $256$ & $132$ & $223$ & $811$ & $222$ & $145$ & $211$ & $104$ & $724$ & $211$ & $104$ \\
            & \# 2 & $206$ & $171$ & $244$ & $107$ & $228$ & $850$ & $285$ & $199$ & $174$ & $99$ & $764$ & $174$ & $99$ \\
            & \# 3 & $239$ & $172$ & $312$ & $115$ & $266$ & $1067$ & $266$ & $199$ & $188$ & $103$ & $949$ & $188$ & $103$ \\
            & \# 4 & $240$ & $176$ & $307$ & $91$ & $274$ & $1023$ & $185$ & $75$ & $133$ & $114$ & $895$ & $133$ & $114$ \\
            & \# 5 & $212$ & $188$ & $238$ & $103$ & $237$ & $823$ & $190$ & $165$ & $224$ & $105$ & $735$ & $224$ & $105$ \\
            & Mean & $\mathbf{221}$ & $\mathbf{174}$ & $\mathbf{271}$ & $\mathbf{110}$ & $\mathbf{245}$ & $\mathbf{915}$ & $\mathbf{230}$ & $\mathbf{157}$ & $\mathbf{186}$ & $\mathbf{105}$ & $\mathbf{813}$ & $\mathbf{186}$ & $\mathbf{105}$ \\
            \bottomrule
        \end{tabular}
    \end{table*}

    \begin{figure*}\captionsetup[subfigure]{font=scriptsize}
        \scalebox{.5}{
        \subfloat[]{
            \includegraphics[width=0.29\textwidth]{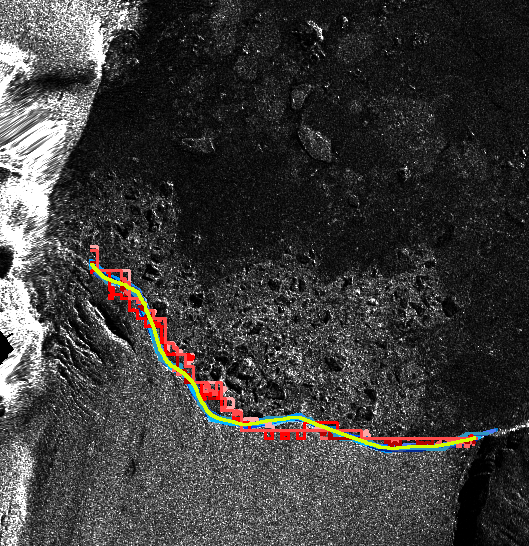}
            \label{fig:Mapple_good}
        }
        \subfloat[]{
            \includegraphics[width=0.68\textwidth]{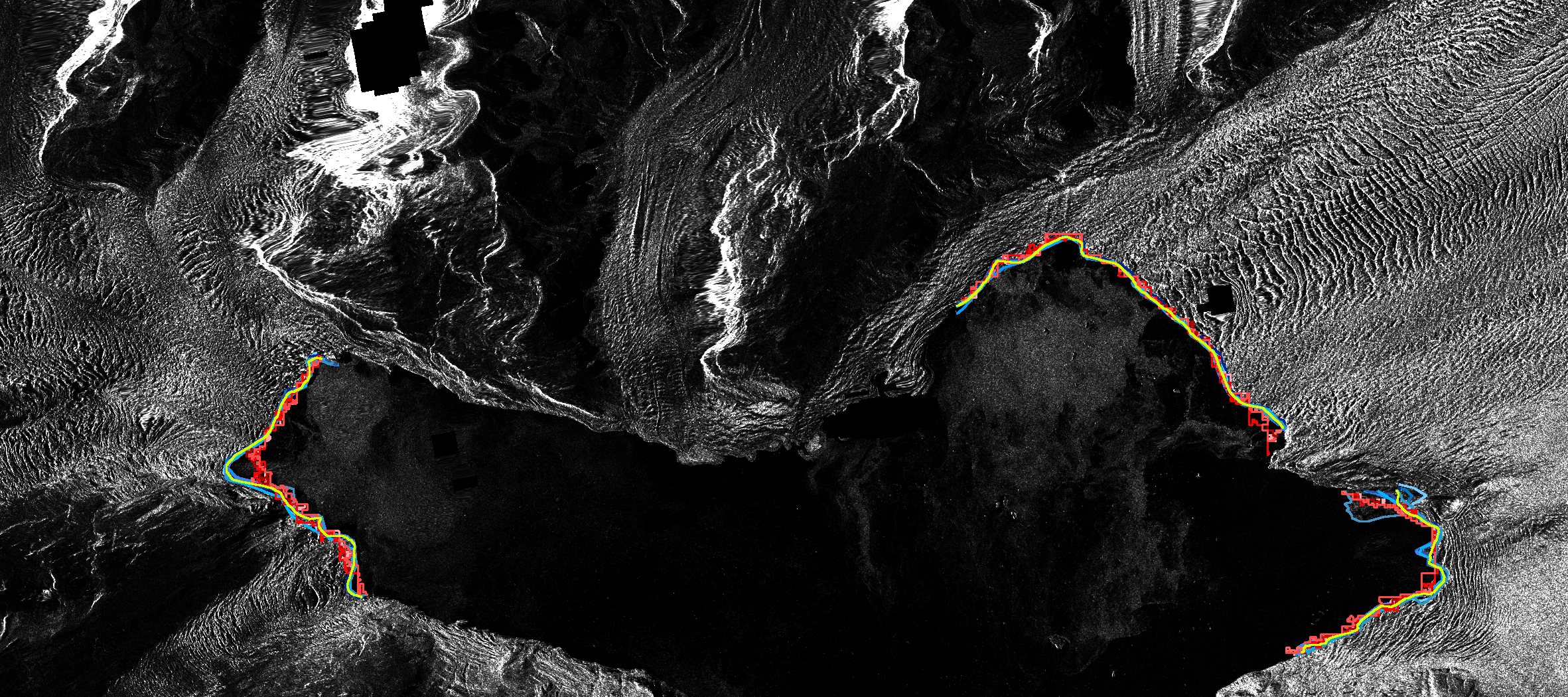}
            \label{fig:Col_good}
        }
        \hfill
        
        \subfloat[]{
            \includegraphics[width=0.29\textwidth]{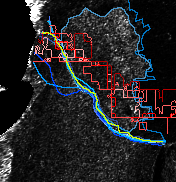}
            \label{fig:Mapple_bad}
        }
        \subfloat[]{
            \includegraphics[width=0.68\textwidth]{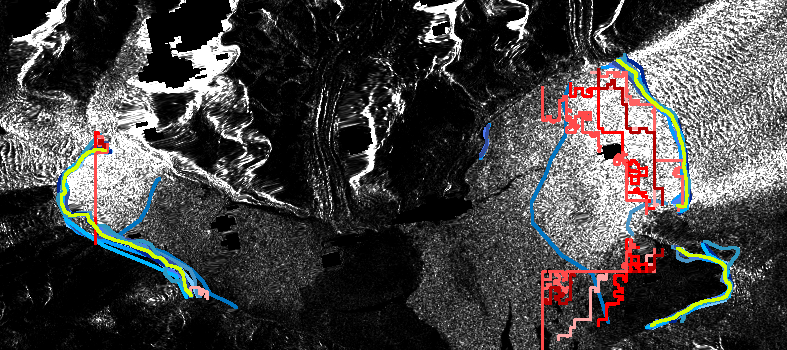}
            \label{fig:Col_bad}
        }  
        }
        \caption{Visualizations for all ten annotations by humans (\fboxsep=1pt\colorbox{blue!100}{\color{white}shades of blue}), the five post-processed HookFormer runs 
        (\fboxsep=1pt\colorbox{red!100}{\color{white}shades of red}), and the aggregation of human annotations  (\fboxsep=1pt\colorbox{yellow!100}{yellow}). (a) shows the Mapple Glacier on 2nd November 2009, acquired by the TSX satellite; (b) shows the Columbia Glacier on 15th March 2016, acquired by the TDX satellite; (c) shows the Mapple Glacier on 9th June 2007, acquired by the Envisat satellite; and (d) shows the Columbia Glacier on 6th January 2018, acquired by the S1 satellite. \ac{SAR} imagery is provided by DLR, ESA, and ASF. Best viewed zoomed in.}
        \label{fig:ai_vs_human}
    \end{figure*}

\subsection{Significant difference between humans and DL}
    \noindent Furthermore, we want to compare the annotators with the best-performing \ac{dl} system - the HookFormer.
    The \ac{mde} of the HookFormer's post-processed automatic calving front predictions is high, with an average of \SI{221}{\meter} and a standard deviation of \SI{15}{\meter}.
    This result is significantly (refer to the supplementary material for the employed statistical tests and their numerical results) higher than the comparatively low average \ac{mde} of the manual annotations with an \ac{mde} of \SI{38}{\meter} and a standard deviation of \SI{15}{\meter}.
    A visual comparison between the \ac{dl} system's and the annotators' performance is shown in Fig.~\ref{fig:ai_vs_ai_vs_human_vs_ai}.
    The \acp{mde} for each run, subdivided into different subsets of the test set, are given in Table~\ref{tab:distance_errors_annotators}.

    The \ac{mde} for the predicted fronts of the \ac{dl} system is higher for winter images than for summer images and higher for images of the Columbia Glacier than for the Mapple Glacier. 
    Of the various sensors, Sentinel-1 has by far the highest \ac{mde} for the outputs of the \ac{dl} system.
    For human annotators, a drop in performance for low-resolution images containing ice mélange is observable. DL systems also show a low MDE for images with a high resolution of 7\,m.
    Two examples where the predictions of the annotators and the runs of the \ac{dl} system closely agree are shown in Fig.~\ref{fig:Mapple_good} and~\ref{fig:Col_good}.

\section{Discussion and outlook}
\label{sec:discussion_outlook}
Many systems were designed for different tasks than the \ac{caffe} dataset, which may explain the low performance. 
The \ac{caffe} dataset provides a challenging basis for calving front delineation because of the use of \ac{SAR} data as opposed to optical imagery, the extraction of laterally bounded glacier calving fronts rather than ice shelf edges, and the construction of the test set containing only glaciers not seen during training.
In general, for automatic monitoring of calving fronts, instead of comparing the performance of different \ac{dl} systems, a larger training dataset with more different glaciers than the benchmark dataset would most likely improve generalizability.

\subsection{SAR imagery}
    \ac{SAR} imagery presents several challenges for calving front detection~\cite{Baumhoer.2019}. Speckle noise, an inherent characteristic of \ac{SAR} data, reduces contrast and obscures key features. Traditional denoising approaches, such as bilateral filtering and Contrast Limited Adaptive Histogram Equalization~\cite{Periyasamy.2022}, can mitigate noise but introduce additional hyperparameters that must be re-tuned for different settings. A more adaptive alternative is integrating denoising within deep learning models, allowing filters to be learned through backpropagation. While known operator learning has been applied in biomedical imaging~\cite{Wagner.2022}, its potential for calving front delineation remains unexplored.
    Beyond noise, contrast variability further complicates front extraction. Certain image features exhibit inherently higher contrast than the calving front, and rough sea surfaces can reduce the contrast between ice and water~\cite{Baumhoer.2019}. Additionally, seasonal variations in ice, snow, and sea ice properties alter backscatter responses throughout the year~\cite{Baumhoer.2019}. Moreover, different ice types, such as glacial and sea ice, may produce similar backscatter values, increasing classification ambiguity~\cite{Baumhoer.2018}.
    Given these limitations, time series analysis of \ac{SAR} imagery emerges as a promising direction. Temporal patterns in backscatter variation might help distinguish ice types and reduce reliance on single-image contrast, potentially improving the robustness of calving front delineation.

\subsection{Sensors}
    A similarity between humans and \ac{dl} systems is the high \acp{mde} for Sentinel-1 images, suggesting that \mbox{Sentinel-1} images are more challenging for humans and \ac{dl} systems to interpret than other sensors. The low performance of \ac{dl} systems on images taken by Sentinel-1 suggests that the training data for \ac{dl} systems designed to work with Sentinel-1 imagery should include more Sentinel-1 samples than the benchmark dataset~\cite{Gourmelon_2022} used for this comparison. Sentinel-1 images are under-represented in \ac{caffe}'s training set (15~Sentinel-1, 52~ERS-1/2, 72~Envisat, 54~RADARSAT-1, 40~ALOS PALSAR, 326~TSX/TDX images) and, additionally, smaller than the average image size in the training set (S1: 998\,$\times$\,651 vs.\ complete training set 2163\,$\times$\,2174). Moreover, the low resolution of \SI{20}{\meter}, combined with ice mélange in front of the calving front, leads to false predictions for Columbia Glacier (see, e.\,g., Fig.~\ref{fig:Col_bad}) and increases the total \ac{mde} for Sentinel-1 images. In general, higher-resolution images seem to be easier for humans and the best performing \ac{dl} system to understand. Therefore, it would be preferable to use a high resolution sensor when delineating calving fronts. If lower resolution sensors shall be used, super-resolution techniques could be explored as a potential enhancement.

\subsection{Calving front shape}
    Our analysis indicates that multi-class labels yield better results than binary front labels, likely due to the strong class imbalance in the latter. However, we argue that the structure of the calving front -- a single-pixel-wide line -- also contributes to this effect. In segmentation tasks, the primary focus is typically on objects rather than their boundaries, as in the case of calving front delineation. Incorporating \ac{mtl} approaches~\cite{Herrmann.2023,Heidler.2021}, or explicitly integrating edge detection into the network as proposed by Heidler et\,al.~\cite{Heidler.2021} can help the model better capture the shape of the front. Additionally, specialized loss functions designed to improve boundary accuracy~\cite{Karimi.2020, Sun.2022} present an interesting direction for future research.

\subsection{Ice mélange}
    During winter, ice mélange is more prevalent, which might cause the drop in performance from summer to winter in \ac{dl} systems.
    The ambiguity in the labeling of Mapple Glacier during the multi-annotator study is likely also due to the presence of ice mélange in front of the calving front in several Mapple Glacier images, which complicated the mapping of the glacier front.
    For example, in Fig.~\ref{fig:Mapple_bad}, the region between the ocean and the glacier was recognized as glacial ice by two annotators, while the other annotators assigned it to the ocean as ice mélange.
    This shows that ice mélange in \ac{SAR} imagery is not only a challenge for \ac{dl} systems but also for humans, thereby constraining the learning possibilities of \ac{dl} systems. Incorporating image acquisition time and geographic location as additional inputs to \ac{dl} systems could help detect the presence of sea ice and ice mélange. More promising, however, would be the application of augmentations that mimic ice mélange. This could be done by simple cut and paste if ice mélange and sea ice were labeled as zones separate from the ocean.

\subsection{Misidentification of rocky coastline}
    Misidentification of rocky coastline as part of the calving front might be an issue of the test set, as the rock class has not been picked manually, so the rock may actually be covered by ice or snow.
    If the system is used to extract calving fronts for new glaciers and not for comparison with other \ac{dl} systems on the benchmark, this situation can easily be avoided by using a static mask that excludes the rocky coastlines of laterally bounding mountains~\cite{Cheng.2021}.
    
\subsection{Patching artifacts}
    The higher \ac{mde} for images of Columbia as opposed to Mapple Glacier could be due to stronger patching artifacts in the Columbia predictions, as the Columbia Glacier images tend to be larger than the Mapple Glacier images and, therefore, need to be split into more patches.
    Patching artifacts could most likely be mitigated if the region-of-interest patches were extracted with an overlap and the outputs at the overlap were averaged with a Gaussian or uncertainty-based weighting.

\subsection{Global- and multi-scale information}
    \ac{dl} systems that incorporate broader spatial context tend to achieve better delineation performance. However, leveraging this information effectively requires more than simply increasing input patch size -- it necessitates dedicated architectural adaptations. Approaches such as deeper U-Net architectures~\cite{Loebel.2022,Herrmann.2023,Heidler.2021}, or separate context branches that integrate global and multi-scale information into local pathways, such as AMD-HookNet~\cite{Wu.2023_1} and the HookFormer~\cite{Wu.2023_2}, have demonstrated success. Additionally, four of the five top-performing models~\cite{Herrmann.2023, Wu.2023_1, Wu.2023_2, Heidler.2021} employ deep supervision, and three utilize attention mechanisms~\cite{Wu.2023_1, Wu.2023_2, Heidler.2021}, further emphasizing the importance of structured multi-scale processing.

\subsection{Neural network architecture and training}
    Based on our results, transformers -- like the HookFormer~\cite{Wu.2023_2} -- are the most effective neural network architecture for calving front delineation, likely due to their ability to capture long-range dependencies and multi-scale features~\cite{Heidari_2023_WACV}. As leveraging global and multi-scale semantic information further improved segmentation performance, we suggest employing deep transformer architectures with large input sizes.

\subsection{Post-processing}
    The post-processing of the results of a network could also be worth further investigation, as demonstrated by our work using conditional random fields~\cite{Gourmelon.2023} which improved the MDE of the baseline system~\cite{Gourmelon_2022}. Other promising initial approaches include smoothing the front~\cite{Cheng.2021}, analyzing the connected components of the ocean~\cite{Gourmelon_2022}, and removing mispredicted fronts based on their confidence values, their length, their shape, or their fit to the time series~\cite{Gourmelon_2022, Baumhoer.2019, Zhang.2021}.

\subsection{Additional Data Sources}
    Previous studies~\cite{Heidler.2021, Loebel.2022} suggest that the addition of digital elevation models as a second input can lead to overfitting.  
    An unexplored research direction is multimodal learning using \ac{SAR} and optical images. A straightforward approach would be to input temporally matched \ac{SAR} and optical images, but this would contradict the goal of monitoring calving fronts under conditions where optical sensors fail -- such as during the polar night or in poor weather. An alternative could be multimodal pretraining, where a model first learns to translate \ac{SAR} images into an optical-like representation. This approach could enable models to extract informative features from optical data while remaining applicable in \ac{SAR}-only scenarios.
    
\subsection{Foundation models}
    For foundation models, further research is required to determine whether fine-tuned versions could outperform specialized models, as they have in other domains~\cite{Ma.2024, Xiao.2024}.
    In addition, future foundation models developed for the segmentation of radar images may be more suitable, as the current versions are generally trained on optical images.

\section{Limitations}
When one of the evaluated \ac{dl} systems shall be used to generate a calving front dataset for analysis, we strongly recommend the use of automated or manual checks to ascertain the plausibility of the delineated calving fronts.
Still, our evaluation is restricted to the scenarios presented in \ac{caffe}'s test set: laterally bounded glaciers not seen during training and captured by \ac{SAR} sensors.
Scenarios like ice shelves, optical images, and glaciers already seen during training are not covered by the \ac{caffe} dataset and have, therefore, not been tested in this study.

\section{Conclusion}
\noindent Our research shows that \ac{dl} calving front delineation systems have not yet reached the quality of standard calving front products.
The best-performing \ac{dl} system produces calving front predictions that are, on average, \SI{183}{\meter} away from the average human-labeled calving front after post-processing.
A direct consequence of this error can be seen in mass balance calculations at the calving gate: the mass error is determined by multiplying the \ac{mde}, the length of the calving front, and the glacier height at the terminus.
As an example, for the Mapple Glacier, the difference of \SI{110}{\metre} to the average human-labeled calving front would lead to an error of \SI{0.38}{\kilo\metre\squared} in glacier area if we multiply the difference by the average length of Mapple's front in \ac{caffe}'s test set. Assuming an average ice thickness of \SI{100}{\meter} at the terminus, this corresponds to an error of \SI{0.038}{\kilo\metre\cubed} in ice volume. Converting this to mass using an ice density of \SI{900}{\kilo\gram\per\metre\cubed}, the error amounts to \SI{34.2}{\mega\tonne}. Such inaccuracies directly impact sea level rise projections~\cite{Rounce.2023} in IPCC reports~\cite{IPCC.2023}. As a result, they influence policy decisions on climate adaptation and mitigation. Reducing these errors in automated delineation systems is therefore essential for improving large-scale glacier monitoring and ensuring that climate policies are based on the most precise data available.

However, an assessment of frontal ablation rates for the large number of tidewater glaciers at high temporal resolution and on regional scales is still missing~\cite{Kochtitzky.2022, Kochtitzky.2023}.
Consequently, we are faced with the need to improve \ac{dl} systems, as manual mapping is not feasible.
From our analyses of the influences on the performance of \ac{dl} systems, several avenues for future research are derived to improve the calving front delineation performance of \ac{dl} systems: 
We suggest that future research should further explore the possibilities of \acp{ViT} and foundation models and focus on the efficient provision and integration of global information.
Until the quality of standard calving front products is achieved, we strongly recommend the use of automated or manual checks to ascertain the plausibility of the delineated calving fronts.

\section*{Acknowledgments}
\noindent The authors thank all annotators for their contribution to this research.
This research was funded by the DFG project CH 2080/5-1 and the Bayerisches Staatsministerium für Wissenschaft und Kunst within the ENB M5613.5.2020.
The authors thank the JPL, California Institute of Technology, for support of their work under a contract with NASA and the NHR@FAU for HPC resources (partially financed by DFG – 440719683) within the projects b110dc and b194dc.
The author team acknowledges the provision of satellite data under various AOs from respective space agencies (DLR, ESA, JAXA, CSA).

\section*{Code and data availability}
\noindent Figures~\ref{fig:five_best_models_Mapple_COL} and~\ref{fig:ai_vs_human} show a subset of \ac{caffe}'s images. 
The full set of visualizations and GIFs showing \ac{SAR} image time series for all glaciers in \ac{caffe} are available at \url{https://zenodo.org/records/15173191}.\\

\section*{Author contribution}
\noindent \textbf{Nora Gourmelon}: Conceptualization, Methodology, Software, Experiments, Statistical Analysis, Project administration, Writing - Original draft preparation. \textbf{Konrad Heidler, Erik Loebel, Julian Klink, Fei Wu}: Software, Experiments, Writing - review \& editing. \textbf{Daniel Cheng}: Software, Writing - review \& editing. \textbf{Noah Maul, Moritz Koch, Marcel Dreier, Dakota Pyles}: Writing - review \& editing. \textbf{Thorsten Seehaus, Matthias Braun, Andreas Maier}: Supervision, Writing - review \& editing. \textbf{Vincent Christlein}: Supervision, Validation, Writing - review \& editing.

\section{Biography Section}
\label{sec:bios}

\vspace{-33pt}
\begin{IEEEbiographynophoto}{Nora Gourmelon} received her B.Sc. and M.Sc. in computer science from Friedrich-Alexander-Universität Erlangen-Nürnberg (FAU). She is pursuing a Ph.D. at the Pattern Recognition Laboratory (PRL), focusing on AI for sustainability and natural sciences. In 2023, she was honored as AI Newcomer by the German Association of Computer Science.
\end{IEEEbiographynophoto}

\vspace{-33pt}
\begin{IEEEbiographynophoto}{Konrad Heidler}
received his Dr.-Ing. from Technical University of Munich in 2024. He is a postdoctoral researcher leading the Visual Learning and Reasoning group, focusing on deep learning in polar regions.
\end{IEEEbiographynophoto}

\vspace{-33pt}
\begin{IEEEbiographynophoto}{Erik Loebel} is pursuing a Ph.D. at Technische Universität Dresden, focusing on remote sensing and machine learning. His work develops deep learning methods to monitor glacier changes in Greenland.
\end{IEEEbiographynophoto}

\vspace{-33pt}
\begin{IEEEbiographynophoto}{Daniel Cheng}
received his Ph.D. in computer science from University of California, focusing on machine learning for glacial feature extraction. He is now a postdoc at Jet Propulsion Laboratory, working on Antarctic Ice Sheet state estimation using ISSM.
\end{IEEEbiographynophoto}

\vspace{-33pt}
\begin{IEEEbiographynophoto}{Julian Klink}
earned his M.Sc. in computer science at FAU's PRL in 2025.
\end{IEEEbiographynophoto}

\vspace{-33pt}
\begin{IEEEbiographynophoto}{Anda Dong}
is pursuing an M.Sc. in computer science at FAU and participated in Tohoku University's Cooperative Laboratory Study program in 2023/24.  
\end{IEEEbiographynophoto}

\vspace{-33pt}
\begin{IEEEbiographynophoto}{Fei Wu}
received his Ph.D. in signal processing from the University of Chinese Academy of Sciences in 2023. He is a postdoctoral researcher at FAU's PRL, focusing on computer vision and machine learning.
\end{IEEEbiographynophoto}

\vspace{-33pt}
\begin{IEEEbiographynophoto}{Noah Maul}
received his B.Sc. and M.Sc. in computer science from FAU. He researches machine learning for blood flow simulation, X-ray imaging, and CT image reconstruction at FAU's PRL.
\end{IEEEbiographynophoto}

\vspace{-33pt}
\begin{IEEEbiographynophoto}{Moritz Koch}
received his B.Sc. and M.Sc. in physical geography and climate sciences from FAU. He is currently pursuing a Ph.D. at FAU's Institute of Geography and is part of the M3OCCA program.
\end{IEEEbiographynophoto}

\vspace{-33pt}
\begin{IEEEbiographynophoto}{Marcel Dreier}
earned his B.Sc. and M.Sc. in computer science from FAU, completing his Master's thesis on diffusion models in 2023. He is now a Ph.D. candidate at FAU's PRL, focusing on machine learning for radargrams.
\end{IEEEbiographynophoto}

\vspace{-33pt}
\begin{IEEEbiographynophoto}{Dakota Pyles}
received his M.Sc. in Geology from the University of Montana and is pursuing a Ph.D. at FAU. His research focuses on estimating frontal ablation and understanding tidewater glacier changes in the Arctic.
\end{IEEEbiographynophoto}

\vspace{-33pt}
\begin{IEEEbiographynophoto}{Thorsten Seehaus}
received his Ph.D. in geography from FAU in 2016 and is now a Junior Research Group Leader in GIS and Remote Sensing. His research focuses on using synthetic aperture radar imagery to monitor glacier changes and mass balances globally.
\end{IEEEbiographynophoto}

\vspace{-33pt}
\begin{IEEEbiographynophoto}{Matthias Braun}
received his Ph.D. in hydrology from the University of Freiburg and is a Professor at FAU. His research focuses on mass change of glaciers using in-situ observations, remote sensing, and modeling, with field campaigns in diverse regions worldwide.
\end{IEEEbiographynophoto}

\vspace{-33pt}
\begin{IEEEbiographynophoto}{Andreas Maier}
is a Professor and Head of the PRL at FAU, specializing in medical imaging, image and audio processing, and interpretable machine learning. He developed the PEAKS tool for speech intelligibility assessment and has led numerous research projects, including the ERC Synergy Grant ``4D nanoscope''.
\end{IEEEbiographynophoto}

\vspace{-33pt}
\begin{IEEEbiographynophoto}{Vincent Christlein}
earned his Ph.D. in computer science from FAU and is now an Academic Councilor leading the Computer Vision Group at FAU's PRL. His research covers a wide range of topics, including glacier segmentation, solar cell crack recognition, and document analysis.
\end{IEEEbiographynophoto}
\vfill

\twocolumn[{\centering{\Huge Supplementary Material - Comparison Study: Glacier Calving Front Delineation in Synthetic Aperture Radar Images With Deep Learning\par}}
\smallbreak
\vspace{2ex}]

\section*{Dataset}
\label{app:dataset}
All \ac{dl} systems are optimized, trained, and evaluated on the same dataset: \ac{caffe}, which was introduced by~Gourmelon et\,al.~\cite{Gourmelon_2022}.
This dataset was selected for the comparison because it is the largest, manually annotated, and publicly available \ac{SAR} calving front dataset that provides both \ac{SAR} images and corresponding labels.
Table~\ref{tab:datasets} provides an overview of publicly available glacier calving front datasets, underpinning the choice of \ac{caffe}.

\begin{table*}[htbp]
    \centering
    \caption{Overview of publicly available glacier calving front line datasets. The \textit{Annotation} entry refers to how the delineations were produced,  i.\,e., manually, by a network or by a model. Manually* means that only the center point of the calving front was digitized. \textit{Img. Avail.} stands for ``Images Available'', which indicates whether the satellite images used to delineate the calving front positions are provided along with the front positions.
    The entry \textit{\# Glaciers} gives the number of presented glaciers and \textit{\# Mapped Fronts} the number of glacier front delineations over all glaciers inherent in a dataset.
    The \textit{Res.} indicates the spatial resolution of images used for the mapping of the glacier fronts.
    The datasets are:
    [Lippl] \cite{lippl.2019_dataset},
    [King] \cite{king.2020_dataset},
    [Fausto] \cite{Fausto.2019_dataset},
    [Schild] \cite{Schild.2013_dataset},
    [Cheng] \cite{Cheng.2020_dataset},
    [ADD] \cite{add.2021_dataset},
    [GLIMS] \cite{glims.2018_dataset},
    [Loebel22] \cite{Loebel.2022},
    [Loebel23] \cite{Loebel.2023},
    [Loebel25] \cite{Loebel.2025},
    [Zhang19a] \cite{Zhang.2019_dataset},
    [Zhang19b] \cite{Zhang.2019_dataset_network},
    IceLines \cite{Baumhoer.2023},
    [ENVEO] \cite{enveo},
    [Zhang20a] \cite{Zhang.2020_dataset},
    [Zhang20b] \cite{Zhang.2020_dataset_network},
    [ESA] \cite{ESA.2019_dataset},
    TermPicks \cite{Goliber.2022}, 
    [Zhang23] \cite{Zhang.2023},
    [Li] \cite{Li.2024}, 
    and the \ac{caffe} dataset \cite{caffe} is denoted as \textit{CaFFe}. The table is adapted from~\cite{Gourmelon_2022}.}
    \label{tab:datasets}
    \begin{tabular}{lcccccccr}
        \toprule
        \textit{Modality} & \textit{Dataset} & \textit{Annotation} & \textit{Img. Avail.} & \textit{Area} & \textit{\# Glaciers} & \textit{\# Mapped Fronts} & \textit{Time Span} & \textit{Res.}\\
        \hline
        
        \multirow{13}{*}{Optical} & [Lippl] & Manually & \xmark & Antarctica & 26 & 656 & 2014 - 2018 & 15\,m \\
        \multirow{13}{*}{or} & [King] & Manually* & \xmark & Greenland & 234 & 128,442 & 1985 - 2018 & 30\,m \\
        \multirow{13}{*}{Multispectral} & [Fausto] & Manually & \xmark & Greenland & 47 & 1180 & 1999 - 2018 & 10 - 30\,m \\
        & [Schild] & Manually* & \xmark & Greenland & 2 & 1862 & 2001 - 2010 & 250\,m\\
        & \multirow{2}{*}{[Cheng]} & Manually & \cmark & \multirow{2}{*}{Greenland} & \multirow{2}{*}{66} & $>$\,1500 & \multirow{2}{*}{1972 - 2019} & \multirow{2}{*}{30\,m} \\
        & & \& Network & \xmark & & & 22,678 & & \\
        & \multirow{2}{*}{[ADD]} & Manually & \multirow{2}{*}{\xmark} & \multirow{2}{*}{Antarctica} & & & \multirow{2}{*}{Since 1843} & \\
        & & \& Semi-Automatic & & & & & & \\
        & \multirow{2}{*}{[GLIMS]} & Manually & \multirow{2}{*}{\xmark} & \multirow{2}{*}{Global} & \multirow{2}{*}{$\sim$200,000} & \multirow{2}{*}{546,300} & \multirow{2}{*}{Since 1750} & \\
        & &\& Semi-Automatic & & & & & & \\
        & \multirow{2}{*}{[Loebel22]} & \multirow{2}{*}{Manually} & \multirow{2}{*}{\cmark} & Greenland & \multirow{2}{*}{25} & \multirow{2}{*}{1,723} & \multirow{2}{*}{2013 - 2021} & \multirow{2}{*}{30 - 100\,m}\\
        & & & & \& Antarctica & & & & \\
        & [Loebel23] & Network & \xmark & Greenland & 23 & 9,243 & 2013 - 2021 & 15 - 100\,m\\
        & [Loebel25] & Network & \xmark & Antarctica & 42 & 4,817 & 2013 - 2023 & 30\,m\\
        \midrule
        
        \multirow{5}{*}{SAR} & [Zhang19a] & Manually & \xmark & Greenland & 1 & 159 & 2009 - 2015 & 3\,m \\
        & [Zhang19b] & Network & \xmark & Greenland & 1 & 159 & 2009 - 2015 & 3\,m \\
        &  \multirow{2}{*}{\textit{CaFFe}} & \multirow{2}{*}{Manually} & \multirow{2}{*}{\cmark} & Alaska, Antarctica & \multirow{2}{*}{7} & \multirow{2}{*}{681} & \multirow{2}{*}{1996 - 2020} & \multirow{2}{*}{6 - 20\,m}\\
        & & & & \& Greenland & & & & \\
        & IceLines & Network & \cmark & Antarctica & 51 & 19,400 & 2014 - 2023 & 40\,m\\
        
        \midrule
        
        \multirow{7}{*}{Mixed} & [ENVEO] & Manually & \xmark & Greenland & 28 & 1,090 & 1990 - 2016 & \\
        & [Zhang20a] & Manually & \xmark & Greenland & 3 & 2087 & 2002 - 2019 & 3 - 40\,m \\
        & [Zhang20b] & Network & \xmark & Greenland & 3 & 2087 & 2002 - 2019 & 3 - 40\,m \\
        & [ESA] & Manually & \xmark & Greenland & 28 & 1089 & 1990 - 2016 & 10 - 30 m\\
        & TermPicks & Manually & \xmark & Greenland & 278 & 39,060 & 1931 - 2021 & \\ 
        & [Zhang23] & Network & \xmark & Greenland & 295 & 278,239 & 1984 - 2021 & 10 - 30\,m\\
        & [Li] & Network & \xmark & Svalbard & 149 & 124,919 & 1985 - 2023 & 10 - 40\,m\\
        \bottomrule
        & & & & & & & \\
    \end{tabular}
\end{table*}
    
\begin{figure*}[htbp]
\captionsetup[subfigure]{font=scriptsize}
    \centering
    
    \subfloat[Jorum Glacier]{
        \includegraphics[width=0.29\textwidth]{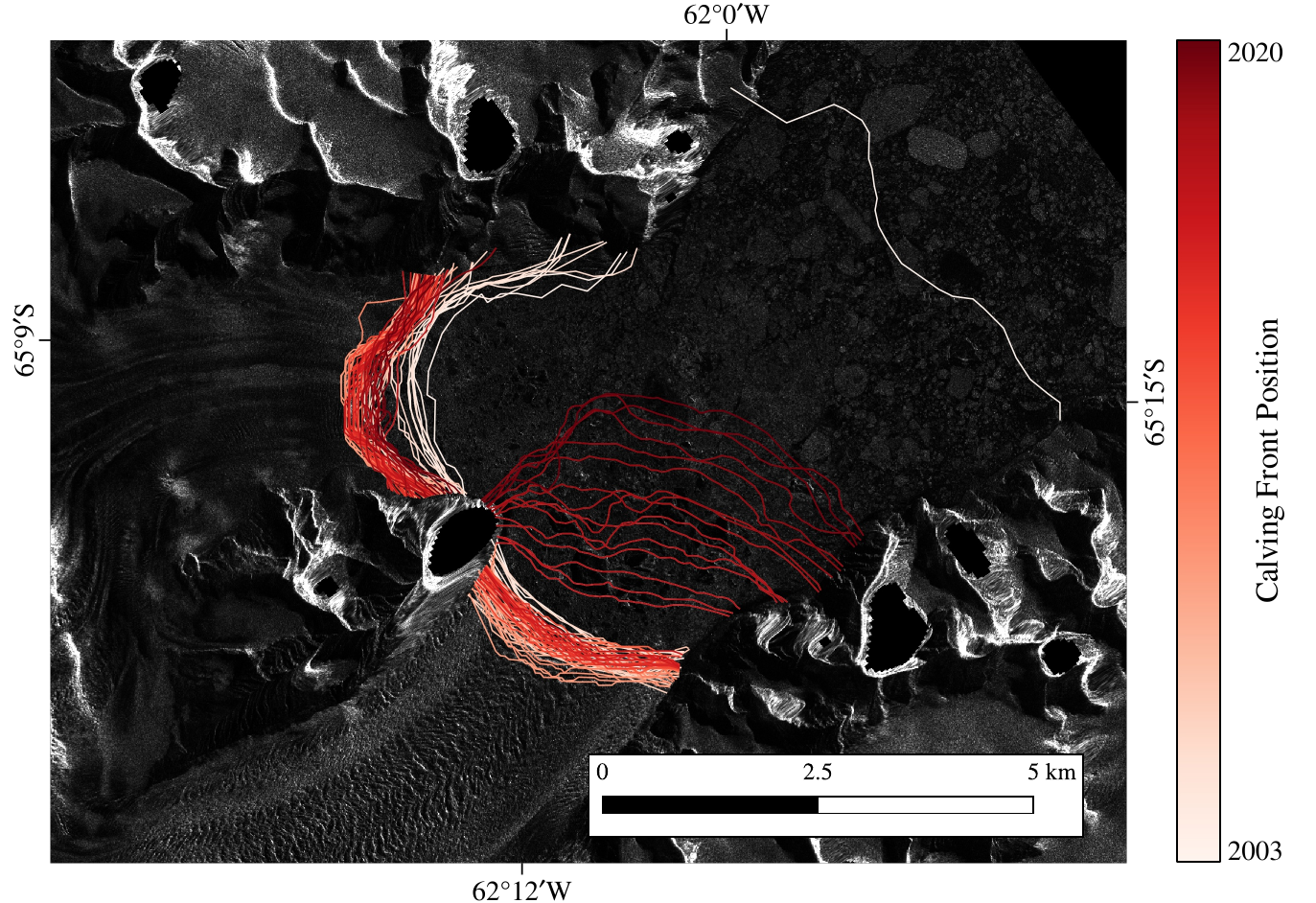}
        \label{fig:Jorum}
    }
    \subfloat[Dinsmoore-Bombardier-Edgeworth]{
        \includegraphics[width=0.25\textwidth]{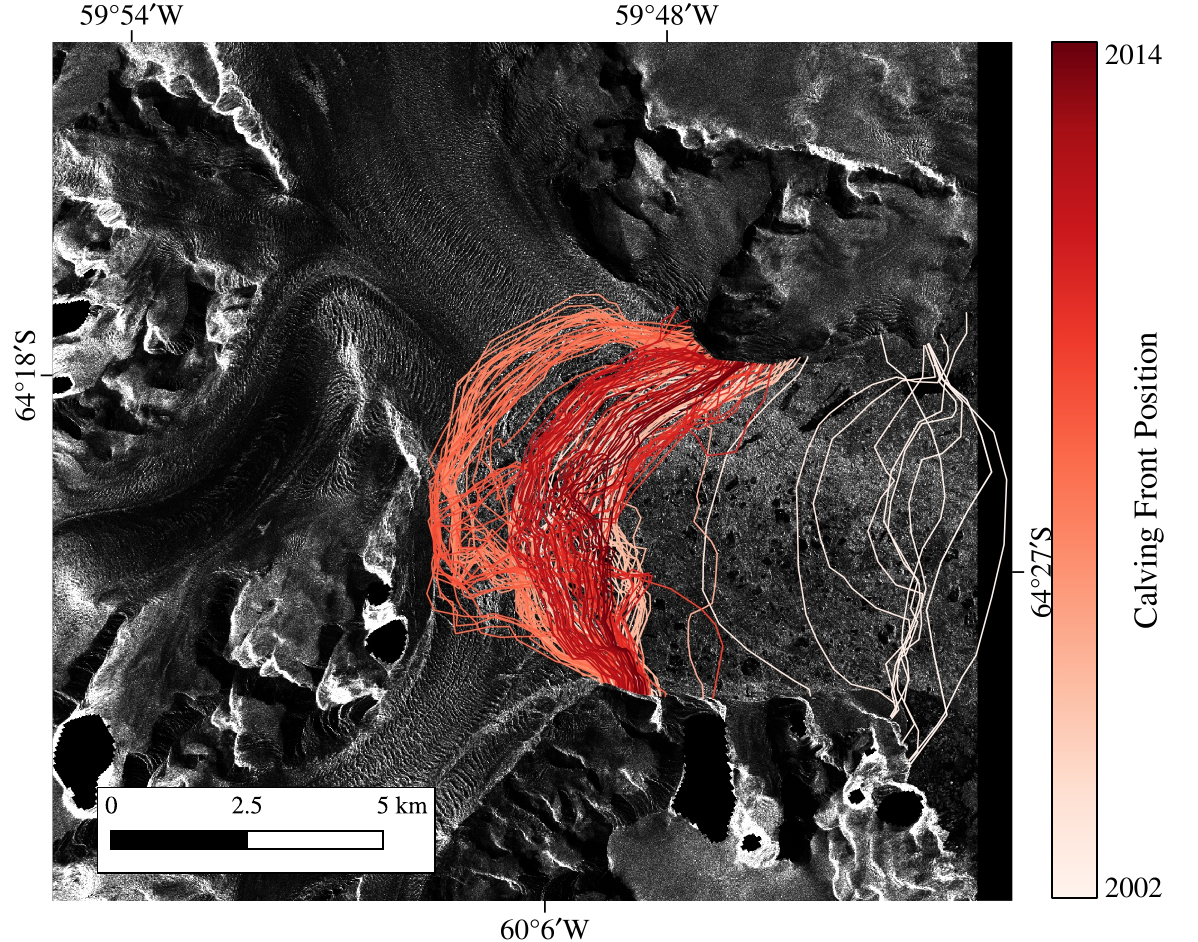}
        \label{fig:DBE}
    }  
   \subfloat[Crane Glacier]{
        \includegraphics[width=0.21\textwidth]{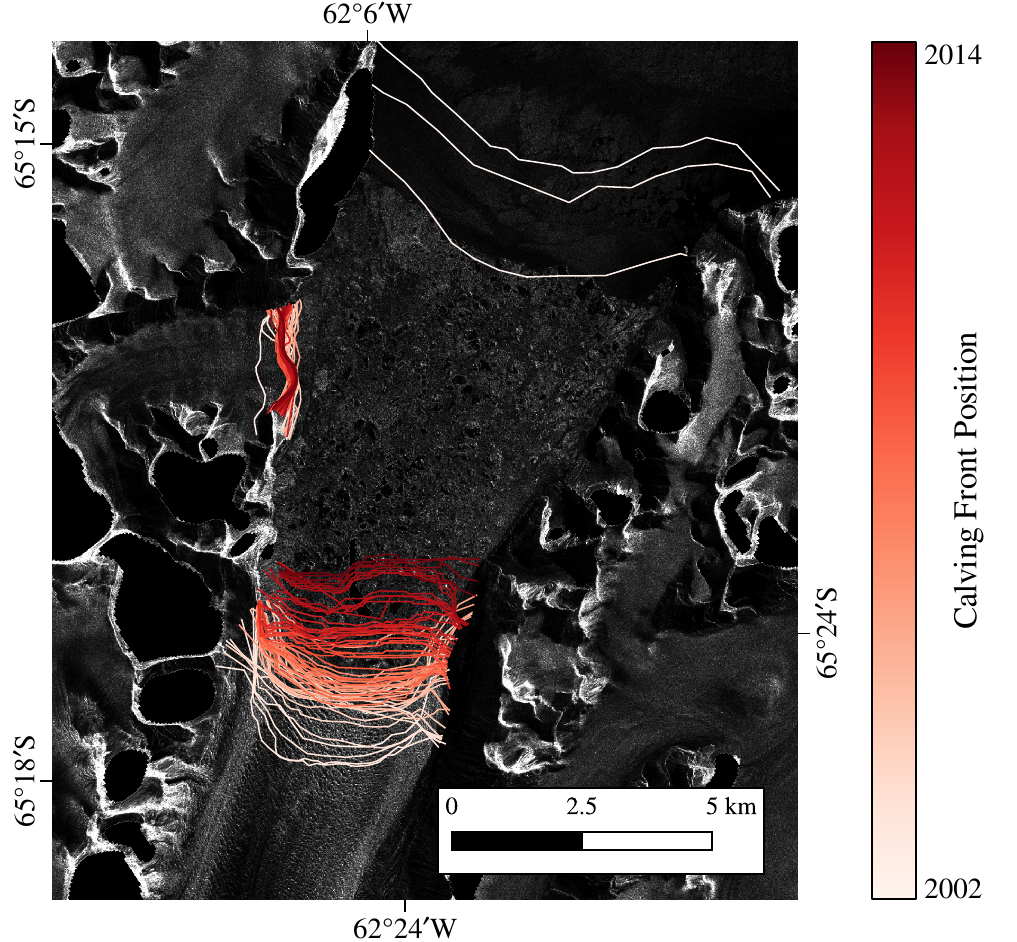}
        \label{fig:Crane}
    }
    \subfloat[Mapple Glacier]{
        \includegraphics[width=0.23\textwidth]{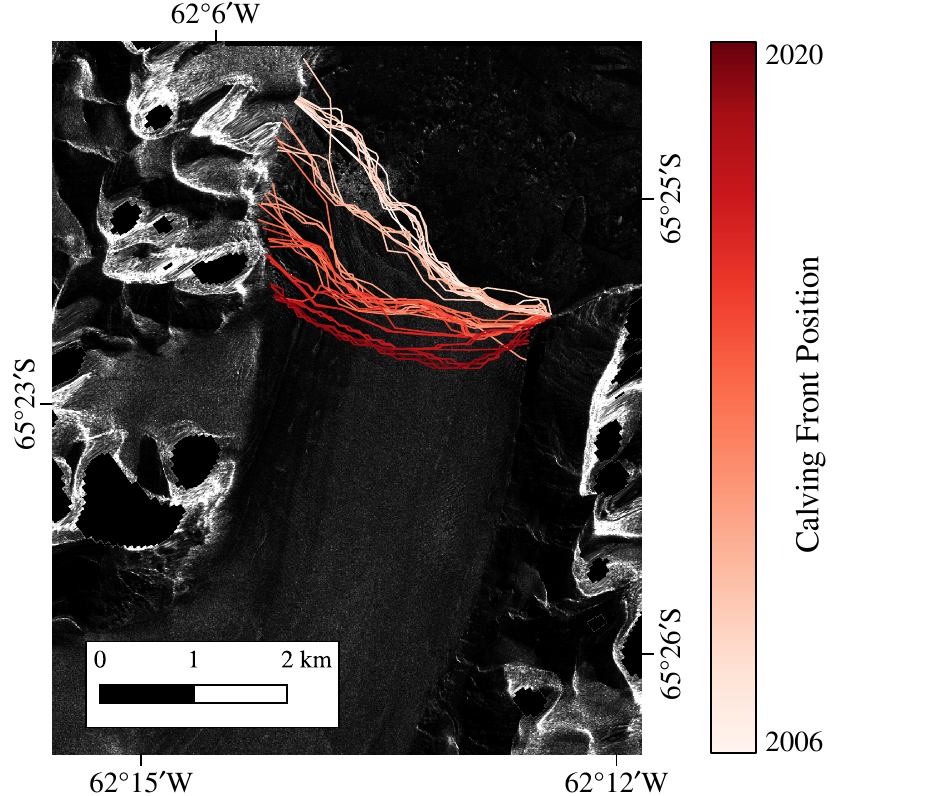}
        \label{fig:Mapple}
    }  
    \hfill
    
    \subfloat[Columbia Glacier]{
        \includegraphics[width=0.40\textwidth]{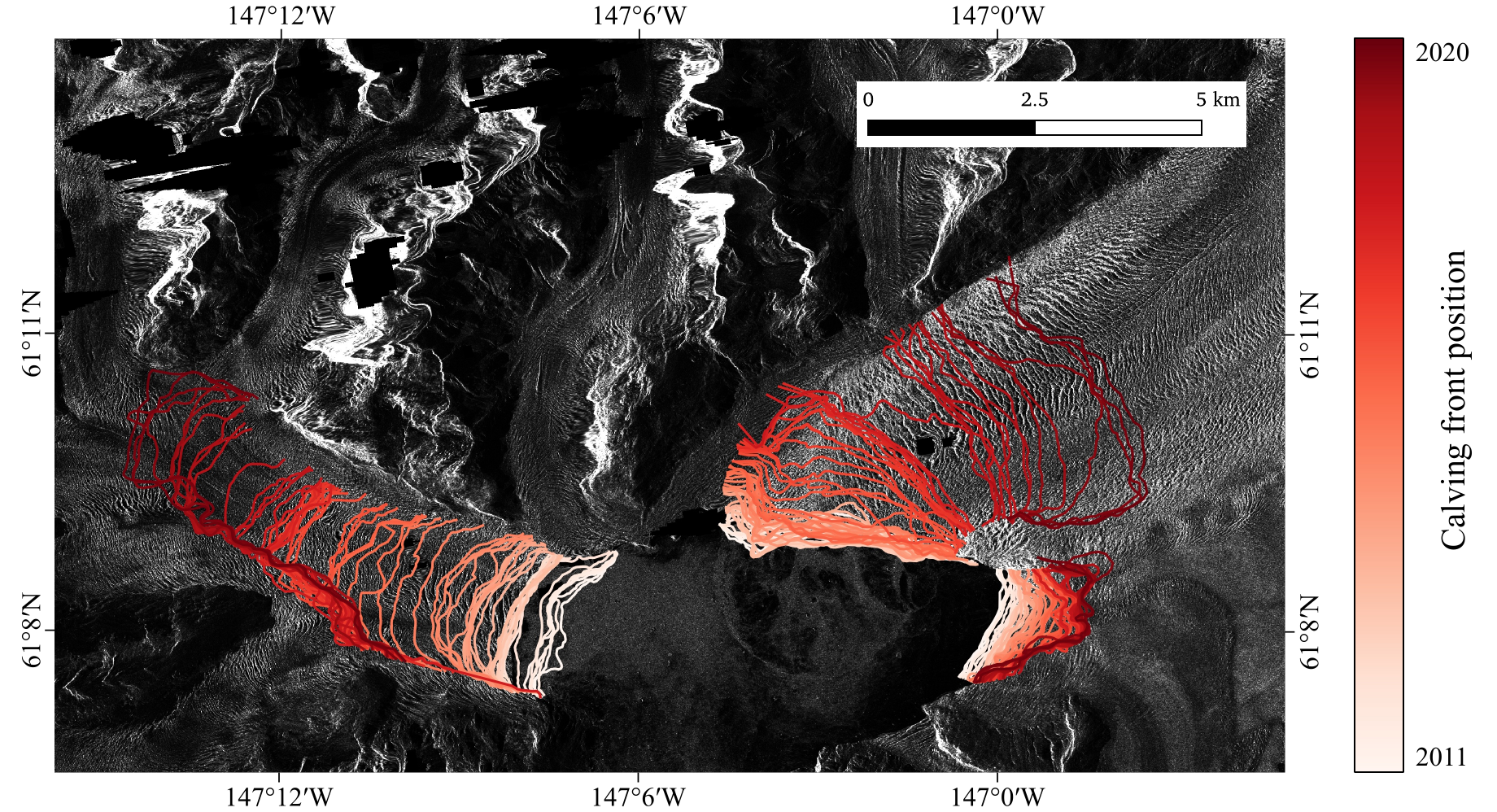}
        \label{fig:COL}
    }  
    \subfloat[Sjörgen-Inlet Glacier]{
        \includegraphics[width=0.35\textwidth]{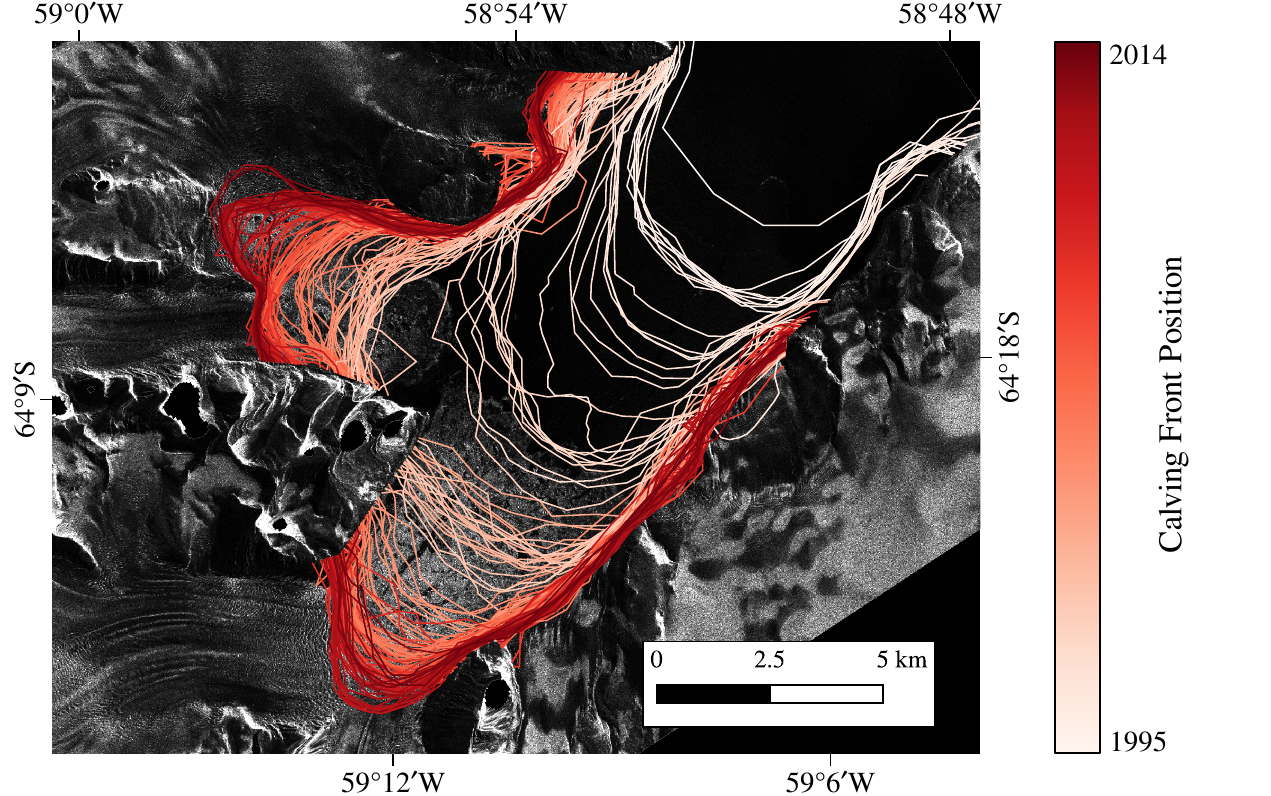}
        \label{fig:SI}
    } 
    \subfloat[Jakobshavn Glacier]{
        \includegraphics[width=0.23\textwidth]{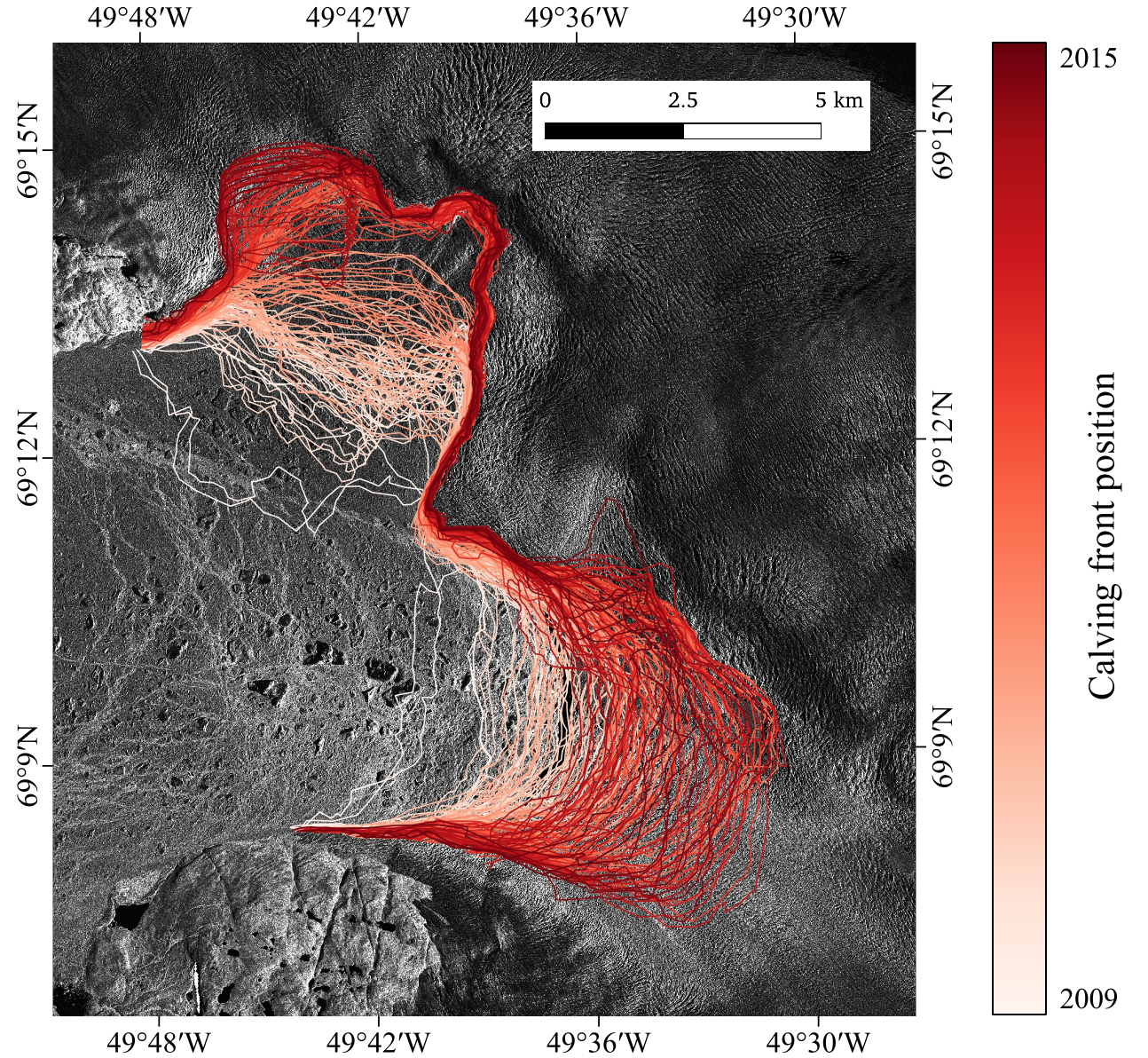}
        \label{fig:JAC}
    } 
    \caption{Visualization of calving front position changes within the \ac{caffe} dataset. \ac{SAR} imagery is provided by DLR, ESA, and ASF.
    Illustrations are taken from Herrmann et\,al.~\cite{Herrmann.2023}. Best viewed zoomed in.}
    \label{fig:calving_front_changes}
\end{figure*}

The \ac{caffe} dataset encompasses 681 \ac{SAR} images from seven tidewater glaciers dating from 1996 to 2020.
Five glaciers are located on the Antarctic Peninsula, one in Greenland, and one in Alaska.
The calving front position changes of the seven glaciers throughout the time span covered by \ac{caffe} are illustrated in Fig.~\ref{fig:calving_front_changes}. GIFs of the changes can be found at \url{https://zenodo.org/records/15173191}.

The dataset comprises multiple missions (ERS-1/2, Envisat, RADARSAT-1, ALOS PALSAR, \ac{tsx}, \ac{tdx}, and Sentinel-1).
The imagery was multi-looked, calibrated, geo-referenced, and ortho-rectified.
Image sizes in pixels vary between $405\,\times\,382$ and $3561\,\times\,2768$, depending on the sensor and captured glacier, while the spatial resolution ranges between \SI{6}{} and \SI{20}{\meter} per pixel.

In addition to the labels, \ac{caffe} provides a bounding box for each image that shows the region of interest and is used to exclude static glacier fronts, which were not manually labeled.
For the zone labels, the calving front is extracted during post-processing as the edge between glacier and ocean zones within this bounding box to not compare with non-manual annotations of static fronts.
For the front labels, the prediction inside the bounding box is taken as the final calving front prediction.
For the evaluation of the trained \ac{dl} systems, the dataset contains a test set of 122 images, which are withheld during training.
Which part of the training set is used for the validation during hyperparameter optimization is the user's choice.
The test set includes all images of the Columbia Glacier in Alaska and the Mapple Glacier on the Antarctic Peninsula. 
Hence, both the Mapple and the Columbia Glacier are not seen during training and validation, only at test time.
This intercontinental spread of the test set and the spatial separation of the test and training sets ensures that the evaluation assesses the reproducibility of the \ac{dl} systems' performance in a global context, thus ensuring generalizability to unseen geographic locations.

\section*{Evaluation metrics}
    For the evaluation, two metrics are employed, which were both introduced by Gourmelon et\,al.~\cite{Gourmelon_2022} alongside the benchmark dataset: the \acf{mde} and the number of images with no predicted front. 
    Both metrics are computed after post-processing. 
    The number of images with no predicted front counts the images where no front is found by the \ac{dl} system.
    The \ac{mde} evaluates the distance between the predicted locations of the calving fronts and the locations of the manually labeled calving fronts.
    It is calculated as:
    \begin{multline}\label{eq:mean_distance_error}
    \mathrm{MDE}(\mathcal{I}) = \frac{1}{\sum_{(\mathcal{P}, \mathcal{Q}) \in \mathcal{I}} (|\mathcal{P}| + |\mathcal{Q}|)} \cdot \\
    \sum_{(\mathcal{P}, \ \mathcal{Q}) \in I} \bigg( \sum_{\vec{p} \in \mathcal{P}} \min_{\vec{q} \in \mathcal{Q}} \lVert \vec{p}-\vec{q} \rVert_2 + \sum_{\vec{q} \in \mathcal{Q}} \min_{\vec{p} \in \mathcal{P}} \lVert \vec{p}-\vec{q} \rVert_2 \bigg)
    \end{multline}
    whereas $\mathcal{I}$ is the set of all images where a front is predicted, $|.|$ the cardinality of a set, $\mathcal{P}$ all ground truth front pixels of one image, and $\mathcal{Q}$ all predicted front pixels of the same image. 
    Images with no predicted front pixels are ignored during the calculation. 
        
\begin{figure}[H]\captionsetup[subfigure]{font={scriptsize, sf}}
        \centering
        \subfloat[\ac{mde}: 1.64; \ac{iou}: 0.11]{
            \includegraphics[width=0.145\textwidth]{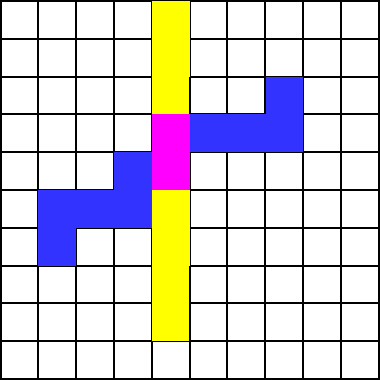}
            \label{fig:iou_mde_bad}
        }
        \subfloat[\ac{mde}: 3.64; \ac{iou}: 0.00]{
            \includegraphics[width=0.145\textwidth]{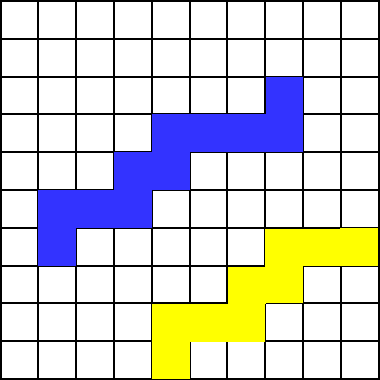}
            \label{fig:iou_mde_far}
        }
        \subfloat[\ac{mde}: 1.13; \ac{iou}: 0.00]{
            \includegraphics[width=0.145\textwidth]{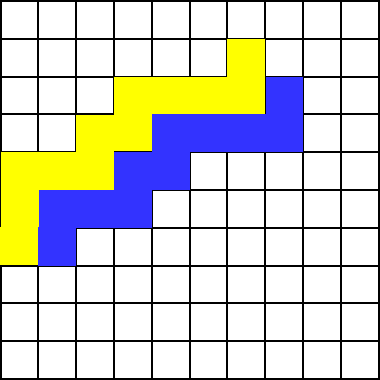}
            \label{fig:iou_mde_good}
        }
        \hfill   
        \caption{Examples of two poorly predicted calving front delineations (a, b) and one well-predicted delineation (c). The predicted front is shown in yellow, the ground truth in blue, and overlapping areas indicating a perfect match in pink. The \ac{mde} is measured in pixels and can be converted to meters by multiplying by the image resolution.}
        \label{fig:iou_vs_mde}
    \end{figure}

    Figure~\ref{fig:iou_vs_mde} illustrates three examples of predicted calving fronts, highlighting why \ac{iou} is not a suitable evaluation metric for this task. In Fig.~\ref{fig:iou_mde_good}, the predicted front closely follows the ground truth, resulting in a low \ac{mde}. However, since no pixels overlap exactly, the \ac{iou} is 0, despite the prediction being accurate. In contrast, Fig.~\ref{fig:iou_mde_far} depicts a prediction that is farther from the ground truth, leading to a higher \ac{mde}, yet its \ac{iou} remains identical to that of the well-predicted front in Figure~\ref{fig:iou_mde_good}. Lastly, Fig.~\ref{fig:iou_mde_bad} shows a less accurate prediction that happens to cross the true front, leading to a higher \ac{iou} than Fig.~\ref{fig:iou_mde_good}, even though the actual delineation is worse. These examples demonstrate that \ac{iou} fails to reliably differentiate between good and poor predictions, whereas \ac{mde} consistently reflects prediction quality, making it the more appropriate metric for calving front delineation. 

    The \ac{mde} is closely related to two other metrics:
    The Average Symmetric Surface Distance~\cite{Yeghiazaryan_2018}, which is a well-known metric in medical image segmentation, and the Chamfer Distance~\cite{Thiel_1994}, which is used for distance calculations between point clouds.

    A trade-off exists between the \ac{mde} and the number of images with no predicted front. 
    For challenging fronts, the model typically either fails to predict a front entirely or generates an inconsistent and distant prediction. In the first case, the number of images without a predicted front increases, but the \ac{mde} remains low since these difficult cases are excluded from its computation. In the second case, fewer images lack a predicted front, but the \ac{mde} increases as it incorporates poorly predicted fronts from challenging images.

\section*{Methodology}
\label{app:methods}

\subsection{Deep Learning system comparison setup}
\label{app:methods_ai_comparisons}
\noindent For the comparison, we selected studies that take satellite imagery as input to a neural network and extract either the calving front of a marine-terminating glacier or the coastline of an ice shelf.
Only three studies are excepted: We do not evaluate the studies of Baumhoer et\,al.~\cite{Baumhoer.2019} and Zhang et\,al.~\cite{Zhang.2019} as both were superseded by their successor models Heidler et\,al.~\cite{Heidler.2021} and Zhang et\,al.~\cite{Zhang.2021}. 
Similarly, we do not evaluate the study by Heidler et\,al.~\cite{Heidler.2023} because their network inherently requires that there is only one coastline and not multiple calving fronts in a single image.
Therefore, Heidler et\,al.~\cite{Heidler.2023}'s model is not applicable to the \ac{caffe} dataset, which shows multiple calving fronts in several images.
Additionally, we explore the performance of foundation models~--~large deep neural networks that have been trained on enormous amounts of data and aim to handle various downstream tasks for which only minimal fine-tuning is required~\cite{bommasani.2022}.
For segmentation tasks, several foundation models have emerged recently~\cite{Kirillov.2023, Wang.2023, Zou.2023}.
As a representative, we choose to evaluate the promptable \ac{sam}~\cite{Kirillov.2023} in the advertised zero-shot manner, i.\,e., no fine-tuning is performed.

Adjustments are necessary to enable comparison between the algorithms of the different studies. 
We regard each paper's code as a system, meaning we try to minimize the adaptations we perform. 
The pre-processing including patch extraction, the \ac{dl} model, and post-processing are kept unchanged as much as possible. 
We only adapt parts of the pipeline to make the code run with the employed dataset, which might differ from the dataset initially used to train and test the code.
For example, if the respective dataset contains binary zone segmentation masks (glacier vs. ocean), the loss function will likely be or contain binary cross-entropy (BCE), which we will have to adapt to categorical cross-entropy to work with our multi-class zone segmentation masks.
We tweak the pipelines of the \ac{dl} systems so that they take in \ac{SAR} imagery and learn to extract the calving front by either using \ac{caffe}'s zone or front labels.
Systems that were previously trained on binary coastline masks or binary calving front masks are trained on \ac{caffe}'s binary calving front masks.
Systems previously trained on binary ocean masks or multi-class segmentation masks are trained on \ac{caffe}'s multi-class zone masks.
Any manual steps in the pre-or post-processing of the systems are skipped, as we want to test the systems' ability to delineate calving fronts fully automatically. 
Since most of the standard pre-processing is already complete for the benchmark dataset (see Sec. \ref{app:dataset}), only pre-processing techniques related to the specific architecture of the neural network need to be applied.
Concerning the post-processing, we add bounding box masking and the deletion of too short fronts to the end of each post-processing schema.
The position of the bounding boxes that exclude the fronts of static glaciers and the minimum length of the fronts of the dynamic glaciers are dataset-specific prior knowledge, which Gourmelon et\,al.~\cite{Gourmelon_2022} use in their post-processing scheme.  
Hence, this prior knowledge needs to be integrated into the other systems to keep the comparison fair.
Bounding box masking alters the prediction so that outside of the bounding box, all pixels belong to the background.
The minimum length of a front for the given dataset is \SI{1.5}{\km}.
All predicted front pixels belonging to a connected line shorter than half of this minimum length are set to background.
System-specific adjustments can be found in the following section (Sec.~\ref{app:dl_systems}).

To ensure fairness in the comparison, we re-optimize the hyperparameters on the benchmark's training set.
Therefore, we split the training set into a train set and a validation set, whereas we train the network on the train set and evaluate it on the validation set. 
The split ratio is taken from the respective study.
For the optimization, we chose the hyperparameters that were specified as being optimized in the corresponding publication.
Additionally, if not already mentioned in the publication, we optimize the learning rate or the base and maximum learning rate if a scheduler is used. 
Next, we trained each system five times.
The number of epochs trained was calculated so that the model would see 150 times the number of pixels in the training set.
For the calculation, the amount of patch overlap, resizing, and the number of iterations in one epoch had to be taken into account.
Lastly, the five trained systems are evaluated on \ac{caffe}'s test set, and the mean and standard deviation of the evaluation metrics are computed over the five runs. Re-training the models only five times was a compromise between statistical interpretability and computational feasibility. Given that parametric significance tests are not robust with such a small sample size, non-parametric tests were employed for the statistical analysis of the results.

\subsection{Deep Learning systems}
\label{app:dl_systems}
\noindent In this section, we will review the methodologies of the compared codes and the adjustments made.
Tables~\ref{tab:Masks_and_Models},~\ref{tab:Image_processing}, and~\ref{tab:augmentations} provide a summary of the segmentation masks originally used, the network architecture on which each system is based, the original strategy for dealing with image sizes, and the augmentations performed.
Table~\ref{tab:hyper-parameters} lists the values of the re-optimized hyper-parameters for each model in the comparison.

\begin{table*}[htbp]
        \centering
        \caption{Overview of the original segmentation masks and base models. \textit{BCF} refers to segmentation masks that distinguish between the calving front and background, \textit{BCL} refers to segmentation masks that distinguish between coastline and background, \textit{BO} refers to segmentation masks that distinguish between ocean and non-ocean, \textit{Multi} refers to segmentation masks that distinguish between multiple landscape zones, and \textit{Conv.} is the abbreviation for convolutional.}
        \label{tab:Masks_and_Models}
        \begin{tabular}{p{0.1\textwidth} p{0.04\textwidth} p{0.06\textwidth} p{0.06\textwidth} p{0.06\textwidth} p{0.07\textwidth} p{0.10\textwidth} p{0.12\textwidth} p{0.07\textwidth} p{0.07\textwidth}} 
            \toprule
            && \multicolumn{4}{c}{\textit{Segmentation Mask}} & \multicolumn{4}{c}{\textit{Base Model}}\\
            \textit{Paper} && \textit{BCL} & \textit{BCF} & \textit{BO} & \textit{Multi} & \textit{Conv. U-Net} & \textit{DeepLabv3+} & \textit{ViT} & \textit{VGG16}\\
            \midrule
            
            Cheng 
            && \checkmark & & \checkmark & & & \checkmark & & \\
            Davari (a) 
            && & \checkmark & & & \checkmark & & & \\
            Davari (b) 
            && & \checkmark & & & \checkmark & & & \\
            \multirow{ 2}{*}{Gourm. (22)}  
            & Front & & \checkmark & & & \checkmark & & & \\
            & Zones & & & & \checkmark & \checkmark & & & \\
            Gourm. (23) 
            && & & & \checkmark & \checkmark & & & \\
            Hartmann 
            && & & \checkmark & & \checkmark & & & \\
            Heidler 
            && \checkmark & & \checkmark & & \checkmark & & & \\
            Herrmann 
            && & \checkmark & & \checkmark & \checkmark & & & \\
            Holzmann 
            && & \checkmark & & & \checkmark & & & \\
            Kirillov 
            && & & \checkmark & & & & \checkmark & \\
            Loebel 
            && & & \checkmark & & \checkmark & & & \\
            Marochov 
            && & & & \checkmark & & & & \checkmark \\
            Mohajerani 
            && & \checkmark & & & \checkmark & & & \\
            Periya. 
            && & & \checkmark & & \checkmark & & & \\
            Wu (a) 
            && & & & \checkmark & \checkmark & & & \\
            Wu (b) 
            && & & & \checkmark & & & \checkmark & \\
            Zhang (21) 
            && & & \checkmark & & & \checkmark & & \\
            Zhang (23) 
            && & & \checkmark & & & \checkmark & & \\
            Zhu 
            && & & \checkmark & & & \checkmark & \checkmark & \\
            \bottomrule
            && & & & & & & & \\
        \end{tabular}
    \end{table*}

\begin{table*}[htbp]
        \centering
        \caption{Overview of original strategies for dealing with the image size. $1024\,\times\,X$ or $X\,\times\,1024$ implies that the image is resized such that the longest side of the image is $1024$ pixels long while preserving the aspect ratio. The abbreviation a.\,n. stands for as necessary for covering the entire glacier region with the specified patch size. $^*$During test time, the images were fed into the network as a whole.}
        \label{tab:Image_processing}
        \begin{tabular}{p{0.18\textwidth} p{0.06\textwidth} p{0.18\textwidth}  p{0.14\textwidth} p{0.15\textwidth} p{0.14\textwidth}} 
            \toprule
            && \textit{Resizing} & \multicolumn{3}{c}{\textit{Patch Extraction}}\\
            \textit{Paper} && \textit{Size} & \textit{Patch Size} & \textit{Train-time Overlap} & \textit{Test-time Overlap}\\
            \midrule
            
            Cheng et\,al.~\cite{Cheng.2021} && $256\,\times\,256$ & $224\,\times\,224$ & / & 208 \\
            Davari et\,al.~\cite{Davari_Baller.2021} && $512\,\times\,512$ & / & / & / \\
            Davari et\,al.~\cite{Davari_Islam.2021} && / & $256\,\times\,256$ & 0 & 0 \\
            Gourmelon et\,al.~\cite{Gourmelon_2022} && / & $256\,\times\,256$ & 0 & 128 \\
            Gourmelon et\,al.~\cite{Gourmelon.2023} && / & $256\,\times\,256$ & 0 & 128 \\
            Hartmann et\,al.~\cite{Hartmann.2021} && / & $256\,\times\,256$ & 0 & 0 \\
            Heidler et\,al.~\cite{Heidler.2021} && / & $768\,\times\,768$ & 384 & 384 \\
            Herrmann et\,al.~\cite{Herrmann.2023} && / & $1280\,\times\,1024$ & 0 & 640, 512 \\
            Holzmann et\,al.~\cite{Holzmann.2021} && $512\,\times\,512$ & / & / & / \\
            Kirillov et\,al.~\cite{Kirillov.2023} && $1024\,\times\,X$ or $X\,\times\,1024$ & / & / & / \\
            Loebel et\,al.~\cite{Loebel.2022} && / & $512\,\times\,512$ & a.\,n. & a.\,n. \\
            \multirow{ 2}{*}{Marochov et\,al.~\cite{Marochov.2021}} &Phase 1& / & $50\,\times\,50$ & 30 & 0 \\
            &Phase 2& / & $15\,\times\,15$ & 14 & 14 \\
             \multirow{ 2}{*}{Mohajerani et\,al.~\cite{Mohajerani.2019}} &Training& $150\,\times\,240$ & / & / & / \\
             &Testing& $200\,\times\,300$ & / & / & / \\
            Periyasamy et\,al.~\cite{Periyasamy.2022} && / & $256\,\times\,256$ & 0 & /$^*$ \\
            \multirow{ 2}{*}{Wu et\,al.~\cite{Wu.2023_1}} & Target & / & $288\,\times\,288$ & 0 & 0 \\
            & Context & $288\,\times\,288$ & $576\,\times\,576$ & 288 & 288 \\
            \multirow{ 2}{*}{Wu et\,al.~\cite{Wu.2023_2}} & Target & / & $224\,\times\,224$ & 0 & 0 \\
            & Context & $224\,\times\,224$ & $448\,\times\,448$ & 224 & 224 \\
            Zhang et\,al.~\cite{Zhang.2021} && / & $960\,\times\,720$ & $320, 240$ & $384, 288$ \\
            Zhang et\,al.~\cite{Zhang.2023} && / & $960\,\times\,720$ & $320, 240$ & $384, 288$ \\
            Zhu et\,al.~\cite{Zhu.2023} && / & $384\,\times\,384$ & $192, 192$ & $192, 192$ \\
            \bottomrule
            && & & & \\
        \end{tabular}
    \end{table*}
    
    \begin{table*}[htbp]
        \centering
        \caption{Overview of used augmentations. \textit{Rot.} stands for rotations; \textit{Sharp.} stands for sharpening; \textit{Crop} includes cropping and rescaling; \textit{Bright.} stands for brightness adjustments; \textit{Elastic} refers to multiple simultaneous elastic transforms; \textit{Gray} stands for inverting grayscale intensities; 
        \textit{Other} includes blurring, contrast changes, simulation of a lower resolution, and gamma augmentations that comprise a grayscale intensity inversion and a nonlinear intensity transformation.}
        \label{tab:augmentations}
        \begin{tabular}{
        p{0.2\textwidth}p{0.09\textwidth} p{0.05\textwidth} p{0.05\textwidth} p{0.05\textwidth} p{0.05\textwidth} p{0.05\textwidth} p{0.05\textwidth} p{0.05\textwidth} p{0.05\textwidth} p{0.05\textwidth}} 
            \toprule
            \textit{Paper} & & \textit{Flips} & \textit{Rot.} & \textit{Noise} & \textit{Sharp.} & \textit{Crop} & \textit{Bright.} & \textit{Elastic} & \textit{Gray} & \textit{Other}\\
            \midrule
            Cheng et\,al.~\cite{Cheng.2021} && \checkmark & \checkmark & \checkmark & \checkmark & \checkmark & & & & \\
            Davari et\,al.~\cite{Davari_Baller.2021} && \checkmark & \checkmark & & & & & & & \\
            Davari et\,al.~\cite{Davari_Islam.2021} && \checkmark & \checkmark & & & & & & & \\
            Gourmelon et\,al.~\cite{Gourmelon_2022} && \checkmark & \checkmark & \checkmark & & & \checkmark & \checkmark & & \\
            Gourmelon et\,al.~\cite{Gourmelon.2023} && \checkmark & \checkmark & \checkmark & & & \checkmark & \checkmark & & \\
            Hartmann et\,al.~\cite{Hartmann.2021} 
            && & & & & & & & & \\
            Heidler et\,al.~\cite{Heidler.2021} && \checkmark & \checkmark & & & & & & & \\
            Herrmann et\,al.~\cite{Herrmann.2023} && \checkmark & \checkmark & \checkmark & & \checkmark & \checkmark & & & \checkmark \\
            Holzmann et\,al.~\cite{Holzmann.2021} && \checkmark & \checkmark & & & & & & & \\
            Kirillov et\,al.~\cite{Kirillov.2023} && & & & & & & & & \\
            Loebel et\,al.~\cite{Loebel.2022} && \checkmark & \checkmark & & & & & & & \\
            \multirow{ 2}{*}{Marochov et\,al.~\cite{Marochov.2021}} 
            & Phase 1 & & \checkmark & \checkmark & & & & & & \\
            & Phase 2 & & & & & & & & & \\
            Mohajerani et\,al.~\cite{Mohajerani.2019} && \checkmark & & & & & & & \checkmark & \\
            Periyasamy et\,al.~\cite{Periyasamy.2022} && \checkmark & \checkmark & & & & & & & \\
            Wu et\,al.~\cite{Wu.2023_1} && \checkmark & \checkmark & & & & & & & \\
            Wu et\,al.~\cite{Wu.2023_2} && \checkmark & \checkmark & & & & & & & \\
            Zhang et\,al.~\cite{Zhang.2021} && \checkmark & \checkmark & & & & & & & \\
            Zhang et\,al.~\cite{Zhang.2023} && \checkmark & \checkmark & & & & & & & \\
            Zhu et\,al.~\cite{Zhu.2023} && \checkmark & & & & & & & & \\
            \bottomrule
            && & & & & & & & & \\
        \end{tabular}
    \end{table*}

    \begin{table*}[htbp]
        \centering
        \caption{Overview of re-optimized hyper-parameter values. The abbreviations are as follows: learning rate (LR); decay parameter for the distance map ($\gamma$); binarization threshold (bin. thres.); $w$, $k$, and $R$ are the parameters for the (optimized) distance map loss; Tile size is the tile size in the first phase of Mohajerani et\,al.~\cite{Mohajerani.2019}; loss function weighting (loss weight.); loss function on the zone predictions ($L_z$); loss function on the front predictions ($L_f$).}
        \label{tab:hyper-parameters}
        \begin{tabular}{p{0.17\textwidth} p{0.06\textwidth} p{0.09\textwidth} p{0.02\textwidth} p{0.05\textwidth} p{0.03\textwidth} p{0.02\textwidth} p{0.02\textwidth} p{0.02\textwidth} p{0.02\textwidth} p{0.06\textwidth} p{0.16\textwidth}} 
            \toprule
            \textit{Paper} &  & \textit{LR} & \textit{$\gamma$} & \textit{Dilation kernel} & \textit{bin. thres.} & \textit{w} & \textit{k} & \textit{R} & \textit{Tile size} & \textit{Kernel size} & \textit{Loss weight.} \\
            \midrule
            
            Cheng et\,al.~\cite{Cheng.2021} && $[3e^{-5}; 3e^{-4}]$ & / & / & / & / & / & / & / & / & $0.01 * L_z + 0.99 * L_f$\\
            Davari et\,al.~\cite{Davari_Baller.2021} && $1e^{-2}; 1e^{-6}$ & $8$ & $2\,\times\,2$ & / & / & / & / & / & / & / \\
            Davari et\,al.~\cite{Davari_Islam.2021} && $[1e^{-7}, 1e^{-4}]$ & / & / & $0.05$ & $3$ & $0.1$ & $1$ & / & / & / \\
            \multirow{ 2}{*}{Gourmelon et\,al.~\cite{Gourmelon_2022}} & Front & / & / & / & / & / & / & / & / & / & / \\
             & Zones & / & / & / & / & / & / & / & / & / & / \\
            Gourmelon et\,al.~\cite{Gourmelon.2023} && / & / & / & / & / & / & / & / & / & / \\
            Hartmann et\,al.~\cite{Hartmann.2021} && $1e^{-4}$ & / & / & / & / & / & / & / & / & / \\
            Heidler et\,al.~\cite{Heidler.2021} && $[4e^{-5}, 2e^{-4}]$& / & / & / & / & / & / & / & / & / \\
            Herrmann et\,al.~\cite{Herrmann.2023} && / & / & / & / & / & / & / & / & / & / \\
            Holzmann et\,al.~\cite{Holzmann.2021} && $[1e^{-6}, 1e^{-3}]$ & / & / & $0.5$ & $8$ & / & / & / & / & / \\
            \multirow{ 2}{*}{Kirillov et\,al.~\cite{Kirillov.2023}} & Iterative & / & / & / & / & / & / & / & / & / & / \\
            & Parallel & / & / & / & / & / & / & / & / & / & / \\
            Loebel et\,al.~\cite{Loebel.2022} && $1e^{-4}$ & / & / & / & / & / & / & / & / & / \\
            Marochov et\,al.~\cite{Marochov.2021} && $1e^{-3}; 1e^{-3}$ & / & / & / & / & / & / & $32$ & $15\,\times\,15$ & / \\
            Mohajerani et\,al.~\cite{Mohajerani.2019} && $1e^{-3}$ & / & / & / & / & / & / & / & / & / \\
            Periyasamy et\,al.~\cite{Periyasamy.2022} && $[1e^{-7}, 1e^{-3}]$ & / & / & / & / & / & / & / & / & $0.8 * CE + 0.2* Dice$ \\
            Wu et\,al.~\cite{Wu.2023_1} && / & / & / & / & / & / & / & / & / & / \\
            Wu et\,al.~\cite{Wu.2023_2} && / & / & / & / & / & / & / & / & / & / \\
            Zhang et\,al.~\cite{Zhang.2021} && $1e^{-4}$ & / & / & / & / & / & / & / & / & / \\
            Zhang et\,al.~\cite{Zhang.2023} && $0.05$ & / & / & / & / & / & / & / & / & / \\
            Zhu et\,al.~\cite{Zhu.2023} && $0.05$ & / & / & / & / & / & / & / & / & / \\
            \bottomrule
             && & & & & & & & & & \\
        \end{tabular}
    \end{table*}

\subsection*{\texorpdfstring{Cheng et\,al.~\cite{Cheng.2021}}{}}
    \noindent Cheng et\,al.~\cite{Cheng.2021} employ the DeepLabv3~\cite{Chen_2018_ECCV} architecture to segment optical and \ac{SAR} imagery into land and sea, including ice mélange.
    They employ the Xception model~\cite{Chollet_2017_CVPR} as the backbone, like in the original DeepLabv3 paper, but add Atrous Spatial Pyramid Pooling \cite{Chen_2018_ECCV} between the encoder and decoder. Their loss function is a weighted sum of the cross-entropy and the Dice loss.
    Cheng et\,al.~\cite{Cheng.2021}'s network, called CALFIN, outputs two probability masks: sea versus land and coastline versus background.
    Their training and testing datasets consist of 1,541 Landsat images of Greenland and 232 Sentinel-1A/B images of Antarctica.
    The dataset is part of the published dataset of Cheng et\,al.~\cite{Cheng.2020_dataset}.
    All images are centered over basins and are precision- and terrain-corrected.
    Only images with low cloud coverage and a low number of NODATA pixels are further considered.
    Next, they are resized to 256\,$\times$\,256 pixels and enhanced using pseudo-HDR toning (HDR) and shadows/highlights (S/H) options in Adobe Photoshop.
    Before feeding the input into CALFIN, patches of size 224\,$\times$\,224 are extracted.
    The input patches are augmented randomly on the fly and have three channels: the original image, the HDR-enhanced, and the S/H-enhanced image.
    Augmentations include flips, Gaussian noise, sharpening filters, rotations of up to 12$^{\circ}$, as well as crops and rescaling.
    A polyline extraction via a minimum spanning tree is performed to extract the final calving front prediction from CALFIN's output probability masks, and the result is masked with the corresponding fjord boundaries.
    To assess their prediction quality, Cheng et\,al.~\cite{Cheng.2021} calculate the mean–median of the distances between the closest pixels in the predicted and target fronts in meters.

    For the comparison, we omit the resizing step during post-processing because information is lost during resizing. 
    Instead, we directly extract patches of size 224\,$\times$\,224 pixels.
    We adjust the number of output layers from two to five so that one channel predicts the front labels of the \ac{caffe} dataset, and the remaining four channels predict the zone labels.
    Masking of fjord boundaries during post-processing is prior knowledge and, therefore, cannot be used for the comparison.
    The polyline extraction without masking the fjord boundaries would result in a coastline prediction, not a calving front prediction.
    Therefore, to extract the final front prediction, we use the post-processing of Gourmelon et\,al.~\cite{Gourmelon_2022}'s Zones network instead of the original post-processing.

\subsection*{\texorpdfstring{Davari et\,al.~\cite{Davari_Baller.2021}}{}}
    \noindent Davari et\,al.~\cite{Davari_Baller.2021} convert the typical binary front segmentation to a regression problem by applying a distance map transform on the calving front segmentation mask.
    Their dataset includes \ac{SAR} imagery of two glacier systems at the Antarctic Peninsula. 
    The images are multi-looked, calibrated to sigma-0, geo-referenced, ortho-rectified, and resized to 512$\,\times\,$512 pixels. 
    Flips and rotations are used to augment the dataset.
    Their \ac{dl} model predicts each pixel's distance to the closest point of the calving front.
    The architecture of their model is a simple U-Net \cite{Ronneberger.2015}, with the mean-squared error loss for training.
    From the predicted distance map, the calving front is extracted during post-processing.
    In their paper, Davari et\,al.~\cite{Davari_Baller.2021} test three different post-processing schemes: statistical thresholding, a conditional random field, and a second U-Net.
    Davari et\,al.~\cite{Davari_Baller.2021} showed that the second U-Net gives the most accurate results.
    The U-Net takes the predicted distance map as input and outputs a segmentation prediction for the front.
    The model is trained on the front segmentation masks using a binary cross-entropy loss.
    The one-pixel-wide front segmentation masks are thickened with a kernel of size $5\,\times\,5$ to ease the class imbalance problem for the second U-Net.
    The output of the second U-Net is post-processed with morphological thinning.

    We add the post-processing of Gourmelon et\,al.~\cite{Gourmelon_2022}'s Front network.
    This results in a one-pixel-wide prediction for the front, which is essential for a fair comparison, as with a broader prediction, the distance error computation would be skewed.
    
\subsection*{\texorpdfstring{Davari et\,al.~\cite{Davari_Islam.2021}}{}}
    \noindent Davari et\,al.~\cite{Davari_Islam.2021} tested three different versions of the U-Net \cite{Ronneberger.2015} for calving front extraction.
    The best-performing version uses \ac{mcc} as an early stopping criterion and is trained on binary segmentation masks showing the calving front versus background using an improved distance map loss.
    Davari et\,al.~\cite{Davari_Islam.2021}'s dataset includes \ac{SAR} imagery of the Jakobshavn Glacier located in Greenland and two glacier systems at the Antarctic Peninsula. 
    The images are multi-looked, calibrated to sigma-0, geo-referenced, and ortho-rectified. 
    Additionally, only the images of Jakobshavn are median-filtered to reduce speckle noise.
    The calving front segmentation masks are dilated with a $5\,\times\,5$ kernel to alleviate the class imbalance.
    All images are divided into non-overlapping patches of size $256\,\times\,256$, and the resulting dataset is artificially enlarged by flip and rotation augmentations.
    No post-processing is performed.

    For the comparison, we omit median-filtering as all images need to be treated similarly, and we add the post-processing of Gourmelon et\,al.~\cite{Gourmelon_2022}'s Front network to extract the calving front.

\subsection*{\texorpdfstring{Gourmelon et\,al.~\cite{Gourmelon_2022}}{}}
    \noindent The baselines for the benchmark dataset were presented in the same paper as the dataset by Gourmelon et\,al.~\cite{Gourmelon_2022}.
    As the benchmark features two label categories, Gourmelon et\,al.~\cite{Gourmelon_2022} provide two separate networks, which from now on will be called ``Zones'' and ``Front'' after the labels used to train the networks. 
    Both networks have a U-Net structure with \ac{aspp}~\cite{Chen.2018} in the bottleneck.
    The Front network is trained with an improved distance map loss~\cite{Davari_Islam.2021}, while the loss function of the Zones network is a weighted combination of Dice~\cite{Sudre.2017} and cross-entropy~\cite{Bishop.1995}. 
    Only the front labels are pre-processed via a morphological dilation employing a rectangular structuring element of size $5\,\times\,5$ pixels.
    For further processing, both networks extract patches of size $256\,\times\,256$ with no overlap for training and 128 pixels overlap for testing.
    Image patches are augmented online by rotations, horizontal flips, brightness adjustments, Gaussian noise, and elastic transforms.
    Neural network outputs are combined by patch merging with Gaussian importance weighting. 
    Post-processing for the Zones network includes filling gaps in the ocean zone prediction and removing all but the largest connected predicted ocean zone. 
    The boundary between the ocean and glacier zones is taken as the predicted calving front.
    For the Front network, the predicted front is skeletonized, and the longest path in each separate skeleton is identified to obtain 1-pixel-wide lines.
    
    No adaptations were performed for the comparison.

\subsection*{\texorpdfstring{Gourmelon et\,al.~\cite{Gourmelon.2023}}{}}
    \noindent Gourmelon et\,al.~\cite{Gourmelon.2023} change the post-processing of Gourmelon et\,al.~\cite{Gourmelon_2022}'s Zones network by introducing a \ac{crf}.
    The \ac{crf} is the replacement for the commonly used argmax, which determines the predicted zone for each pixel based on the output logits of the network.
    Instead of just considering each pixel individually as argmax does, the \ac{crf} optimizes the predicted zones while considering the predictions and logits of all other pixels.
    
    No further adaptations were made to the system pipeline.
    Moreover, retraining the network was not necessary.

\subsection*{\texorpdfstring{Hartmann et\,al.~\cite{Hartmann.2021}}{}}
    \noindent To increase the accuracy in uncertain image regions, Hartmann et\,al.~\cite{Hartmann.2021} simulate two Bayesian U-Nets with random sampling layers using dropout and concatenate the two networks, generating a two-stage pipeline that determines uncertain regions and then focuses on these regions to enhance the prediction.
    Hartmann et\,al.~\cite{Hartmann.2021}'s multi-looked, geo-referenced, and ortho-rectified dataset comprises \ac{SAR} imagery of two glacier systems at the Antarctic Peninsula.
    For training and testing, patch extraction with a patch size of $256\,\times\,256$ and no overlap is conducted.
    The first Bayesian U-Net takes the \ac{SAR} image as input, while the second, in addition to the \ac{SAR} image, receives an uncertainty map, which is computed as the binarized variance of 20 forward passes of the first U-Net.
    Both networks are trained to segment ocean versus non-ocean regions using the binary cross-entropy and early stopping on the validation loss with a patience of 30 epochs.

    For the comparison, we adapt the U-Nets from binary zone segmentation to multi-zone segmentation with four output channels and categorical cross-entropy loss.
    The second U-Net receives four uncertainty maps, one for each zone. 
    To get the final prediction, an argmax is applied to the four output channels of the second U-Net, and the post-processing of Gourmelon et\,al.~\cite{Gourmelon_2022}'s Zones network is applied.
        
\subsection*{\texorpdfstring{Heidler et\,al.~\cite{Heidler.2021}}{}}        
    \noindent Heidler et\,al.~\cite{Heidler.2021}'s network is based on the U-Net architecture with a down-sampling depth of six but has two output heads: one for edge detection of the coastline and one for the segmentation into sea and land. Incorporating the edge detection is inspired by classical coastline delineation approaches.
    Both heads separately merge up-scaled feature maps from the U-Net's decoder using an attention mechanism and employ deep supervision with an adaptively balanced cross-entropy loss function.
    The dataset used to train and test the network includes 16 Antarctic Sentinel-1 scenes taken between June 2017 and December 2018, each covering an area of 315~km\,$\times$\,263~km.
    During pre-processing, all scenes are processed in the Antarctic Polar Stereographic Projection (EPSG:3031), converted to decibels, and divided into overlapping patches of 768\,$\times$\,768 pixels.
    Applied augmentations are rotations with multiples of 90$^{\circ}$ and mirroring both horizontally and vertically.
    As both polarizations of Sentinel-1 are used, the input to the network has two channels (HH and HV).
    Heidler et\,al.~\cite{Heidler.2021} conduct no post-processing of the network's output.
    The sea/land segmentation is evaluated using the mean IoU, and for the edge detection result, both the F1 scores at the optimal image and dataset scale and the average distance to the target coastline over all predicted coastline pixels are employed.
    Moreover, Heidler et\,al.~\cite{Heidler.2021} showed that adding down-sampled Tandem-X elevation maps as a third input channel can be beneficial.

    The input channels are reduced to one to accommodate for the \ac{caffe} dataset.
    In addition, the sea and land segmentation network head is extended to encompass multiple landscape zones.
    The loss function for this multi-class head is set to categorical cross-entropy, and the post-processing of Gourmelon et\,al.~\cite{Gourmelon_2022}'s Zones network is applied to extract the calving front.
    The coastline segmentation head did not need any adaptation to be used for calving front segmentation.
    Only the post-processing of Gourmelon et\,al.~\cite{Gourmelon_2022}'s Front network is used to obtain a calving front prediction from the binary segmentation head.

\subsection*{\texorpdfstring{Herrmann et\,al.~\cite{Herrmann.2023}}{}}
    \noindent The nnU-Net~\cite{Isensee.2021}, a framework initially designed for biomedical image segmentation, adapts the U-Net to a given dataset and automates design decisions and hyperparameter tuning, eliminating the need for manual intervention. In addition, the nnU-Net uses deep supervision. 
    Herrmann et\,al.~\cite{Herrmann.2023} train and test the nnU-Net, fixed to a down-sampling depth of eight, on the \ac{caffe} dataset and experiment with multi-task learning, concluding that fusing the front and zone label and training the nnU-Net with this fused label yields the lowest \ac{mde}.
    Front labels are dilated with a structuring element of $5\,\times\,5$ pixels and inserted into the zone label.
    Patch extraction is performed with the median image size, which for the \ac{caffe} dataset is $1280\,\times\,1024$.
    The dataset is augmented online using rotations and scaling, Gaussian noise, Gaussian blur, brightness and contrast adjustments, simulation of low resolution, gamma augmentation, and mirroring.
    nnU-Net's loss function is a combination of cross-entropy and Dice score.
    Since the nnU-Net assumes that the final segmentation objective is the label itself, Herrmann et\,al.~\cite{Herrmann.2023} add additional post-processing to extract the calving front.
    For this purpose, the front zone in the fused label is assigned to the ocean zone, and the glacier zone is dilated with a structuring element of $7\,\times\,7$ pixels. 
    Afterward, the post-processing of Gourmelon et\,al.~\cite{Gourmelon_2022}'s Zones network is applied.

    The nnU-Net usually employs five-fold cross-validation and takes the ensemble of the five trained networks as the final prediction. 
    Instead of taking the ensemble, we treat the cross-validation networks as the five training runs and compute the mean and standard deviation of \ac{mde} for our comparison over the five cross-validation networks.        

\subsection*{\texorpdfstring{Holzmann et\,al.~\cite{Holzmann.2021}}{}}
    \noindent Holzmann et\,al.~\cite{Holzmann.2021} introduce attention gates into the skip connections of the U-Net and train the U-Net on labels distinguishing front and background using a distance-weighted loss function.
    Their dataset consists of \ac{SAR} imagery showing two glacier systems in the Antarctic Peninsula.
    For pre-processing, Holzmann et\,al.~\cite{Holzmann.2021} apply a median filter on the \ac{SAR} images and resize both labels and images to $512\,\times\,512$ pixels.
    The front labels are dilated to a width of six pixels to ease the class imbalance.
    Flipping and rotation augmentations are applied to enlarge the dataset.
    During post-processing, the output of the U-Net is simply binarized.

    As we need a one-pixel-wide calving front to calculate the \ac{mde}, we add skeletonization after the binarization.

\subsection*{\texorpdfstring{Kirillov et\,al.~\cite{Kirillov.2023}}{}}
    \noindent The recently introduced \ac{SAM} is a promptable foundation model for zero-shot image segmentation.
    We test the version pre-trained on the SA-1B dataset and ViT-H as the backbone in a zero-shot way on \ac{caffe}; i.\,e., we do not fine-tune \ac{SAM} on \ac{caffe}, but simply use \ac{SAM} as is.
    As \ac{SAM} is trained using RGB images, we repeat our single-channel input three times to artificially create three input channels, as suggested by the authors.
    The images are rescaled for \ac{SAM}'s image encoder - a \ac{ViT} - so that the longest image side has $1024$ pixels and the aspect ratio is preserved.
    The resulting image embeddings are fed into the mask decoder alongside prompts specifying the object to be segmented and an optional segmentation mask that can be used for refinement.
    \ac{SAM} can take prompts in text, point, dense (i.\,e., coarse segmentation map), and bounding box form.
    We generate point prompts using the Contextual HookFormer~\cite{Wu.2023_2} with the goal of enhancing the zone segmentations already created by the Contextual HookFormer.
    For each zone, a sigmoid is applied to the corresponding output channel to receive probability maps. 
    Next, these probability maps are thresholded such that only areas with the highest probability remain. 
    Then, the high probability maps are additionally eroded to focus on points in the center of the specific zone.         
    As \ac{SAM} is not able to conduct semantic segmentation, we focus on predicting the ocean zone.
    Hence, positive prompts are randomly drawn from the eroded high-probability ocean map.
    Negative prompts are randomly drawn from the three remaining eroded high-probability maps.
    We tested two approaches to feed prompts to \ac{SAM}: iteratively and parallel.
    For parallel prompt feeding, we draw ten positive prompts and ten negative prompts per zone (rock, glacier, \ac{NA}) and pass all prompts to \ac{SAM} at once such that \ac{SAM}'s mask decoder is just run once.
    Additionally, we use the Contextual HookFormer's logits of the ocean channel as a dense prompt.
    For iterative prompt feeding, the point prompts are drawn in the same way, but instead of being handed to \ac{SAM} altogether, the prompts are fed into \ac{SAM} one after another.
    \ac{SAM}'s mask decoder is run after every new prompt, receiving the new prompt and the last segmentation output as a dense prompt.
    Like this, the segmentation masks are iteratively enhanced.
    To extract the calving front from the segmentation mask, we overlay the binary ocean mask with the rock and \ac{NA} predictions from the Contextual HookFormer and add the post-processing of Gourmelon et\,al.~\cite{Gourmelon_2022}'s Zones network.

\subsection*{\texorpdfstring{Loebel et\,al.~\cite{Loebel.2022}}{}}
    \noindent Loebel et\,al.~\cite{Loebel.2022} analyze the effect of different inputs on a neural network, including multi-spectral, topographic, and textural inputs.
    For this purpose, Loebel et\,al.~\cite{Loebel.2022} train a U-Net with six down- and upsampling layers on binary labels distinguishing ocean and non-ocean areas.
    The employed loss function is the binary cross-entropy.
    Loebel et\,al.~\cite{Loebel.2022}'s dataset includes radiometrically calibrated and ortho-rectified level-1 Landsat-8 imagery of 23 Greenland outlet glaciers and two glaciers at the Antarctic Peninsula.
    Each glacier is either covered by one $512\,\times\,512$ image or multiple overlapping $512\,\times\,512$ images if the area is too large for a single image.
    During pre-processing, histogram clipping is performed for each multi-spectral band.
    The dataset is augmented eight-fold by rotations and flipping.
    During post-processing, images of the same glacier are merged, if necessary, by averaging the overlap. 
    Next, the coastline is binarized and vectorized using a contour algorithm.
    The calving front is extracted from the coastline with a static mask, which is manually created for each glacier.

    For the comparison, we stick to only \ac{SAR} images as input and alter the U-Net to perform multi-class instead of binary segmentation.
    To do this, we change the number of output channels to four and replace the binary cross-entropy with the categorical cross-entropy.
    Additionally, we employ the post-processing of Gourmelon et\,al.~\cite{Gourmelon_2022}'s Zones network to extract the final calving front.

\subsection*{\texorpdfstring{Marochov et\,al.~\cite{Marochov.2021}}{}}
    \noindent A different approach to front delineation is taken by Marochov et\,al.~\cite{Marochov.2021}.
    Instead of segmenting the entire images directly into the desired classes, Marochov et\,al.~\cite{Marochov.2021} use classification networks to determine the class of each single pixel in each image separately.
    The differentiated classes include open water, iceberg water, mélange, glacier ice, snow on ice, snow on rock, and bare bedrock.
    The employed dataset comprises Sentinel-2 images from three glaciers in Greenland.
    The paper's approach is separated into two phases:
    In the first phase, a VGG16 network \cite{Simonyan_2015_ICLR} is trained on image tiles with $50\,\times\,50$ pixels, in which more than 95~\% of pixels have the same class. 
    The tiles are augmented using rotation and uniformly distributed noise.
    Hence, the input is an image tile, and the output is the predominant class in this image tile.
    With this first phase, the authors aim to overcome the need to produce pixel-wise labels for training, as the training labels for the VGG16 network can be coarse polygons, and the trained VGG16 network can then generate the pixel-wise labels for the second phase by classifying each pixel in the given training images.
    In the second phase, a small \ac{CNN} takes in a small image patch of $15\,\times\,15$ pixels and is trained to predict the center pixel's class.
    Both networks employ the categorical cross-entropy loss function.
    After training, the small \ac{CNN} is then used to classify each pixel in the test images.
    The calving front is extracted during post-processing.
    The largest glacier object is isolated and refined with morphologic geodesic active contours, and the boundary pixels of this glacier object are extracted.
    The classes associated with the ocean (open water, mélange, icebergs) are taken together and objects larger than \SI{1}{\km\squared} are dilated by 30~pixels. 
    The intersection of these ocean objects and the extracted glacier boundary gives the front prediction.
    Moreover, Marochov et\,al.~\cite{Marochov.2021} fine-tuned the trained model on one image from each of the glaciers in the test set.
    These images are not taken from the test set directly but from the glaciers in the test set at a time point, which is not included in the test set.

    For a fair comparison, we omit the fine-tuning. 
    We adapt the networks to predict the four classes prevalent in \ac{caffe}'s zone labels.
    To counter class imbalance, we did not perform augmentations for glacier tiles, as glacier tiles occur much more frequently in the training set than the other three classes.
    Moreover, as the prominent feature of the \ac{NA} class is a smooth black region, we do not add Gaussian noise to the tiles of this class.
    In the original code of phase 1, training is stopped when a validation accuracy of 0.985 is reached.
    We change this to early stopping when the change of validation accuracy is less than 0.005 with a patience of 10 epochs, as a validation accuracy of 0.985 is never reached for the \ac{caffe} dataset.

\subsection*{\texorpdfstring{Mohajerani et\,al.~\cite{Mohajerani.2019}}{}}
    \noindent Mohajerani et\,al.~\cite{Mohajerani.2019} employ a U-Net with a weighted binary cross-entropy as a loss function to segment multi-spectral Landsat images into calving front and background.
    Their data comprises 123 images of four Greenlandic glaciers.
    During pre-processing, these images are cropped to the region around the front with a buffer of 300\,m, rotated such that the front is oriented in the y-direction, and resized to 200\,$\times$\,300 pixels using cubic interpolation.
    For training, the resulting 200\,$\times$\,300 sized images are cropped to a size of 150\,$\times$\,240 pixels.
    Moreover, Mohajerani et\,al.~\cite{Mohajerani.2019} normalize the image contrast, equalize grey-scale intensities to create a uniform distribution, and apply smoothing and edge enhancement kernels.
    As augmentation, the images are additionally flipped horizontally, and grey-scale intensities are inverted.
    The U-Net produces a probability mask, which must be post-processed to attain the final calving front prediction.
    The post-processing entails computing the least-cost path through the probability mask, with the values of the probability mask as step weights.

    Since rotating the images so that the front is oriented in a certain way also requires prior knowledge of the test set, this part is omitted for the comparison.
    We replaced the multiple cropping and resizing steps in pre-processing by rescaling the images to the average bounding box size, as resizing to the average of the entire images resulted in a memory error. This procedure gave better validation results than cropping to the bounding box size and then resizing to 150\,$\times$\,240 pixels.
    During testing, we omit the rescaling altogether.
    Further pre-processing steps are kept unchanged.
    The labels used are the front labels of \ac{caffe}.
    Hence, no architecture or loss function changes were needed.
    The post-processing was exchanged with that of Gourmelon et\,al.~\cite{Gourmelon_2022}'s Front network since the original is based on knowledge of the fjord boundaries, which we consider prior knowledge.
    
\subsection*{\texorpdfstring{Periyasamy et\,al.~\cite{Periyasamy.2022}}{}}
    \noindent Periyasamy et\,al.~\cite{Periyasamy.2022} aim to find an optimal configuration for a U-Net trained to differentiate between ocean and non-ocean regions by optimizing data pre-processing, data augmentation, the loss function, normalization layer, dropout rate, bottleneck layer, and transfer learning.
    Their dataset consists of multi-looked, geo-referenced, ortho-rectified SAR imagery of two glaciers in the Antarctic Peninsula and one glacier in Greenland.
    The best-performing model takes inputs pre-processed with a bilateral and a CLAHE filter. This denoising during preprocessing was originally inspired by classical methods for calving front delineation.
    For training, images are divided into non-overlapping patches of size $256\,\times\,256$ pixels and augmented eight-fold by rotation and flipping.
    During inference, images are fed into the network as a whole.
    The bottleneck of the best-performing U-Net includes a residual connection and dilated convolutions.
    The loss function combines the binary cross-entropy and the Dice loss with equal weighting.
    During post-processing, the calving front is extracted by dropping all but the largest connected ocean component and applying the canny edge detector to receive the contour of the ocean.
    
    For the comparison, we employ the optimized U-Net and alter the binary zone segmentation to a multi-zone segmentation.
    Hence, binary cross-entropy is replaced with a categorical cross-entropy.
    Moreover, we use a softmax as the final activation layer instead of a sigmoid and employ an argmax instead of a simple threshold to receive the zone predictions.
    Lastly, we replace the post-processing with the post-processing of Gourmelon et\,al.~\cite{Gourmelon_2022}'s Zones network.

\subsection*{\texorpdfstring{Wu et\,al.~\cite{Wu.2023_1}}{}}
    \noindent In all systems designed for calving front extraction, images are either divided into patches or resized to alleviate GPU memory issues. 
    Both resizing and patch extraction have their downsides: During resizing, high-frequency details are lost, while patches miss the global information around the patch.
    Wu et\,al.~\cite{Wu.2023_1} address this trade-off by employing the HookNet~\cite{vanRijthoven.2021}.
    The HookNet consists of two connected U-Nets, each with a down-sampling depth of four. The first U-Net takes in the target patch, while the other receives a downsized patch of the context that covers both the target patch and the surrounding area.
    Therefore, this approach combines local high-frequency details and coarse global information in the input.
    Wu et\,al.~\cite{Wu.2023_1} improve the HookNet by integrating an attention mechanism into multihooking U-Nets with deep supervision of the feature pyramid in the architecture. 
    The improved network is called \ac{AMD-HookNet}.
    The zone labels of the \ac{caffe} dataset are employed for training and testing.
    Wu et\,al.~\cite{Wu.2023_1} extract non-overlapping target patches with a size of $288\,\times\,288$ pixels.
    The extracted context patches are of size $567\,\times\,567$ pixels, with the corresponding target patch in the center. 
    The context patches overlap by 288 pixels and are resized to a size of $288\,\times\,288$ pixels before being fed into the U-Net of the AMD-HookNet's context branch.
    The patches are jointly augmented via rotations and flipping.
    Wu et\,al.~\cite{Wu.2023_1}'s AMD-HookNet is trained with a combination of the categorical cross-entropy and Dice loss of the target branch's and context branch's output as well as deep supervision of upsampled feature maps of the hooking mechanism.
    The output patches of the target branch are stitched together and post-processed to extract the calving front using the post-processing of Gourmelon et\,al.~\cite{Gourmelon_2022}'s Zones network.
    
    No adaptation except the length of training had to be performed for the comparison.

\subsection*{\texorpdfstring{Wu et\,al.~\cite{Wu.2023_2}}{}}
    \noindent The HookFormer is the first fully Transformer-based network for calving front extraction.
    Wu et\,al.~\cite{Wu.2023_2} base the HookFormer on the AMD-HookNet but exchange the convolution blocks with Swin Transformer blocks~\cite{Liu.2021} and improve the hooking mechanism by introducing a Cross-Attention Swin-Transformer module and a Cross-Interaction module.
    The dataset and labels are the \ac{caffe} dataset and its zone labels, the same as for the AMD-HookNet.
    The target patch size is $224\,\times\,224$, while the context patch size is $488\,\times\,488$, which is rescaled to $224\,\times\,224$ as well.
    Context patches are extracted with an overlap of $224$ pixels, while the predictions of non-overlapping target patches are used as network outputs.
    All patches are augmented by rotation and flipping.
    During training, a combination of categorical cross-entropy and Dice loss is used to supervise target and context branch outputs and the upsampled target bottleneck map.
    To attain the final calving front prediction, the post-processing of Gourmelon et\,al.~\cite{Gourmelon_2022}'s Zones network is applied. 

    For the comparison, no adaptations were necessary.

\subsection*{\texorpdfstring{Zhang et\,al.~\cite{Zhang.2021}}{}}
    \noindent Zhang et\,al.~\cite{Zhang.2021} replace the U-Net with the DeepLabv3~\cite{Chen_2018_ECCV} architecture to segment optical and \ac{SAR} imagery into land and sea.
    Their dataset, with corresponding manual delineations, is published by Zhang et\,al.~\cite{Zhang.2020_dataset}.
    As pre-processing, the images are cropped to the region of interest, de-speckled, and their histograms normalized.
    Before rotation and flipping augmentations are applied, the images are subdivided into patches of 960\,$\times$\,720 pixels.
    Zhang et\,al.~\cite{Zhang.2021} perform a comparison between the U-Net and the DeepLabv3+ with different backbones. 
    The tested backbones include ResNet \cite{He_2016_CVPR}, DRN \cite{Yu_2017_CVPR}, and MobileNet \cite{Howard_2017_MobileNets}.
    Their post-processing is the same as of Zhang et\,al.~\cite{Zhang.2019}, except that a final step is added where fronts with too complex shapes are omitted based on their frequency, amplitude of vibration, and convexity of the polygon. 
    The performance metric, the mean difference, is likewise taken from Zhang et\,al.~\cite{Zhang.2019}.

    For the comparison, the mentioned pre-processing steps are omitted, as these have already been performed for the \ac{caffe} dataset.
    The binary segmentation is altered to a multi-class segmentation to accommodate for \ac{caffe}'s zone labels.
    For this purpose, the output channels have been increased to four, and the categorical cross-entropy loss instead of the binary cross-entropy loss has been applied.
    Moreover, the post-processing of Gourmelon et\,al.~\cite{Gourmelon_2022}'s Zones network is integrated, which includes an argmax instead of a threshold to receive the prediction.
    The original post-processing could not be applied because, first, it assumes the existence of only two classes, and second, it would require prior knowledge of the test set.    
    
\subsection*{\texorpdfstring{Zhang et\,al.~\cite{Zhang.2023}}{}}
    \noindent A complete calving front delineation pipeline for \ac{gee} is presented by Zhang et\,al.~\cite{Zhang.2023}.
    The automated pipeline includes a screening module for erroneous predictions as well as an uncertainty estimation.
    To train the included DeepLabv3+~\cite{Chen_2018_ECCV}, Zhang et\,al.~\cite{Zhang.2023} curated the TermPicks dataset~\cite{Goliber.2022} and added additional manually annotated fronts summing up to 17,906 samples from 249 glaciers in Greenland.
    Only satellites available on \ac{gee} are included.
    Before the images are fed into the model, a cloud screening is performed to ensure the calving front is visible.
    Next, histogram equalization is conducted and images with a width of less than 1000 pixels are resized to a width that is just larger than 1000 pixels.
    Patches of size $960\,\times\,720$ are extracted with an overlap of 320, 240 (width, height) for training and 384, 288 for testing.
    Using flipping and rotation, Zhang et\,al.~\cite{Zhang.2023} enlarge the dataset artificially.
    The model learns to differentiate ocean from non-ocean using a binary cross-entropy loss.
    During post-processing, patches are merged by averaging the overlap, and the values are thresholded with 0.5.
    To extract the calving front, the prediction is converted to a polygon; small polygons and the image border are removed, leaving the predicted calving front.
    Lastly, the predicted calving fronts undergo a screening to remove erroneous fronts.
    The screening checks the calving front curvature and length, the number of intersections between glacier flowlines and front, and the size of enclosed areas between temporally adjacent calving fronts.
    
    During the comparison, we omitted the cloud screening and the histogram normalization, as \ac{SAR} penetrates cloud cover, and histogram normalization was already performed on the benchmark dataset.
    We changed the output channels of DeepLabv3+ to four, trained the network using the categorical cross-entropy, and used an argmax instead of a threshold for binarization to accommodate for \ac{caffe}'s zone labels.
    Moreover, the screening module could not be applied, as three of the four checks are based on thresholds that can only be calculated using optical imagery, and the last check relies on glacier flowlines, which would be prior knowledge of the test set.
    
\subsection*{\texorpdfstring{Zhu et\,al.~\cite{Zhu.2023}}{}}
    \noindent Zhu et\,al.~\cite{Zhu.2023} leverage the properties of both \acp{CNN} and \acp{ViT} by incorporating \acp{gla-st} into DeepLabv3+.
    They dub the resulting model \ac{gla} and train it with a weighted combination of the binary cross-entropy loss and the Dice loss on the final output and the binary cross-entropy loss on an auxiliary output.
    Experiments to assess the model's performance are based on the \ac{caffe} dataset.
    Zhu et\,al.~\cite{Zhu.2023} fuse all classes but the ocean class in the zone labels, leading to a binary ocean segmentation.
    Consistent with Swin-L~\cite{Liu.2021}, patches of size $384\,\times\,384$ are extracted with an overlap of 50\,\% for both training and testing.
    The only augmentation performed during the model's training is random horizontal flipping.
    The conducted post-processing that is needed to calculate the \ac{mde} is not described in the study nor published with the code.

    To enable the calculation of the \ac{mde}, the network is adjusted to predict all zones provided by the \ac{caffe} dataset, the binary cross-entropy loss terms in the combined loss function are exchanged with the categorical cross-entropy loss and the post-processing of Gourmelon et\,al.~\cite{Gourmelon_2022}'s Zones network is applied.

\subsection{Statistical Analysis}
\label{app:statistical_analysis_methods}
\noindent To check whether the resulting differences between the \ac{dl} systems are significant, a Kruskal-Wallis test~\cite{Kruskal_1952} is conducted.
The best-performing model is compared with the second, third, and fourth best-performing models to check whether the performance gain is significant.
For this purpose, six one-sided Mann-Whitney U-tests are performed. 
Three of them are based on the models' \acp{mde}, and three are based on the models' number of images with no predicted front.
Subsequently, the results are grouped according to different properties of the \ac{dl} systems to discover whether these properties have an impact on the performance.
To test the hypothesis that a certain group is more suitable, the \acp{mde} of the groups are compared with a Kruskal-Wallis test~\cite{Kruskal_1952}. 
The Kruskal-Wallis test is followed by one-sided Mann-Whitney U-tests that check whether the performance differences between the best group and the remaining ones are significant.
First, the results are grouped by base architecture.
The \ac{dl} systems have a total of four basic architectures on which they are built: the U-Net~\cite{Ronneberger.2015}, DeepLabv3+~\cite{Chen_2018_ECCV}, the \ac{ViT}~\cite{Dosovitskiy.2020}, and VGG16~\cite{Simonyan_2015_ICLR} (see Table~\ref{tab:Masks_and_Models}).
Only one model, the GLA-STDeepLab~\cite{Zhu.2023}, mixes \ac{ViT} and DeepLabv3+, which we then regard as a fifth type of base architecture.
Second, the results are grouped by models trained on \ac{caffe}'s binary front labels, \ac{caffe}'s zone labels, and models trained in a multi-task manner on both labels (see Table~\ref{tab:Masks_and_Models}).
To test the hypothesis that more global-scale semantic information is beneficial for performance, the correlation factors between the \ac{mde} and two variables are calculated.
For the first variable, each model's mean input size in pixels during training is taken as a surrogate of how much information goes into the networks.
For the second variable, the down-sampling depth of used U-net architectures is taken as an estimate of how much local-global information interaction takes place.

For the statistical analysis of the results, all posthoc tests following Kruskal-Wallis tests were carried out hypothesis-driven and are Bonferroni-corrected if applicable.

\subsection{Multi-annotator study}
\label{app:methods_multi_annotator}
Nine annotators participated in our study, which, together with the original annotator of the \ac{caffe} dataset, results in ten annotations for each image in the test set.
The annotators' levels of proficiency in QGIS and knowledge about glaciers are given in the supplementary information (Fig.~\ref{fig:annotators}).
The annotators were asked to delineate the calving fronts in QGIS, following a provided manual.
In addition to the \ac{SAR} images, they were assisted with a catchment for each glacier and one optical image per glacier (not per \ac{SAR} image) for initial orientation.
The resulting shape files were to be post-processed by removing everything within the catchment area. 
However, some annotators also deliberately labeled the rocky coastline, resulting in fragmented, spurious fronts when everything within the catchment was removed.
Therefore, to remove false fronts, we had to buffer all catchments by 120\,m and expand the Columbia Glacier catchments at the coastline between the eastern and western glacier tongues. Any remaining front fragments shorter than half of the minimum front length in the CaFFe dataset (750\,m) were removed. This threshold accounts for the buffered catchments and potential minor variations in annotation, where some annotators may not have delineated calving fronts precisely at the lateral glacier boundaries.
Following this procedure, we receive standard calving front products like the ones commonly provided and used in the community. This allows us to compare the \ac{dl} models to the quality of a standard product.

As no objective ground truth exists due to the subjectivity of labeling, we consider the aggregation of all people as ground truth.
For calculating how much human annotations deviate, we would, however, get a bias if we simply calculated the \ac{mde} between each annotator and the combination of all ten annotators.
Instead, for each annotator, we aggregate the nine remaining annotators and compare the annotator with this combined version.
For the aggregation of the nine annotators, we conduct a majority vote.
To combine the manually labeled calving fronts, the fronts are used together with the catchment areas to create one PNG per annotator showing the ocean area.
For each pixel in the combined image, the number of annotators that assigned that pixel to the ocean area is counted.
If more than or equal to five annotators assign this pixel to the ocean, this pixel is also attributed to the ocean in the combined image.
We shrink the ocean area by morphological erosion and subtract this eroded version from the original combined ocean area to obtain the coastline.
Next, we remove the parts of the coastline that lie within the catchment that was also used for the individual annotations, leaving us with the calving front.
Finally, we delete fronts that are shorter than \SI{750}{\meter} and occur due to rocky coastlines that are labeled as front and are not covered by the buffered catchment area.

\subsection{DL versus humans}
\label{app:dl_vs_humans}
\noindent We use the \ac{dl} system with the lowest \ac{mde} and compare it to human performance.
Since training a neural network is a stochastic process, the \ac{dl} system is trained five times.
As the \ac{caffe} benchmark dataset was labeled by annotator number ten, the \ac{dl} system is trained on annotator number ten, and, therefore, the system might have a bias towards this annotator.
Still, as the \ac{dl} system's outputs are not equal to the annotations of annotator number ten, we compare the \ac{dl} system to the aggregation of all ten annotators.
The combination of the ten annotators is performed in exactly the same way as it is done for the combination of the nine annotators.
To make the comparison between the \ac{dl} system and annotators fairer, the predictions of HookFormer~\cite{Wu.2023_2} are further post-processed just like the multi-annotator annotations (removal of predicted front pixels within the buffered catchment area, deletion of fronts shorter than \SI{750}{\meter}). 
We then calculate the number of images with no predicted front and the \ac{mde} to the combined annotations instead of \ac{caffe}'s ground truth.
To test whether the difference in \ac{mde} between humans and the \ac{dl} system is significant, a one-sided Mann-Whitney U-test is carried out.

\section*{Results}
    \subsection*{Deep Learning system comparison}
    \label{app:numerical_results_ai_comparison}
\begin{figure*}    
            \includegraphics[width=\textwidth]{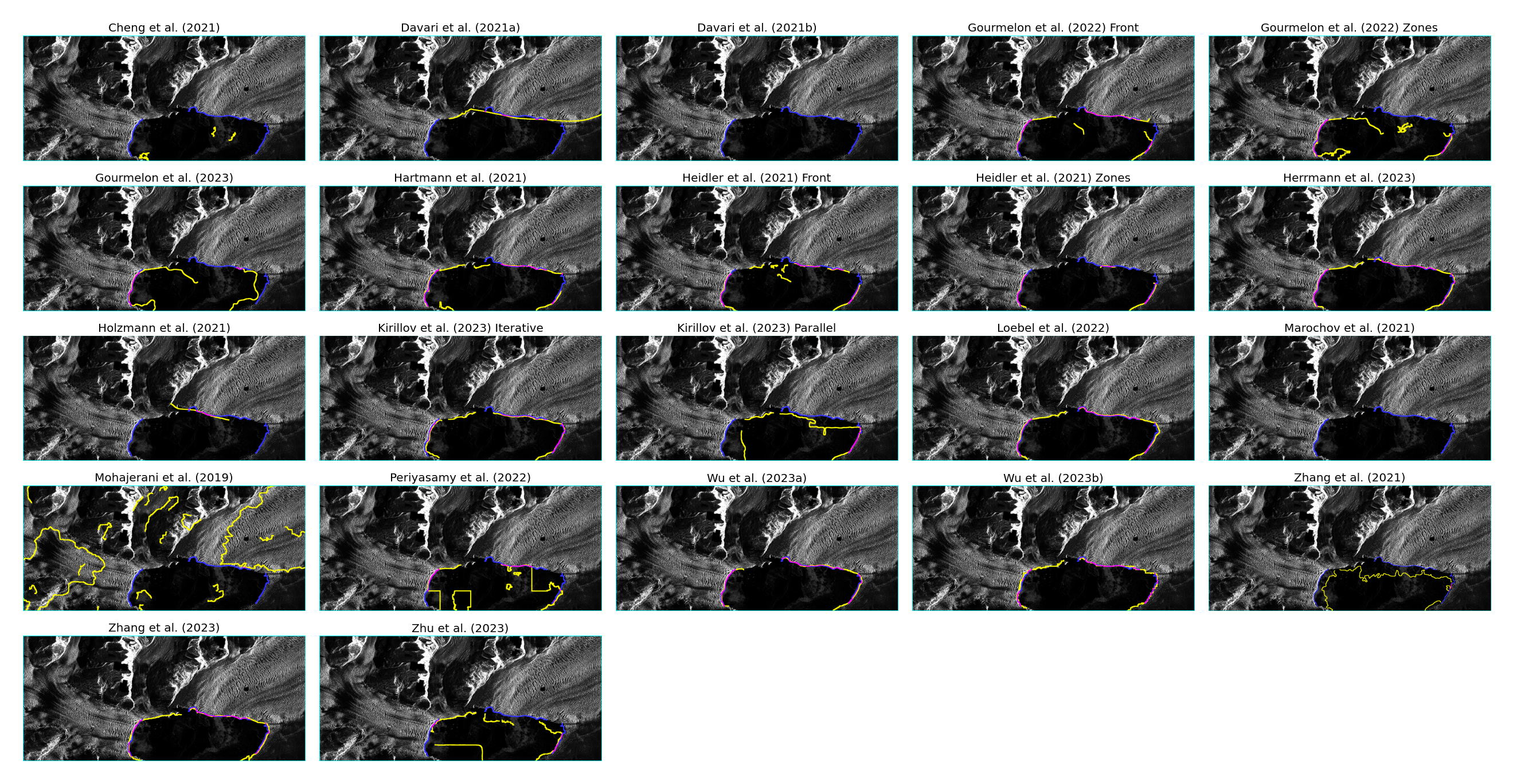}
            \caption{Predicted calving fronts of all 22 \ac{dl} systems for an image of the Columbia Glacier taken on 2nd January 2012 by the TanDEM-X satellite. \fboxsep=1pt\colorbox{yellow!100}{Yellow} depicts the prediction, 
            \fboxsep=1pt\colorbox{blue!100}{\color{white}blue} is used for the ground truth front, and 
            \fboxsep=1pt\colorbox{magenta!100}{\color{white}pink} signifies a perfect match between prediction and ground truth. The bounding box is given in 
            \fboxsep=1pt\colorbox{Turquoise!100}{turquoise}. \ac{SAR} imagery is provided by DLR, ESA, and ASF.}
            \label{fig:all_models_Col}
        \end{figure*}

\begin{figure*}
            \includegraphics[width=\textwidth]{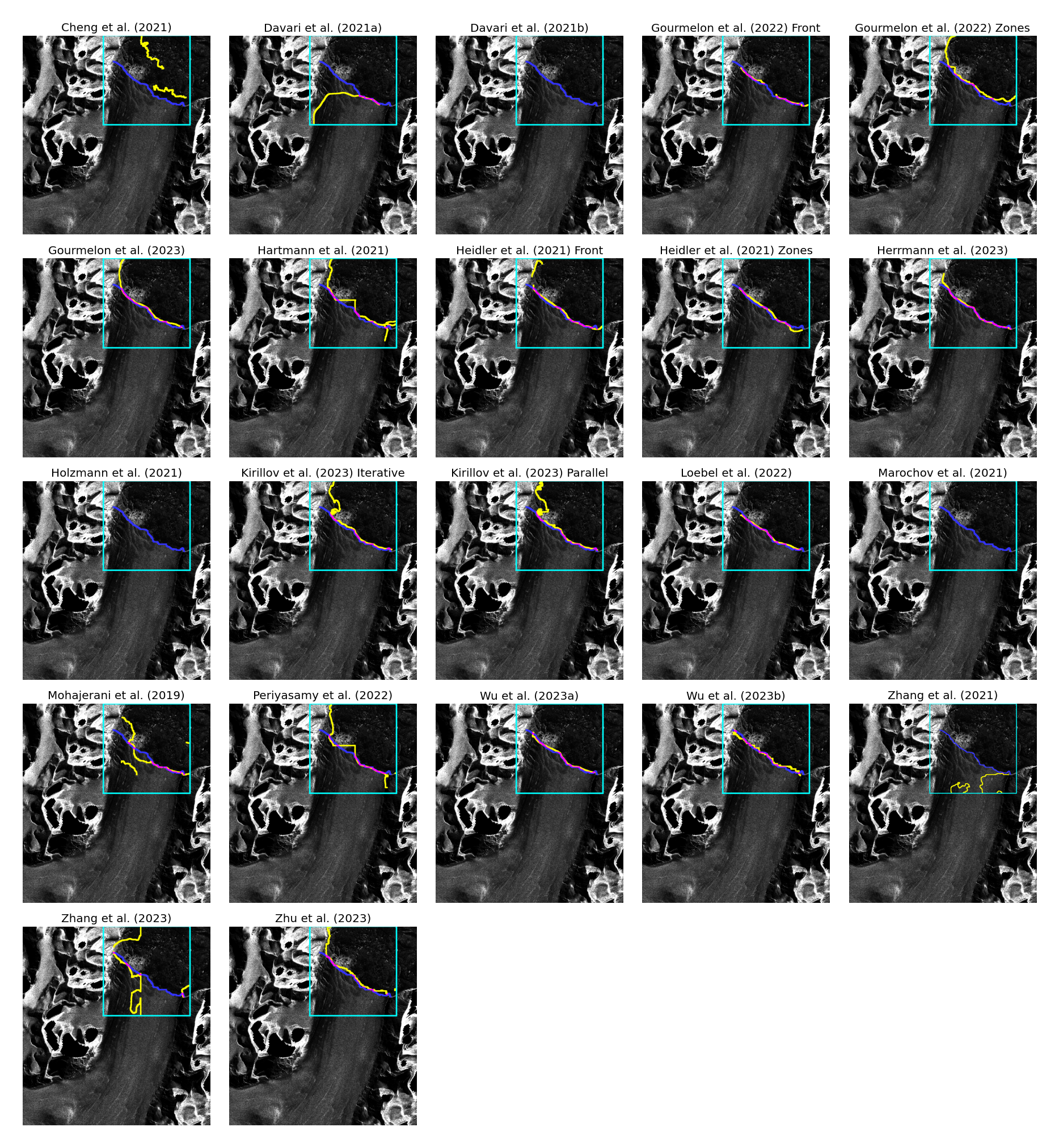}
            \caption{Predicted calving fronts of all 22 \ac{dl} systems for an image of the Mapple Glacier taken on 13th October 2008 by the TerraSAR-X satellite. \fboxsep=1pt\colorbox{yellow!100}{Yellow} depicts the prediction, 
            \fboxsep=1pt\colorbox{blue!100}{\color{white}blue} is used for the ground truth front, and 
            \fboxsep=1pt\colorbox{magenta!100}{\color{white}pink} signifies a perfect match between prediction and ground truth. The bounding box is given in 
            \fboxsep=1pt\colorbox{Turquoise!100}{turquoise}. \ac{SAR} imagery is provided by DLR, ESA, and ASF.}
            \label{fig:all_models_Mapple}
        \end{figure*}

        \noindent This section provides the mean and standard deviations of the \acp{mde} for subsets of the test set for all 22 \ac{dl} systems, a visual examination of the predictions (Figures \ref{fig:all_models_Col} and \ref{fig:all_models_Mapple}), and the numerical results for the statistical analyses.
        Table~\ref{tab:distance_errors} shows the \ac{mde} for the complete test set as well as for only summer and only winter images.
        Table~\ref{tab:distance_errors_mapple} provides \acp{mde} for the Mapple Glacier, encompassing all its images and further categorized into summer and winter sets.
        Similarly, \acp{mde} for the Columbia Glacier are given in Table~\ref{tab:distance_errors_columbia}, with a breakdown into summer and winter images as well as an overall measure for all images of the glacier.
        A breakdown of the test set results into the different sensors is provided in Tables~\ref{tab:distance_errors_sensors_mde} and~\ref{tab:distance_errors_sensors_no_front}.
        
        \begin{table*}[htbp]
            \centering
            \caption{\acp{mde} in meters for \ac{caffe}'s complete test set. $\varnothing$ stands for the number of images for which no front was predicted. A \textbf{bold font} signifies the best value in a column, and a \fboxsep=1pt\colorbox{black!20}{grey background} indicates the best value in a row of both the \ac{mde} and the number of no predicted fronts. Ties lead to multiple marked entries.}
            \label{tab:distance_errors}
            \begin{tabular}{p{0.2\textwidth} p{0.07\textwidth} p{0.09\textwidth} p{0.09\textwidth} p{0.09\textwidth} p{0.09\textwidth} p{0.09\textwidth} p{0.09\textwidth}} 
                \toprule
                & & & & \multicolumn{2}{c}{\textit{Summer}} & \multicolumn{2}{c}{\textit{Winter}}\\
                \textit{Paper} & &  	$\downarrow$ \textit{MDE} &  	$\downarrow$ $\varnothing \in 122$ & $\downarrow$ \textit{MDE} & $\downarrow$ $\varnothing \in 68$ & $\downarrow$ \textit{MDE} & $\downarrow$ $\varnothing \in 54$\\
                \midrule
    
                Cheng et\,al.~\cite{Cheng.2021} && $1767 \pm 536$ & $15 \pm 9$ & \fboxsep=1pt\colorbox{black!20}{$1697 \pm 433$} & $10 \pm 6$ & $1828 \pm 660$ & \fboxsep=1pt\colorbox{black!20}{$5 \pm 4$} \\
                
                Davari et\,al.~\cite{Davari_Baller.2021} & & $2414 \pm 425$ & $27 \pm 5$ & \fboxsep=1pt\colorbox{black!20}{$2205 \pm 288$} & $15 \pm 5$ & $2626 \pm 618$ & \fboxsep=1pt\colorbox{black!20}{$12 \pm 3$} \\
                
                Davari et\,al.~\cite{Davari_Islam.2021} & & $4327 \pm 248$ & $69 \pm 1$ & \fboxsep=1pt\colorbox{black!20}{$4159 \pm 304$} & $43 \pm 0$ & $4523 \pm 273$ & \fboxsep=1pt\colorbox{black!20}{$26 \pm 1$} \\
    
                \multirow{ 2}{*}{Gourmelon et\,al.~\cite{Gourmelon_2022}} & Front & $887 \pm 189$ & $7 \pm 3$ & \fboxsep=1pt\colorbox{black!20}{$738 \pm 111$} & \fboxsep=1pt\colorbox{black!20}{$4 \pm 1$} & $1054 \pm 308$ & $4 \pm 2$ \\
                
                & Zones & $753 \pm 76$ & $1 \pm 1$ & \fboxsep=1pt\colorbox{black!20}{$732 \pm 93$} & $1 \pm 1$ & $776 \pm 65$ & \fboxsep=1pt\colorbox{black!20}{$0 \pm 0$} \\
                
                Gourmelon et\,al.~\cite{Gourmelon.2023} && $726 \pm 76$ & $1 \pm 1$ & \fboxsep=1pt\colorbox{black!20}{$696 \pm 93$} & $1 \pm 1$ & $757 \pm 67$ & \fboxsep=1pt\colorbox{black!20}{$0 \pm 0$} \\
                
                Hartmann et\,al.~\cite{Hartmann.2021} & & $1011 \pm 46$ & $12 \pm 10$ & $1085 \pm 82$ & $8 \pm 5$ & \fboxsep=1pt\colorbox{black!20}{$942 \pm 59$} & \fboxsep=1pt\colorbox{black!20}{$5 \pm 4$} \\
                
                \multirow{ 2}{*}{Heidler et\,al.~\cite{Heidler.2021}} & Front & $499 \pm 31$ & $2 \pm 2$ & \fboxsep=1pt\colorbox{black!20}{$478 \pm 43$} & \fboxsep=1pt\colorbox{black!20}{$1 \pm 1$} & $519 \pm 37$ & \fboxsep=1pt\colorbox{black!20}{$1 \pm 1$} \\
                
                & Zones & $646 \pm 67$ & $6 \pm 5$ & \fboxsep=1pt\colorbox{black!20}{$640 \pm 74$} & \fboxsep=1pt\colorbox{black!20}{$3 \pm 2$} & $648 \pm 95$ & $3 \pm 4$ \\
                
                Herrmann et\,al.~\cite{Herrmann.2023} & & $546 \pm 98$ & $4 \pm 2$ & \fboxsep=1pt\colorbox{black!20}{$459 \pm 121$}& \fboxsep=1pt\colorbox{black!20}{$1 \pm 1$} & $636 \pm 82$ & $2 \pm 1$ \\
                
                Holzmann et\,al.~\cite{Holzmann.2021} & & $2498 \pm 283$ & $77 \pm 4$ & $2587 \pm 314$ & $50 \pm 1$ & \fboxsep=1pt\colorbox{black!20}{$2445 \pm 300$} & \fboxsep=1pt\colorbox{black!20}{$26 \pm 5$} \\
                
                \multirow{ 2}{*}{Kirillov et\,al.~\cite{Kirillov.2023}} & Iterative & $708 \pm 74$ & $13 \pm 2$ & $726 \pm 88$ & \fboxsep=1pt\colorbox{black!20}{$6 \pm 2$} & \fboxsep=1pt\colorbox{black!20}{$688 \pm 109$} & $7 \pm 1$\\
                
                & Parallel & $753 \pm 105$ & $9 \pm 0$ & \fboxsep=1pt\colorbox{black!20}{$576 \pm 79$} & \fboxsep=1pt\colorbox{black!20}{$4 \pm 0$} & $929 \pm 144$ & $5 \pm 0$ \\
                
                Loebel et\,al.~\cite{Loebel.2022} & & $582 \pm 41$ & $7 \pm 2$ & \fboxsep=1pt\colorbox{black!20}{$521 \pm 52$} & $5 \pm 2$ & $645 \pm 38$ & \fboxsep=1pt\colorbox{black!20}{$2 \pm 1$} \\
                
                Marochov et\,al.~\cite{Marochov.2021} & & $2670 \pm 349$ & $97 \pm 2$ & \fboxsep=1pt\colorbox{black!20}{$2279 \pm 290$} & $56 \pm 1$ & $2880 \pm 420$ & \fboxsep=1pt\colorbox{black!20}{$40 \pm 1$} \\
                
                Mohajerani et\,al.~\cite{Mohajerani.2019} & & $1990 \pm 33$ & \fboxsep=1pt\colorbox{black!20}{$\mathbf{0 \pm 0}$} & \fboxsep=1pt\colorbox{black!20}{$1883 \pm 47$} & \fboxsep=1pt\colorbox{black!20}{$\mathbf{0 \pm 0}$} & $2099 \pm 55$ & \fboxsep=1pt\colorbox{black!20}{$\mathbf{0 \pm 0}$} \\
                
                Periyasamy et\,al.~\cite{Periyasamy.2022} & & $1065 \pm 47$ & $12 \pm 4$ & $1144 \pm 55$ & $6 \pm 3$ & \fboxsep=1pt\colorbox{black!20}{$992 \pm 36$} & \fboxsep=1pt\colorbox{black!20}{$6 \pm 1$} \\
                
                Wu et\,al.~\cite{Wu.2023_1} & & $451 \pm 34$ & $4 \pm 1$ & \fboxsep=1pt\colorbox{black!20}{$421 \pm 43$} & $3 \pm 1$ & $482 \pm 41$ & \fboxsep=1pt\colorbox{black!20}{$1 \pm 1$} \\
                
                Wu et\,al.~\cite{Wu.2023_2} & & $\mathbf{360 \pm 13}$ & \fboxsep=1pt\colorbox{black!20}{$\mathbf{0 \pm 0}$} & \fboxsep=1pt\colorbox{black!20}{$\mathbf{333 \pm 13}$} & \fboxsep=1pt\colorbox{black!20}{$\mathbf{0 \pm 0}$} & $\mathbf{389 \pm 21}$ & \fboxsep=1pt\colorbox{black!20}{$\mathbf{0 \pm 0}$} \\
                
                Zhang et\,al.~\cite{Zhang.2021} & & $1297 \pm 273$ & $45 \pm 5$ & $1455 \pm 268$ & $31 \pm 5$ & \fboxsep=1pt\colorbox{black!20}{$1162 \pm 335$} & \fboxsep=1pt\colorbox{black!20}{$15 \pm 3$} \\
                
                Zhang et\,al.~\cite{Zhang.2023} & & $909 \pm 180$ & $12 \pm 5$ & $1047 \pm 233$ & $9 \pm 3$ & \fboxsep=1pt\colorbox{black!20}{$800 \pm 148$} & \fboxsep=1pt\colorbox{black!20}{$3 \pm 2$} \\
                
                Zhu et\,al.~\cite{Zhu.2023} & & $914 \pm 77$ & $26 \pm 3$ & $988 \pm 74$ & $13 \pm 3$ & \fboxsep=1pt\colorbox{black!20}{$838 \pm 101$} & \fboxsep=1pt\colorbox{black!20}{$13 \pm 2$} \\
                
                \bottomrule
                 & & & & & & & \\
            \end{tabular}
        \end{table*}
    
        \begin{table*}[htbp]
            \centering
            \caption{\acp{mde} in meters for images of the Mapple Glacier in \ac{caffe}'s test set. $\varnothing$ stands for the number of images for which no front was predicted. A \textbf{bold font} signifies the best value in a column, and a \fboxsep=1pt\colorbox{black!20}{grey background} indicates the best value in a row of both the \ac{mde} and the number of no predicted fronts. Ties lead to multiple marked entries. / means that no \ac{mde} could be calculated as no front was predicted for all images in the subset.}
            \label{tab:distance_errors_mapple}
            \begin{tabular}{p{0.2\textwidth} p{0.07\textwidth} p{0.09\textwidth} p{0.09\textwidth} p{0.09\textwidth} p{0.09\textwidth} p{0.09\textwidth} p{0.09\textwidth}} 
                \toprule
                & & & & \multicolumn{2}{c}{\textit{Summer}} & \multicolumn{2}{c}{\textit{Winter}}\\
                \textit{Paper} & & $\downarrow$ \textit{MDE} & $\downarrow$ $\varnothing \in 57$ & $\downarrow$ \textit{MDE} & $\downarrow$ $\varnothing \in 40$ & $\downarrow$ \textit{MDE} & $\downarrow$ $\varnothing \in 17$\\
                \midrule
    
                Cheng et\,al.~\cite{Cheng.2021} && $696 \pm 250$ & $4 \pm 2$ & \fboxsep=1pt\colorbox{black!20}{$688 \pm 268$} & $3 \pm 1$ & $705 \pm 213$ & \fboxsep=1pt\colorbox{black!20}{$1 \pm 1$} \\
                
                Davari et\,al.~\cite{Davari_Baller.2021} & & $233 \pm 29$ & $8 \pm 3$ & $251 \pm 38$ & $6 \pm 2$ & \fboxsep=1pt\colorbox{black!20}{$192 \pm 34$} & \fboxsep=1pt\colorbox{black!20}{$2 \pm 2$} \\
                
                Davari et\,al.~\cite{Davari_Islam.2021} & & \fboxsep=1pt\colorbox{black!20}{$2140 \pm 41$} & $56 \pm 1$ & \fboxsep=1pt\colorbox{black!20}{$2140 \pm 41$} & $39 \pm 1$ & \slash & \fboxsep=1pt\colorbox{black!20}{$17 \pm 0$} \\
    
                \multirow{ 2}{*}{Gourmelon et\,al.~\cite{Gourmelon_2022}} & Front & $150 \pm 24$ & $6 \pm 2$ & \fboxsep=1pt\colorbox{black!20}{$140 \pm 26$} & \fboxsep=1pt\colorbox{black!20}{$2 \pm 1$} & $173 \pm 33$ & \fboxsep=1pt\colorbox{black!20}{$2 \pm 1$} \\
                
                & Zones & $287 \pm 48$ & $0 \pm 1$ & \fboxsep=1pt\colorbox{black!20}{$262 \pm 29$} & $0 \pm 1$ & $340 \pm 93$ & \fboxsep=1pt\colorbox{black!20}{$0 \pm 0$} \\
                
                Gourmelon et\,al.~\cite{Gourmelon.2023} && $263 \pm 40$ & $1 \pm 1$ & \fboxsep=1pt\colorbox{black!20}{$241 \pm 20$} & $1 \pm 1$ & $311 \pm 86$ & \fboxsep=1pt\colorbox{black!20}{$0 \pm 0$} \\
                
                Hartmann et\,al.~\cite{Hartmann.2021}  & & $411 \pm 28$ & $1 \pm 1$ & \fboxsep=1pt\colorbox{black!20}{$346 \pm 27$} & $1 \pm 1$ & $546 \pm 45$ & \fboxsep=1pt\colorbox{black!20}{$0 \pm 0$} \\
                
                \multirow{ 2}{*}{Heidler et\,al.~\cite{Heidler.2021}} & Front & $308 \pm 43$ & $2 \pm 2$ & \fboxsep=1pt\colorbox{black!20}{$291 \pm 39$} & \fboxsep=1pt\colorbox{black!20}{$1 \pm 1$} & $346 \pm 61$ & \fboxsep=1pt\colorbox{black!20}{$1 \pm 1$} \\
                
                & Zones & $256 \pm 32$ & $3 \pm 3$ & \fboxsep=1pt\colorbox{black!20}{$225 \pm 17$} & $2 \pm 1$ & $325 \pm 69$ & \fboxsep=1pt\colorbox{black!20}{$1 \pm 1$} \\
                
                Herrmann et\,al.~\cite{Herrmann.2023} & & $\mathbf{107 \pm 8}$ & $1 \pm 1$ & $\mathbf{108 \pm 9}$ & \fboxsep=1pt\colorbox{black!20}{$\mathbf{0 \pm 0}$} & \fboxsep=1pt\colorbox{black!20}{$\mathbf{104 \pm 18}$} & \fboxsep=1pt\colorbox{black!20}{$\mathbf{0 \pm 0}$} \\
                
                Holzmann et\,al.~\cite{Holzmann.2021} & & $609 \pm 348$ & $56 \pm 1$ & \fboxsep=1pt\colorbox{black!20}{$709 \pm 448$} & $39 \pm 1$ & $775 \pm 0$ & \fboxsep=1pt\colorbox{black!20}{$17 \pm 0$} \\
                
                \multirow{ 2}{*}{Kirillov et\,al.~\cite{Kirillov.2023}} & Iterative & $373 \pm 89$ & $7 \pm 2$ & \fboxsep=1pt\colorbox{black!20}{$216 \pm 17$} & \fboxsep=1pt\colorbox{black!20}{$3 \pm 1$} & $658 \pm 213$ & $4 \pm 0$ \\
                
                & Parallel & $219 \pm 20$ & $4 \pm 0$ & \fboxsep=1pt\colorbox{black!20}{$167 \pm 13$} & \fboxsep=1pt\colorbox{black!20}{$1 \pm 0$} & $342 \pm 43$ & $2 \pm 1$ \\
                
                Loebel et\,al.~\cite{Loebel.2022} & & $215 \pm 43$ & $6 \pm 2$ & \fboxsep=1pt\colorbox{black!20}{$195 \pm 27$} & $4 \pm 2$ & $254 \pm 89$ & \fboxsep=1pt\colorbox{black!20}{$2 \pm 2$} \\
                
                Marochov et\,al.~\cite{Marochov.2021} & & $945 \pm 202$ & $48 \pm 1$ & $1011 \pm 182$ & $34 \pm 1$ & \fboxsep=1pt\colorbox{black!20}{$888 \pm 280$} & \fboxsep=1pt\colorbox{black!20}{$15 \pm 1$} \\
                
                Mohajerani et\,al.~\cite{Mohajerani.2019} & & $607 \pm 9$ & \fboxsep=1pt\colorbox{black!20}{$\mathbf{0 \pm 0}$} & \fboxsep=1pt\colorbox{black!20}{$508 \pm 24$} & \fboxsep=1pt\colorbox{black!20}{$\mathbf{0 \pm 0}$} & $822 \pm 58$ & \fboxsep=1pt\colorbox{black!20}{$\mathbf{0 \pm 0}$} \\
                
                Periyasamy et\,al.~\cite{Periyasamy.2022} & & $567 \pm 22$ & $4 \pm 1$ & \fboxsep=1pt\colorbox{black!20}{$439 \pm 27$} & \fboxsep=1pt\colorbox{black!20}{$2 \pm 1$} & $817 \pm 53$ & $3 \pm 1$ \\
                
                Wu et\,al.~\cite{Wu.2023_1} & & $207 \pm 42$ & $3 \pm 1$ & \fboxsep=1pt\colorbox{black!20}{$202 \pm 52$} & \fboxsep=1pt\colorbox{black!20}{$1 \pm 1$} & $217 \pm 37$ & \fboxsep=1pt\colorbox{black!20}{$1 \pm 1$} \\
                
                Wu et\,al.~\cite{Wu.2023_2} & & $184 \pm 19$ & \fboxsep=1pt\colorbox{black!20}{$\mathbf{0 \pm 0}$} & \fboxsep=1pt\colorbox{black!20}{$138 \pm 30$} & \fboxsep=1pt\colorbox{black!20}{$\mathbf{0 \pm 0}$} & $285 \pm 21$ & \fboxsep=1pt\colorbox{black!20}{$\mathbf{0 \pm 0}$} \\
                
                Zhang et\,al.~\cite{Zhang.2021} & & $652 \pm 260$ & $33 \pm 3$ & \fboxsep=1pt\colorbox{black!20}{$626 \pm 224$} & $24 \pm 3$ & $702 \pm 355$ & \fboxsep=1pt\colorbox{black!20}{$9 \pm 1$} \\
                
                Zhang et\,al.~\cite{Zhang.2023} & & $534 \pm 78$ & $5 \pm 3$ & \fboxsep=1pt\colorbox{black!20}{$506 \pm 106$} & \fboxsep=1pt\colorbox{black!20}{$2 \pm 1$} & $603 \pm 100$ & $2 \pm 2$ \\
                
                Zhu et\,al.~\cite{Zhu.2023} & & $466 \pm 10$ & $14 \pm 3$ & \fboxsep=1pt\colorbox{black!20}{$421 \pm 23$} & $9 \pm 3$ & $560 \pm 20$ & \fboxsep=1pt\colorbox{black!20}{$4 \pm 1$} \\
                
                \bottomrule
                 & & & & & & & \\
            \end{tabular}
        \end{table*}
    
        \begin{table*}[htbp]
            \centering
            \caption{\acp{mde} in meters for images of the Columbia Glacier in \ac{caffe}'s test set. $\varnothing$ stands for the number of images for which no front was predicted. A \textbf{bold font} signifies the best value in a column, and a \fboxsep=1pt\colorbox{black!20}{grey background} indicates the best value in a row of both the \ac{mde} and the number of no predicted fronts. Ties lead to multiple marked entries.}
            \label{tab:distance_errors_columbia}
            \begin{tabular}{p{0.2\textwidth} p{0.07\textwidth} p{0.09\textwidth} p{0.09\textwidth} p{0.09\textwidth} p{0.09\textwidth} p{0.09\textwidth} p{0.09\textwidth}} 
                \toprule
                & & & & \multicolumn{2}{c}{\textit{Summer}} & \multicolumn{2}{c}{\textit{Winter}}\\
                \textit{Paper} & & $\downarrow$ \textit{MDE} & $\downarrow$ $\varnothing \in 65$ & $\downarrow$ \textit{MDE} & $\downarrow$ $\varnothing \in 28$ & $\downarrow$ \textit{MDE} & $\downarrow$ $\varnothing \in 37$\\
                \midrule
    
                Cheng et\,al.~\cite{Cheng.2021} && $2375 \pm 884$ & $11 \pm 8$ & $2633 \pm 872$ & $7 \pm 5$ & \fboxsep=1pt\colorbox{black!20}{$2197 \pm 917$} & \fboxsep=1pt\colorbox{black!20}{$4 \pm 3$} \\
                
                Davari et\,al.~\cite{Davari_Baller.2021} & & $3102 \pm 510$ & $19 \pm 5$ & $3170 \pm 413$ & \fboxsep=1pt\colorbox{black!20}{$9 \pm 4$} & \fboxsep=1pt\colorbox{black!20}{$3054 \pm 650$} & $10 \pm 3$ \\
                
                Davari et\,al.~\cite{Davari_Islam.2021} & & $4331 \pm 252$ & $12 \pm 1$ & \fboxsep=1pt\colorbox{black!20}{$4166 \pm 308$} & \fboxsep=1pt\colorbox{black!20}{$3 \pm 1$} & $4523 \pm 273$ & $9 \pm 1$ \\
    
                \multirow{ 2}{*}{Gourmelon et\,al.~\cite{Gourmelon_2022}} & Front & $1032 \pm 227$ & $2 \pm 1$ & \fboxsep=1pt\colorbox{black!20}{$907 \pm 131$} & \fboxsep=1pt\colorbox{black!20}{$\mathbf{0 \pm 0}$} & $1157 \pm 350$ & $2 \pm 1$ \\
                
                & Zones & $840 \pm 84$ & \fboxsep=1pt\colorbox{black!20}{$\mathbf{0 \pm 0}$} & $854 \pm 111$ & \fboxsep=1pt\colorbox{black!20}{$\mathbf{0 \pm 0}$} & \fboxsep=1pt\colorbox{black!20}{$826 \pm 66$} & \fboxsep=1pt\colorbox{black!20}{$\mathbf{0 \pm 0}$} \\
                
                Gourmelon et\,al.~\cite{Gourmelon.2023} && $814 \pm 86$ & \fboxsep=1pt\colorbox{black!20}{$\mathbf{0 \pm 0}$} & $822 \pm 115$ & \fboxsep=1pt\colorbox{black!20}{$\mathbf{0 \pm 0}$} & \fboxsep=1pt\colorbox{black!20}{$807 \pm 71$} & \fboxsep=1pt\colorbox{black!20}{$\mathbf{0 \pm 0}$} \\
                
                Hartmann et\,al.~\cite{Hartmann.2021}  & & $1158 \pm 96$ & $12 \pm 9$ & $1372 \pm 203$ & $7 \pm 5$ & \fboxsep=1pt\colorbox{black!20}{$998 \pm 76$} & \fboxsep=1pt\colorbox{black!20}{$5 \pm 4$} \\
                
                \multirow{ 2}{*}{Heidler et\,al.~\cite{Heidler.2021}} & Front & $536 \pm 38$ & \fboxsep=1pt\colorbox{black!20}{$\mathbf{0 \pm 0}$} & \fboxsep=1pt\colorbox{black!20}{$532 \pm 56$} & \fboxsep=1pt\colorbox{black!20}{$\mathbf{0 \pm 0}$} & $539 \pm 41$ & \fboxsep=1pt\colorbox{black!20}{$\mathbf{0 \pm 0}$} \\
                
                & Zones & $716 \pm 77$ & $3 \pm 3$ & $745 \pm 94$ & \fboxsep=1pt\colorbox{black!20}{$1 \pm 0$} & \fboxsep=1pt\colorbox{black!20}{$684 \pm 102$} & $2\pm 3$ \\
                
                Herrmann et\,al.~\cite{Herrmann.2023} & & $628 \pm 117$ & $3 \pm 2$ & \fboxsep=1pt\colorbox{black!20}{$556 \pm 157$} & \fboxsep=1pt\colorbox{black!20}{$1 \pm 1$} & $693 \pm 91$ & $2 \pm 1$ \\
                
                Holzmann et\,al.~\cite{Holzmann.2021} & & $2510 \pm 277$ & $21 \pm 4$ & $2608 \pm 289$ & $11 \pm 1$ & \fboxsep=1pt\colorbox{black!20}{$2449 \pm 297$} & \fboxsep=1pt\colorbox{black!20}{$10 \pm 5$} \\
                
                \multirow{ 2}{*}{Kirillov et\,al.~\cite{Kirillov.2023}}  & Iterative & $787 \pm 76$ & $5 \pm 1$ & $892 \pm 110$ & \fboxsep=1pt\colorbox{black!20}{$3 \pm 1$} & \fboxsep=1pt\colorbox{black!20}{$690 \pm 112$} & \fboxsep=1pt\colorbox{black!20}{$3 \pm 1$} \\
                
                & Parallel & $860 \pm 128$ & $5 \pm 0$ & \fboxsep=1pt\colorbox{black!20}{$702 \pm 109$} & $3 \pm 0$ & $993 \pm 158$ & \fboxsep=1pt\colorbox{black!20}{$2 \pm 0$} \\
                
                Loebel et\,al.~\cite{Loebel.2022} & & $642 \pm 44$ & $1 \pm 1$ & \fboxsep=1pt\colorbox{black!20}{$598 \pm 60$} & $1 \pm 0$ & $684 \pm 40$ & \fboxsep=1pt\colorbox{black!20}{$\mathbf{0 \pm 0}$} \\
                
                Marochov et\,al.~\cite{Marochov.2021} & & $2855 \pm 346$ & $48 \pm 2$ & \fboxsep=1pt\colorbox{black!20}{$2558 \pm 394$} & \fboxsep=1pt\colorbox{black!20}{$22 \pm 1$} & $2995 \pm 372$ & $26 \pm 1$ \\
                
                Mohajerani et\,al.~\cite{Mohajerani.2019} & & $2155 \pm 55$ & \fboxsep=1pt\colorbox{black!20}{$\mathbf{0 \pm 0}$} & \fboxsep=1pt\colorbox{black!20}{$2118 \pm 80$} & \fboxsep=1pt\colorbox{black!20}{$\mathbf{0 \pm 0}$} & $2191 \pm 65$ & \fboxsep=1pt\colorbox{black!20}{$\mathbf{0 \pm 0}$} \\
                
                Periyasamy et\,al.~\cite{Periyasamy.2022} & & $1155 \pm 52$ & $7 \pm 4$ & $1332 \pm 55$ & $5 \pm 3$ & \fboxsep=1pt\colorbox{black!20}{$1011 \pm 42$} & \fboxsep=1pt\colorbox{black!20}{$3 \pm 2$} \\
                
                Wu et\,al.~\cite{Wu.2023_1} & & $497 \pm 44$ & $1 \pm 1$ & \fboxsep=1pt\colorbox{black!20}{$481 \pm 62$} & $1 \pm 1$ & $511 \pm 45$ & \fboxsep=1pt\colorbox{black!20}{$\mathbf{0 \pm 0}$} \\
                
                Wu et\,al.~\cite{Wu.2023_2} & & $\mathbf{392 \pm 14}$ & \fboxsep=1pt\colorbox{black!20}{$\mathbf{0 \pm 0}$} & \fboxsep=1pt\colorbox{black!20}{$\mathbf{383 \pm 11}$} & \fboxsep=1pt\colorbox{black!20}{$\mathbf{0 \pm 0}$} & $\mathbf{400 \pm 23}$ & \fboxsep=1pt\colorbox{black!20}{$\mathbf{0 \pm 0}$} \\
                
                Zhang et\,al.~\cite{Zhang.2021} & & $1407 \pm 283$ & $13 \pm 3$ & $1681 \pm 306$ & $7 \pm 2$ & \fboxsep=1pt\colorbox{black!20}{$1208 \pm 335$} & \fboxsep=1pt\colorbox{black!20}{$6 \pm 2$} \\
                
                Zhang et\,al.~\cite{Zhang.2023} & & $989 \pm 209$ & $7 \pm 2$ & $1254 \pm 314$ & $7 \pm 1$ & \fboxsep=1pt\colorbox{black!20}{$820 \pm 155$} & \fboxsep=1pt\colorbox{black!20}{$1 \pm 1$} \\
                
                Zhu et\,al.~\cite{Zhu.2023} & & $999 \pm 91$ & $12 \pm 3$ & $1144 \pm 95$ & \fboxsep=1pt\colorbox{black!20}{$3 \pm 1$} & \fboxsep=1pt\colorbox{black!20}{$870 \pm 112$} & $9 \pm 2$ \\
                
                \bottomrule
                 & & & & & & & \\
            \end{tabular}
        \end{table*}

        \begin{table*}[htbp]
            \centering
            \caption{\acp{mde} in meters for \ac{caffe}'s test set divided by capturing sensor. A \textbf{bold font} signifies the best value in a column, and a \fboxsep=1pt\colorbox{black!20}{grey background} indicates the best value in a row. Ties lead to multiple marked entries. / means that no \ac{mde} could be calculated as no front was predicted for all images in the subset.}
            \label{tab:distance_errors_sensors_mde}
            \begin{tabular}{p{0.2\textwidth} p{0.08\textwidth} p{0.11\textwidth} p{0.11\textwidth} p{0.11\textwidth} p{0.11\textwidth} p{0.11\textwidth}} 
                \toprule
                && \textit{Sentinel-1} & \textit{ENVISAT} & \textit{ERS} & \textit{PALSAR} & \textit{TSX} \\
                \textit{Paper} & & $\downarrow$ \textit{MDE} & $\downarrow$ \textit{MDE} & $\downarrow$ \textit{MDE} & $\downarrow$ \textit{MDE} & $\downarrow$ \textit{MDE} \\
                \midrule
    
                Cheng et\,al.~\cite{Cheng.2021} & & $2510 \pm 813$ & $604 \pm 96$ & \fboxsep=1pt\colorbox{black!20}{$466 \pm 165$} & $644 \pm 243$ & $1737 \pm 561$ \\
                
                Davari et\,al.~\cite{Davari_Baller.2021} & & $3549 \pm 112$ & $462 \pm 197$ & $422 \pm 196$ & \fboxsep=1pt\colorbox{black!20}{$258 \pm 31$} & $2336 \pm 523$ \\
                
                Davari et\,al.~\cite{Davari_Islam.2021} & & $4285 \pm 412$ & \fboxsep=1pt\colorbox{black!20}{$2140 \pm 41$} & / & / & $4342 \pm 264$ \\
    
                \multirow{ 2}{*}{Gourmelon et\,al.~\cite{Gourmelon_2022}} & Front & $2806 \pm 300$ & $\mathbf{191 \pm 32}$ & $127 \pm 38$ & $197 \pm 41$ & \fboxsep=1pt\colorbox{black!20}{$\mathbf{63 \pm 188}$} \\
                
                & Zones & $2201 \pm 246$ & $493 \pm 119$ & \fboxsep=1pt\colorbox{black!20}{$403 \pm 172$} & $437 \pm 172$ & $547 \pm 61$ \\

                Gourmelon et\,al.~\cite{Gourmelon.2023} && $2287 \pm 260$ & $491 \pm 86$ & $449 \pm 153$ & $408 \pm 48$ & \fboxsep=1pt\colorbox{black!20}{$218 \pm 51$} \\
                
                Hartmann et\,al.~\cite{Hartmann.2021} & & $2255 \pm 206$ & $583 \pm 81$ & \fboxsep=1pt\colorbox{black!20}{$465 \pm 133$} & $524 \pm 140$ & $850 \pm 34$ \\
                
                \multirow{ 2}{*}{Heilder et\,al.~\cite{Heidler.2021}} & Front & $1167 \pm 142$ & $354 \pm 138$ & \fboxsep=1pt\colorbox{black!20}{$152 \pm 21$} & $595 \pm 99$ & $395 \pm 38$ \\
                
                & Zones & $2106 \pm 372$ & $441 \pm 103$ & \fboxsep=1pt\colorbox{black!20}{$156 \pm 49$} & $481 \pm 114$ & $474 \pm 73$ \\
                
                Herrmann et\,al.~\cite{Herrmann.2023} & & $2605 \pm 316$ & $270 \pm 85$ & \fboxsep=1pt\colorbox{black!20}{$99 \pm 43$} & $\mathbf{195 \pm 44}$ & $302 \pm 118$ \\
                
                Holzmann et\,al.~\cite{Holzmann.2021} & & $3908 \pm 78$ & / & \fboxsep=1pt\colorbox{black!20}{$1135 \pm 0$} & $1176 \pm 0$ & $2103 \pm 314$ \\
                
                \multirow{ 2}{*}{Kirillov et\,al.~\cite{Kirillov.2023}}  & Iterative & $1650 \pm 126$ & $499 \pm 54$ & \fboxsep=1pt\colorbox{black!20}{$215 \pm 168$} & $420 \pm 86$ & $598 \pm 77$ \\
                
                & Parallel & $1653 \pm 103$ & $325 \pm 90$ & \fboxsep=1pt\colorbox{black!20}{$\mathbf{69 \pm 3}$} & $383 \pm 53$ & $655 \pm 119$ \\
                
                Loebel et\,al.~\cite{Loebel.2022} & & $2196 \pm 187$ & $608 \pm 200$ & $469 \pm 278$ & $360 \pm 107$ & \fboxsep=1pt\colorbox{black!20}{$344 \pm 43$} \\
                
                Marochov et\,al.~\cite{Marochov.2021} & & $1924 \pm 122$ & / & $1469 \pm 0$ & \fboxsep=1pt\colorbox{black!20}{$380 \pm 241$} & $4251 \pm 867$ \\
                
                Mohajerani et\,al.~\cite{Mohajerani.2019} & & $1491 \pm 221$ & \fboxsep=1pt\colorbox{black!20}{$431 \pm 54$} & $682 \pm 135$ & $457 \pm 70$ & $2085 \pm 48$ \\
                
                Periyasamy et\,al.~\cite{Periyasamy.2022} & & $2175 \pm 86$ & $1032 \pm 339$ & $801 \pm 240$ & \fboxsep=1pt\colorbox{black!20}{$633 \pm 89$} & $950 \pm 44$ \\
                
                Wu et\,al.~\cite{Wu.2023_1} & & $1504 \pm 207$ & $468 \pm 70$ & \fboxsep=1pt\colorbox{black!20}{$208 \pm 112$} & $328 \pm 130$ & $303 \pm 20$ \\
                
                Wu et\,al.~\cite{Wu.2023_2} & & $\mathbf{918 \pm 76}$ & $253 \pm 42$ & \fboxsep=1pt\colorbox{black!20}{$174 \pm 47$} & $263 \pm 28$ & $286 \pm 8$ \\
                
                Zhang et\,al.~\cite{Zhang.2021} & & $3927 \pm 837$ & $1926 \pm 68$ & $1368 \pm 709$ & $1838 \pm 402$ & \fboxsep=1pt\colorbox{black!20}{$1158 \pm 257$} \\
                
                Zhang et\,al.~\cite{Zhang.2023} & & $1905 \pm 540$ & $688 \pm 53$ & $642 \pm 388$ & \fboxsep=1pt\colorbox{black!20}{$557 \pm 90$} & $725 \pm 135$ \\
                
                Zhu et\,al.~\cite{Zhu.2023} & & $3094 \pm 730$ & $1276 \pm 266$ & \fboxsep=1pt\colorbox{black!20}{$395 \pm 172$} & $700 \pm 106$ & $812 \pm 63$ \\
                
                \bottomrule
                & & & & & & \\
            \end{tabular}
        \end{table*}

        \begin{table*}[htbp]
            \centering
            \caption{Number of images with no predicted front ($\varnothing$) for \ac{caffe}'s test set divided by capturing sensor. A \textbf{bold font} signifies the best value in a column, and a \fboxsep=1pt\colorbox{black!20}{grey background} indicates the best value in a row. Ties lead to multiple marked entries.}
            \label{tab:distance_errors_sensors_no_front}
            \begin{tabular}{p{0.2\textwidth} p{0.08\textwidth} p{0.11\textwidth} p{0.11\textwidth} p{0.11\textwidth} p{0.11\textwidth} p{0.11\textwidth}} 
                \toprule
                && \textit{Sentinel-1} & \textit{ENVISAT} & \textit{ERS} & \textit{PALSAR} & \textit{TSX} \\
                \textit{Paper} & & $\downarrow$ $\varnothing \in 33$ & $\downarrow$ $\varnothing \in 10$ & $\downarrow$ $\varnothing \in 2$ & $\downarrow$ $\varnothing \in 8$ & $\downarrow$ $\varnothing \in 69$ \\
                \midrule
    
                Cheng et\,al.~\cite{Cheng.2021} & & $4 \pm 3$ & $1 \pm 1$ & \fboxsep=1pt\colorbox{black!20}{$\mathbf{0 \pm 0}$} & \fboxsep=1pt\colorbox{black!20}{$\mathbf{0 \pm 0}$} & $10 \pm 7$ \\
                
                Davari et\,al.~\cite{Davari_Baller.2021} & & $8 \pm 1$ & $4 \pm 2$ & \fboxsep=1pt\colorbox{black!20}{$\mathbf{0 \pm 0}$} & $2 \pm 1$ & $33 \pm 1$ \\
                
                Davari et\,al.~\cite{Davari_Islam.2021} & & $16 \pm 1$ & $9 \pm 1$ & \fboxsep=1pt\colorbox{black!20}{$2 \pm 0$} & $8 \pm 0$ & $33 \pm 1$ \\
    
                \multirow{ 2}{*}{Gourmelon et\,al.~\cite{Gourmelon_2022}} & Front & $2 \pm 1$ & $2 \pm 2$ & \fboxsep=1pt\colorbox{black!20}{$\mathbf{0 \pm 0}$} & $3 \pm 2$ & \fboxsep=1pt\colorbox{black!20}{$\mathbf{0 \pm 0}$} \\
                
                & Zones & \fboxsep=1pt\colorbox{black!20}{$\mathbf{0 \pm 0}$} & \fboxsep=1pt\colorbox{black!20}{$\mathbf{0 \pm 0}$} & \fboxsep=1pt\colorbox{black!20}{$\mathbf{0 \pm 0}$} & \fboxsep=1pt\colorbox{black!20}{$\mathbf{0 \pm 0}$} & \fboxsep=1pt\colorbox{black!20}{$\mathbf{0 \pm 0}$} \\
                
                Gourmelon et\,al.~\cite{Gourmelon.2023} & & \fboxsep=1pt\colorbox{black!20}{$\mathbf{0 \pm 0}$} & \fboxsep=1pt\colorbox{black!20}{$\mathbf{0 \pm 0}$} & \fboxsep=1pt\colorbox{black!20}{$\mathbf{0 \pm 0}$} & \fboxsep=1pt\colorbox{black!20}{$\mathbf{0 \pm 0}$} & \fboxsep=1pt\colorbox{black!20}{$\mathbf{0 \pm 0}$} \\
                
                Hartmann et\,al.~\cite{Hartmann.2021} & & $2 \pm 3$ & \fboxsep=1pt\colorbox{black!20}{$\mathbf{0 \pm 0}$} & \fboxsep=1pt\colorbox{black!20}{$\mathbf{0 \pm 0}$} & \fboxsep=1pt\colorbox{black!20}{$\mathbf{0 \pm 0}$} & $9 \pm 10$ \\
                
                \multirow{ 2}{*}{Heidler et\,al.~\cite{Heidler.2021}} & Front & \fboxsep=1pt\colorbox{black!20}{$\mathbf{0 \pm 0}$} & $2 \pm 1$ & \fboxsep=1pt\colorbox{black!20}{$\mathbf{0 \pm 0}$} & $1 \pm 0$ & \fboxsep=1pt\colorbox{black!20}{$\mathbf{0 \pm 0}$} \\
                
                & Zones & $3 \pm 3$ & $2 \pm 2$ & \fboxsep=1pt\colorbox{black!20}{$\mathbf{0 \pm 0}$} & $1 \pm 1$ & \fboxsep=1pt\colorbox{black!20}{$\mathbf{0 \pm 0}$} \\
                
                Herrmann et\,al.~\cite{Herrmann.2023} & & $3 \pm 2$ & \fboxsep=1pt\colorbox{black!20}{$\mathbf{0 \pm 0}$} & \fboxsep=1pt\colorbox{black!20}{$\mathbf{0 \pm 0}$} & \fboxsep=1pt\colorbox{black!20}{$\mathbf{0 \pm 0}$} & \fboxsep=1pt\colorbox{black!20}{$\mathbf{0 \pm 0}$} \\
                
                Holzmann et\,al.~\cite{Holzmann.2021} & & $16 \pm 1$ & $10 \pm 0$ & \fboxsep=1pt\colorbox{black!20}{$2 \pm 0$} & $8 \pm 0$ & $41 \pm 4$ \\
                
                \multirow{ 2}{*}{Kirillov et\,al.~\cite{Kirillov.2023}}  & Iterative & $8 \pm 1$ & $1 \pm 0$ & \fboxsep=1pt\colorbox{black!20}{$\mathbf{0 \pm 0}$} & \fboxsep=1pt\colorbox{black!20}{$\mathbf{0 \pm 0}$} & $3 \pm 1$ \\
                
                & Parallel & $4 \pm 0$ & $2 \pm 1$ & \fboxsep=1pt\colorbox{black!20}{$\mathbf{0 \pm 0}$} & $1 \pm 0$ & $1 \pm 0$ \\
                
                Loebel et\,al.~\cite{Loebel.2022} & & $2 \pm 2$ & $3 \pm 2$ & \fboxsep=1pt\colorbox{black!20}{$\mathbf{0 \pm 0}$} & $1 \pm 1$ & \fboxsep=1pt\colorbox{black!20}{$\mathbf{0 \pm 0}$} \\
                
                Marochov et\,al.~\cite{Marochov.2021} & & $22 \pm 2$ & $10 \pm 0$ & \fboxsep=1pt\colorbox{black!20}{$2 \pm 0$} & $7 \pm 0$ & $57 \pm 3$ \\
                
                Mohajerani et\,al.~\cite{Mohajerani.2019} & & \fboxsep=1pt\colorbox{black!20}{$\mathbf{0 \pm 0}$} & \fboxsep=1pt\colorbox{black!20}{$\mathbf{0 \pm 0}$} & \fboxsep=1pt\colorbox{black!20}{$\mathbf{0 \pm 0}$} & \fboxsep=1pt\colorbox{black!20}{$\mathbf{0 \pm 0}$} & \fboxsep=1pt\colorbox{black!20}{$\mathbf{0 \pm 0}$} \\
                
                Periyasamy et\,al.~\cite{Periyasamy.2022} & & $4 \pm 3$ & $3 \pm 1$ & $1 \pm 0$ & \fboxsep=1pt\colorbox{black!20}{$\mathbf{0 \pm 0}$} & $3 \pm 1$ \\
                
                Wu et\,al.~\cite{Wu.2023_1} & & $1 \pm 1$ & $1 \pm 1$ & \fboxsep=1pt\colorbox{black!20}{$\mathbf{0 \pm 0}$} & $1 \pm 1$ & $1 \pm 1$ \\
                
                Wu et\,al.~\cite{Wu.2023_2} & & \fboxsep=1pt\colorbox{black!20}{$\mathbf{0 \pm 0}$} & \fboxsep=1pt\colorbox{black!20}{$\mathbf{0 \pm 0}$} & \fboxsep=1pt\colorbox{black!20}{$\mathbf{0 \pm 0}$} & \fboxsep=1pt\colorbox{black!20}{$\mathbf{0 \pm 0}$} & \fboxsep=1pt\colorbox{black!20}{$\mathbf{0 \pm 0}$} \\
                
                Zhang et\,al.~\cite{Zhang.2021} & & $24 \pm 2$ & $9 \pm 1$ & \fboxsep=1pt\colorbox{black!20}{$1 \pm 1$} & $6 \pm 1$ & $4 \pm 3$ \\
                
                Zhang et\,al.~\cite{Zhang.2023} & & \fboxsep=1pt\colorbox{black!20}{$\mathbf{0 \pm 0}$} & $ 4 \pm 3$ & \fboxsep=1pt\colorbox{black!20}{$\mathbf{0 \pm 0}$} & $0 \pm 1$ & $7 \pm 2$ \\
                
                Zhu et\,al.~\cite{Zhu.2023} & & $15 \pm 1$ & $8 \pm 2$ & $1 \pm 0$ & $1 \pm 1$ & \fboxsep=1pt\colorbox{black!20}{$\mathbf{0 \pm 0}$} \\
                
                \bottomrule
                & & & & & & \\
            \end{tabular}
        \end{table*}
        
        There is no single reason why a system has a lower \ac{mde} than another, but several factors contribute to different \acp{mde}.
    
        For systems with an \ac{mde} higher than \SI{1200}{\meter}, the possible reasons diverge:
        For Davari et\,al.~\cite{Davari_Islam.2021}, the network output is heavily speckled. 
        For some images, edges, such as the calving front and the edge between glacier and rock, show a higher density of predicted front pixels but still no connected front line. 
        Marochov et\,al.~\cite{Marochov.2021}'s system recognizes some higher-level structures, such as the approximate position of rocks, but cannot assign the patterns to the correct classes.
        The system of Holzmann et\,al.~\cite{Holzmann.2021} predicts too few front pixels, and the resulting fronts do not show enough curvature and detail and are not in close proximity to the ground truth front.
        Davari et\,al.~\cite{Davari_Baller.2021}'s system sometimes predicts the front in the wrong place. 
        In addition, the predicted front is usually too short and does not have enough curvature and detail. 
        The edge between the rock and glacier zones is often recognized as part of the front.
        Mohajerani et\,al.~\cite{Mohajerani.2019}'s system acts as a pixel-level edge detector, i.\,e., at a level where noise has a big influence, rather than recognizing global information. 
        This is also the reason why the number of images with no predicted front is zero. 
        Each image has pixel-level edges, which are thus incorrectly predicted as calving fronts.
        For Cheng et\,al.~\cite{Cheng.2021}, the predictions are speckled, and the system cannot recognize the classes correctly.
        Sometimes, some edges are found in the images but not between the correct classes.
        Lastly, Zhang et\,al.~\cite{Zhang.2021}'s system does not seem to be able to capture the general, global structure of the \ac{SAR} images; classes are mixed up, and the \ac{NA} region is not predicted correctly.
    
        For systems with an \ac{mde} between \SIrange{1200}{600}{\meter}, the main influences are varying degrees of patching artifacts (\cite{Periyasamy.2022}; \cite{Hartmann.2021}; \cite{Zhu.2023}; \cite{Zhang.2023}; \cite{Gourmelon_2022} Zones, \cite{Kirillov.2023} Parallel; \cite{Gourmelon.2023}; \cite{Kirillov.2023} Iterative; \cite{Heidler.2021} Zones), confusion of glacier and ocean class (\cite{Periyasamy.2022}; \cite{Hartmann.2021}; \cite{Zhu.2023};\cite{Zhang.2023}; \cite{Gourmelon_2022} Zones, \cite{Kirillov.2023} Parallel; \cite{Gourmelon.2023}; \cite{Kirillov.2023} Iterative; \cite{Heidler.2021} Zones), confusion of ice mélange as glacial ice (\cite{Periyasamy.2022}; \cite{Hartmann.2021}; \cite{Zhu.2023}; \cite{Zhang.2023}; \cite{Gourmelon_2022} Zones, \cite{Kirillov.2023} Parallel; \cite{Gourmelon.2023}; \cite{Kirillov.2023} Iterative; \cite{Heidler.2021} Zones), and confusion of the coastline and other edges between different zones as calving front (\cite{Periyasamy.2022}; \cite{Hartmann.2021}; \cite{Zhu.2023}; \cite{Zhang.2023}; \cite{Gourmelon_2022} Front; \cite{Gourmelon_2022} Zones, \cite{Kirillov.2023} Parallel; \cite{Gourmelon.2023}; \cite{Kirillov.2023} Iterative; \cite{Heidler.2021} Zones).
        In addition, the ocean class has many false positive predictions (\cite{Periyasamy.2022}; \cite{Hartmann.2021}; \cite{Zhu.2023}; \cite{Zhang.2023}; \cite{Gourmelon_2022} Zones, \cite{Kirillov.2023} Parallel; \cite{Gourmelon.2023}; \cite{Kirillov.2023} Iterative; \cite{Heidler.2021} Zones) and sometimes no ocean is predicted at all (\cite{Zhu.2023}; \cite{Kirillov.2023} Parallel; \cite{Kirillov.2023} Iterative).
        When the ocean is predicted in the correct location of the image, the ocean outline and, thus, the calving front often do not have the correct shape (\cite{Periyasamy.2022}; \cite{Hartmann.2021}; \cite{Zhu.2023};\cite{Zhang.2023}; \cite{Gourmelon_2022} Zones, \cite{Kirillov.2023} Parallel; \cite{Gourmelon.2023}; \cite{Kirillov.2023} Iterative; \cite{Heidler.2021} Zones).
        In binary front segmentation, the predicted fronts in the majority of images only cover parts of the ground truth, and many additional false positive fronts are predicted~\cite{Gourmelon_2022}.
    
        Only five systems have an \ac{mde} lower than \SI{600}{\meter}: Loebel et\,al.~\cite{Loebel.2022}, Herrmann et\,al.~\cite{Herrmann.2023}, Heidler et\,al.~\cite{Heidler.2021}'s front output, Wu et\,al.~\cite{Wu.2023_1} and Wu et\,al.~\cite{Wu.2023_2}. 
        All five systems confuse parts of the rocky coastline as calving front, have slight issues with ice mélange, and show a decreased delineation performance for images of the Columbia Glacier captured by Sentinel-1.
        The outputs of the model with the lowest average \ac{mde}, the HookFormer~\cite{Wu.2023_2}, additionally show slight patching artifacts and ragged edges between the classes.
        
    \subsection*{Statistical analysis}
    \label{app:statistical_analysis_results}
        \noindent The reported differences in the metrics between our \ac{dl} systems suggest that there is a significant difference for both the \ac{mde} (Chi-Squared(21)\;=\;$101.72$, p\;=\;$1.43e^{-12} < 0.05$) and the number of images with no predicted front (Chi-Squared(21) = $96.99$, p = $9.80e^{-12} < 0.05$).
        On average, the HookFormer~\cite{Wu.2023_2} has the predictions with the lowest \ac{mde}.
        All four differences in \ac{mde} to systems with an \ac{mde} lower than \SI{600}{m}, i.\;e., Wu et\,al.~\cite{Wu.2023_1}'s system, Heidler et\,al.~\cite{Heidler.2021}'s system's front output, Herrmann et\,al.~\cite{Herrmann.2023}'s system, and Loebel et\,al.~\cite{Loebel.2022}'s system are significant ($U = 0.0$, p\;=\;$3.97e^{-3} < 1.25e^{-2}$; $U = 0.0$, p\;=\;$3.97e^{-3} < 1.25e^{-2}$; $U = 0.0$, p\;=\;$3.97e^{-3} < 1.25e^{-2}$; $U = 0.0$, p\;=\;$3.97e^{-3} < 1.25^{-2}$), with effect sizes of $-3.58$, $-5.89$, $-2.66$, and $-7.36$ (Cohen's d), respectively.
        For the number of images with no predicted front, the differences to Wu et\,al.~\cite{Wu.2023_1}'s, Herrmann et\,al.~\cite{Herrmann.2023}'s, and Loebel et\,al.~\cite{Loebel.2022}'s systems are significant ($U = 0.0$, p\;=\;$3.54e^{-3} < 1.25e^{-2}$; $U = 0.0$, p\;=\;$3.35e^{-3} < 1.25e^{-2}$; $U = 0.0$, p\;=\;$3.65e^{-3} < 1.25e^{-2}$), with effect sizes of $-6.32$, $-2.18$ and $-4.50$ (Cohen's d).
        However, the difference to Heidler et\,al.~\cite{Heidler.2021}'s system's front output is not significant ($U = 5.0$, p\;=\;$3.60e^{-2} > 1.25e^{-2}$).
        
        The differences between base architecture groups are significant (Chi-square(4)\;=\;$24.82$, p\;=\;$5.47e^{-5} < 0.05$).
        The average \ac{mde} for each architecture group is \SI{2670}{\meter} for VGG16~\cite{Simonyan_2015_ICLR}, \SI{1324}{\meter} for DeepLabv3+~\cite{Chen_2018_ECCV}, \SI{1314}{\meter} for U-Nets~\cite{Ronneberger.2015}, \SI{914}{\meter} for a mix of DeepLabv3+ and \ac{ViT}, and \SI{607}{\meter} for \acp{ViT}~\cite{Dosovitskiy.2020}.
        The \ac{ViT}-based architectures outperform the mixed architecture, DeepLabv3+, U-Net, and VGG16-based architectures significantly ($U = 4.0$, p\;=\;$7.74e^{-4}<1.25e^{-2}$; $U = 12.0$, p\;=\;$1.68e^{-5}<1.25e^{-2}$; $U = 283.0$, p\;=\;$2.68e^{-3}<1.25e^{-2}$; $U = 0.0$, p\;=\;$6.45e^{-5}<1.25e^{-2}$), with effect sizes of $-1.78$, $-1.88$, $-0.71$, and $-8.78$ (Cohen's d), respectively.
        The differences between models trained on \ac{caffe}'s binary front labels, \ac{caffe}'s zone labels, and models trained in a multi-task manner on both labels are significant (Chi-Squared(2)\;=\;$36.30$, p\;=\;$1.31e^{-8} < 0.05$). The average \acp{mde} are \SI{2423}{\meter} for binary, \SI{938}{\meter} for zones, and \SI{864}{\meter} for \ac{mtl}. Both \ac{mtl} \ac{dl} systems and systems trained solely on the zone labels have a significantly lower \ac{mde} than \ac{dl} systems trained solely on the binary front labels ($U = 45.0$, p\;=\;$1.50e^{-6}<1.67e^{-2}$; $U = 198.0$, p\;=\;$1.59e^{-8}<1.67e^{-2}$), with effect sizes of $-1.66$ and $-1.92$ (Cohen's d). The difference of \ac{mtl} to training on the zone labels is not significant ($U = 480.0$, p\;=\;$3.95e^{-2} > 1.67e^{-2}$).
        
        With a Kendall's $\tau$ of $-0.15$ (p\;=\;$2.53e^{-2} < 0.05$), the \ac{mde} and the mean input size in pixels during training are significantly negatively correlated, i.\,e., the bigger the input size, the lower the \ac{mde}.
        Moreover, the number of down-sampling steps in U-Nets is significantly negatively correlated with the \ac{mde}, with a Kendall's $\tau$ of $-0.48$ (p\;=\;$2.03e^{-7} < 0.05$), i.\,e., the more local-global information interaction, the lower the \ac{mde}.

    \subsection*{Multi-annotator study}
    \label{app:results_multi_annotator}
        \noindent Fig.~\ref{fig:annotators} gives an overview of the annotators' levels of expertise.
        The \ac{mde} of the automatic annotations from the best-performing \ac{dl} system is significantly higher than that of the manual annotations ($U = 50.0$, $p = 3.33e^{-4}$), with an effect size of $11.82$ (Cohen's d).
        Table~\ref{tab:mde_hookformer_vs_annotators} provides the \acp{mde} between the best-performing \ac{dl} system and each single annotator as well as the combined ground truth. No bias towards annotator number ten, who annotated the \ac{caffe} dataset, can be observed.
        
        \begin{figure*}
            \centering
            \begin{tikzpicture}

    \begin{axis} [xbar = .05cm,
    	bar width = 12pt,
    	xmin = 0,
    	xmax = 6,
    	ytick = {0,1,2},
        yticklabels = {Basic, Intermediate, Expert},
    	enlarge x limits = {value = .25, upper},
    	enlarge y limits = {abs = .8},
        legend pos= south east,
        height=7cm,
        width=0.6\textwidth, 
        legend cell align={left},
    ]
    
        \addplot coordinates {(3,0) (4,1) (3,2)};   
        \addplot coordinates {(1,0) (6,1) (3,2)};   
        
        \legend { QGIS Proficiency, Glacier Knowledge};
    
    \end{axis}

\end{tikzpicture}
            \caption{Overview of the annotators' QGIS proficiency and their knowledge of glaciers.}
            \label{fig:annotators}
        \end{figure*}

        \begin{sidewaystable*}[htbp]
        \centering
        \caption{Mean Distance Errors (MDEs) between the post-processed HookFormer predictions and the single annotators as well as the combined ground truth. 
        The HookFormer is trained on annotator number ten.
        The \ac{mde} is calculated for different subsets: complete test set (All), summer (Sum.), winter (Win.), Mapple Glacier (Map.), Columbia Glacier (Col.), Sentinel-1 (S1), Envisat (Envi.), ERS-1/2 (ERS), ALOS PALSAR (PAL.), TerraSAR-X and TanDEM-X (TSX), a resolution of \SI{20}{\meter} (20), a resolution of \SI{17}{\meter} (17), and a resolution of \SI{7}{\meter} (7). The first column gives the number of the annotator, where ``Com.'' stands for the combined ground truth. A \textbf{bold font} signifies the best value in a column.}
        \label{tab:mde_hookformer_vs_annotators}
        \resizebox{\columnwidth}{!}{
        \begin{tabular} {p{0.031\textwidth} p{0.059\textwidth} p{0.055\textwidth} p{0.059\textwidth} p{0.059\textwidth} p{0.059\textwidth} p{0.067\textwidth} p{0.059\textwidth} p{0.059\textwidth} p{0.059\textwidth} p{0.055\textwidth} p{0.059\textwidth} p{0.059\textwidth} p{0.055\textwidth}} 
            \toprule
            && \multicolumn{2}{c}{\textit{Season}} & \multicolumn{2}{c}{\textit{Glacier}} & \multicolumn{5}{c}{\textit{Sensor}} & \multicolumn{3}{c}{\textit{Resolution}}\\
            
            & \textit{All} & \textit{Sum.} & \textit{Win.} & \textit{Map.} & \textit{Col.} & \textit{S1} & \textit{Envi.} & \textit{ERS} & \textit{PAL.} & \textit{TSX} & \textit{20} & \textit{17} & \textit{7}\\
            \midrule
             
            \# 1 & $259 \pm 15$ & $220 \pm 8$ & $301 \pm 32$ & $277 \pm 16$ & $255 \pm 18$ & $925 \pm 108$ & $768 \pm 30$ & $\mathbf{152 \pm 40}$ & $723 \pm 86$ & $119 \pm 5$ & $882 \pm 89$ & $723 \pm 86$ & $119 \pm 5$ \\
            \# 2 & $234 \pm 16$ & $180 \pm 8$ & $291 \pm 32$ & $121 \pm 14$ & $259 \pm 21$ & $962 \pm 108$ & $242 \pm 38$ & $159 \pm 43$ & $195 \pm 23$ & $110 \pm 5$ & $858 \pm 91$ & $195 \pm 23$ & $110 \pm 5$ \\
            \# 3 & $223 \pm 14$ & $178 \pm 9$ & $269 \pm 31$ & $135 \pm 14$ & $242 \pm 19$ & $891 \pm 102$ & $256 \pm 48$ & $163 \pm 48$ & $296 \pm 27$ & $103 \pm 3$ & $798 \pm 87$ & $296 \pm 27$ & $103 \pm 3$ \\
            \# 4 & $215 \pm 16$ & $174 \pm 8$ & $\mathbf{260 \pm 31}$ & $124 \pm 14$ & $\mathbf{236 \pm 22}$ & $\mathbf{851 \pm 104}$ & $287 \pm 18$ & $153 \pm 44$ & $200 \pm 27$ & $103 \pm 5$ & $\mathbf{762 \pm 88}$ & $200 \pm 27$ & $103 \pm 5$ \\
            \# 5 & $226 \pm 17$ & $183 \pm 9$ & $271 \pm 33$ & $113 \pm 13$ & $251 \pm 23$ & $900 \pm 112$ & $247 \pm 43$ & $\mathbf{152 \pm 38}$ & $\mathbf{183 \pm 36}$ & $112 \pm 6$ & $804 \pm 95$ & $\mathbf{183 \pm 36}$ & $112 \pm 6$ \\
            \# 6 & $223 \pm 16$ & $\mathbf{173 \pm 9}$ & $276 \pm 33$ & $122 \pm 14$ & $246 \pm 21$ & $916 \pm 112$ & $271 \pm 38$ & $160 \pm 47$ & $210 \pm 30$ & $\mathbf{102 \pm 5}$ & $819 \pm 94$ & $210 \pm 30$ & $\mathbf{102 \pm 5}$ \\
            \# 7 & $226 \pm 17$ & $184 \pm 9$ & $271 \pm 33$ & $114 \pm 13$ & $251 \pm 23$ & $905 \pm 113$ & $230 \pm 35$ & $156 \pm 39$ & $196 \pm35$ & $112 \pm 6$ & $805 \pm 95$ & $196 \pm 35$ & $112 \pm 6$ \\
            \# 8 & $232 \pm 18$ & $177 \pm 9$ & $290 \pm 36$ & $117 \pm 13$ & $257 \pm 23$ & $960 \pm 120$ & $\mathbf{218 \pm 37}$ & $154 \pm 43$ & $206 \pm 37$ & $105 \pm 4$ & $853 \pm 102$ & $206 \pm 37$ & $105 \pm 4$ \\
            \# 9 & $223 \pm 15$ & $178 \pm 9$ & $271 \pm 31$ & $124 \pm 15$ & $245 \pm 20$ & $907 \pm 101$ & $266 \pm 39$ & $158 \pm 43$ & $199 \pm 31$ & $106 \pm 5$ & $811 \pm 85$ & $199 \pm 31$ & $106 \pm 5$ \\
            \# 10 & $238 \pm 16$ & $187 \pm 8$ & $291 \pm 33$ & $115 \pm 14$ & $265 \pm 21$ & $941 \pm 114$ & $239 \pm 39$ & $155 \pm 46$ & $211 \pm 38$ & $116 \pm 5$ & $839 \pm 96$ & $211 \pm 38$ & $116 \pm 5$ \\
            Com. & $\mathbf{221 \pm 15}$ & $174 \pm 9$ & $271 \pm 32$ & $\mathbf{110 \pm 14}$ & $245 \pm 20$ & $915 \pm 108$ & $230 \pm 40$ & $157 \pm 46$ & $186 \pm 32$ & $105 \pm 5$ & $813 \pm 91$ & $186 \pm 32$ & $105 \pm 5$ \\
            \bottomrule
        \end{tabular}
        }
    \end{sidewaystable*}

\section*{Code and data availability}
\noindent The benchmark dataset \ac{caffe} is available at \url{https://doi.org/10.1594/PANGAEA.940950}~\cite{gourmelon2022ccfa}.\\
Codes for the \ac{dl} systems can be found in their studies' respective repositories:\\
\url{https://github.com/daniel-cheng/CALFIN},\\
\url{https://github.com/VChristlein/PixelwiseDistanceRegression4GlacierSegmentation},\\
\url{https://github.com/zetaSaahil/Glacier-CFL-detection_DMapBCE},\\
\url{https://github.com/Nora-Go/Calving_Fronts_and_Where_to_Find_Them},\\
\url{https://github.com/EntChanelt/GlacierCRF},\\
\url{https://github.com/VChristlein/BayesianUNet4GlacierSegmentation/},\\
\url{https://github.com/khdlr/HED-UNet},\\
\url{https://github.com/ho11laqe/nnUNet_calvingfront_detection},\\
\url{https://github.com/VChristlein/AttentionUNet4GlacierSegmentation/},\\
\url{https://github.com/facebookresearch/segment-anything},\\
\url{https://github.com/eloebel/glacier-front-extraction},\\
\url{https://github.com/PCdurham/SEE_ICE},\\
\url{https://github.com/yaramohajerani/FrontLearning},\\
\url{https://github.com/VChristlein/MostOutOfUNet4GlacierSegmentation},\\
\url{https://github.com/RiverNA/AMD-HookNet},\\
\url{https://github.com/RiverNA/HookFormer},\\
\url{https://github.com/enzezhang/FrontDL3},\\
\url{https://zenodo.org/records/8270875}, and\\
\url{https://github.com/Tangyu35/Calving-front-detection}.\\
Figures~\ref{fig:all_models_Col} and~\ref{fig:all_models_Mapple} show a subset of \ac{caffe}'s images. 
The full set of visualizations is provided at \url{ https://doi.org/10.5281/zenodo.11484341}.\\

\bibliographystyle{IEEEtran}
\bibliography{references}

@Article{Kruskal_1952,
    AUTHOR = {Kruskal, W.\,H. and Wallis, W.\,W.},
    YEAR = {1952},
    TITLE = {Use of Ranks in One-Criterion Variance Analysis},
    JOURNAL = {Journal of the American Statistical Association},
    VOLUME = {47},
    ISSUE = {260},
    PAGES = {583--621},
    DOI = {10.1080/01621459.1952.10483441}
}

@article{Gourmelon_2022,
 author = {Gourmelon, N. and Seehaus, T. and Braun, M. and Maier, A. and Christlein, V.},
 year = {2022},
 title = {Calving fronts and where to find them: a benchmark dataset and methodology for automatic glacier calving front extraction from synthetic aperture radar imagery},
 url = {https://essd.copernicus.org/articles/14/4287/2022/},
 pages = {4287--4313},
 volume = {14},
 number = {9},
 journal = {Earth System Science Data},
 doi = {10.5194/essd-14-4287-2022 }
}

@INPROCEEDINGS{Gourmelon.2023,
  author={Gourmelon, N. and Klink, J. and Seehaus, T. and Braun, M. and Maier, A. and Christlein, V.},
  booktitle={IEEE International Geoscience and Remote Sensing Symposium (IGARSS)},
  title={Conditional Random Fields for improving deep learning-based glacier calving front delineations},
  year={2023},
  volume={},
  number={},
  pages={4939-4942},
  doi={}
}

@book{Bishop.1995,
  title={Neural networks for pattern recognition},
  author={Bishop, C. M.},
  year={1995},
  edition={14},
  publisher={Clarendon Press},
  isbn={978-0198538493}
}

@inproceedings{Dosovitskiy.2020,
title={An Image is Worth 16x16 Words: Transformers for Image Recognition at Scale},
author={Dosovitskiy et al., A.},
booktitle={International Conference on Learning Representations},
year={2021},
url={https://openreview.net/forum?id=YicbFdNTTy}
}

@InProceedings{Chollet_2017_CVPR,
author = {Chollet, F.},
title = {Xception: Deep Learning With Depthwise Separable Convolutions},
booktitle = {Proceedings of the IEEE Conference on Computer Vision and Pattern Recognition (CVPR)},
month = {July},
year = {2017}
}

@InProceedings{He_2016_CVPR,
author = {He, K. and Zhang, X. and Ren, S. and Sun, J.},
title = {Deep Residual Learning for Image Recognition},
booktitle = {Proceedings of the IEEE Conference on Computer Vision and Pattern Recognition (CVPR)},
month = {June},
year = {2016}
}

@InProceedings{Yu_2017_CVPR,
author = {Yu, F. and Koltun, V. and Funkhouser, T.},
title = {Dilated Residual Networks},
booktitle = {Proceedings of the IEEE Conference on Computer Vision and Pattern Recognition (CVPR)},
month = {July},
year = {2017}
}

@article{vanRijthoven.2021,
 author = {{van Rijthoven}, M. and Balkenhol, M. and Siliņa, K. and {van der Laak}, J. and Ciompi, F.},
 year = {2021},
 title = {HookNet: Multi-resolution convolutional neural networks for semantic segmentation in histopathology whole-slide images},
 url = {https://www.sciencedirect.com/science/article/pii/S1361841520302541},
 pages = {101890},
 volume = {68},
 issn = {1361-8415},
 journal = {Medical Image Analysis},
 doi = {10.1016/j.media.2020.101890 },
}

@misc{Howard_2017_MobileNets,
      title={MobileNets: Efficient Convolutional Neural Networks for Mobile Vision Applications}, 
      author={Howard et al., A. G.},
      year={2017},
      eprint={1704.04861},
      archivePrefix={arXiv},
      primaryClass={cs.CV},
        note = {Preprint at http://arxiv.org/abs/1704.04861}
}

@article{Isensee.2021,
 author = {Isensee, F. and Jaeger, P. F. and Kohl, S. A. A. and Petersen, J. and Maier-Hein, K. H.},
 year = {2021},
 title = {nnU-Net: a self-configuring method for deep learning-based biomedical image segmentation},
 pages = {203--211},
 volume = {18},
 number = {2},
 journal = {Nature methods},
 doi = {10.1038/s41592-020-01008-z }
}

@ARTICLE{Chen.2018,
  author={Chen, L.-C. and Papandreou, G. and Kokkinos, I. and Murphy, K. and Yuille, A. L.},
  journal={IEEE T. Pattern. Anal.}, 
  title={DeepLab: Semantic Image Segmentation with Deep Convolutional Nets, Atrous Convolution, and Fully Connected CRFs}, 
  year={2018},
  volume={40},
  number={4},
  pages={834--848},
  doi={10.1109/TPAMI.2017.2699184 }
 }

@InProceedings{Chen_2018_ECCV,
author = {Chen, L.-C. and Zhu, Y. and Papandreou, G. and Schroff, F. and Adam, H.},
title = {Encoder-Decoder with Atrous Separable Convolution for Semantic Image Segmentation},
booktitle = {Proceedings of the European Conference on Computer Vision (ECCV)},
month = {09},
year = {2018}
}

@inproceedings{Simonyan_2015_ICLR,
  author    = {Simonyan, K. and
               Zisserman, A.},
  editor    = {Bengio, Y. and
               LeCun, Y.},
  title     = {Very Deep Convolutional Networks for Large-Scale Image Recognition},
  booktitle = {International Conference on Learning Representations (ICLR)},
  year      = {2015},
  url       = {http://arxiv.org/abs/1409.1556},
  bibsource = {dblp computer science bibliography, https://dblp.org}
}

@inproceedings{Sudre.2017,
 author = {Sudre, C. H. and Li, W. and Vercauteren, T. and Ourselin, S. and {Jorge Cardoso}, M.},
 title = {Generalised Dice Overlap as a Deep Learning Loss Function for Highly Unbalanced Segmentations},
 pages = {240--248},
 booktitle = {Deep Learning in Medical Image Analysis and Multimodal Learning for Clinical Decision Support},
 year = {2017},
 doi = {10.1007/978-3-319-67558-9_28},
}

@article{Goliber.2022,
 author = {Goliber et al., S.},
 year = {2022},
 title = {TermPicks: a century of Greenland glacier terminus data for use in scientific and machine learning applications},
 url = {https://tc.copernicus.org/articles/16/3215/2022/},
 pages = {3215--3233},
 volume = {16},
 number = {8},
 journal = {The Cryosphere},
 doi = {10.5194/tc-16-3215-2022},
}

@inproceedings{Ronneberger.2015,
 author = {Ronneberger, O. and Fischer, P. and Brox, T.},
 title = {U-Net: Convolutional Networks for Biomedical Image Segmentation},
 pages = {234--241},
 publisher = {{Springer International Publishing}},
 isbn = {978-3-319-24574-4},
 editor = {Navab, Nassir and Hornegger, Joachim and Wells, William M. and Frangi, Alejandro F.},
 booktitle = {Medical Image Computing and Computer-Assisted Intervention (MICCAI)},
 year = {2015},
 address = {Cham},
}

@article{Recinos_2019,
	title = {Impact of frontal ablation on the ice thickness estimation of marine-terminating glaciers in {Alaska}},
	volume = {13},
	issn = {1994-0416},
	url = {https://tc.copernicus.org/articles/13/2657/2019/},
	doi = {10.5194/tc-13-2657-2019},
	language = {English},
	number = {10},
	journal = {The Cryosphere},
	author = {Recinos, B. and Maussion, F. and Rothenpieler, T. and Marzeion, B.},
	month = oct,
	year = {2019},
	pages = {2657--2672}
}

@article{Marochov.2021,
 author = {Marochov, M. and Stokes, C. R. and Carbonneau, P. E.},
 year = {2021},
 title = {Image classification of marine-terminating outlet glaciers in Greenland  using deep learning methods},
 url = {https://tc.copernicus.org/articles/15/5041/2021/},
 pages = {5041--5059},
 volume = {15},
 number = {11},
 journal = {The Cryosphere},
 doi = {10.5194/tc-15-5041-2021}
}

@article{Periyasamy.2022,
 author = {Periyasamy, M. and Davari, A. and Seehaus, T. and Braun, M. and Maier, A. and Christlein, V.},
 year = {2022},
 title = {How to Get the Most Out of U-Net for Glacier Calving Front Segmentation},
 volume={15},
 number={},
 pages={1712-1723},
 issn = {1939-1404},
 journal = {IEEE Journal of Selected Topics in Applied Earth Observations and Remote Sensing},
 doi = {10.1109/JSTARS.2022.3148033},
}

@inproceedings{Hartmann.2021,
 author = {Hartmann, A. and Davari, A. and Seehaus, T. and Braun, M. and Maier, A. and Christlein, V.},
 title = {Bayesian U-Net for Segmenting Glaciers in SAR Imagery},
 url = {10.1109/IGARSS47720.2021.9554292},
 pages = {3479--3482},
 booktitle = {IEEE International Geoscience and Remote Sensing Symposium (IGARSS)},
 year = {2021},
}

@inproceedings{Holzmann.2021,
 author = {Holzmann, M. and Davari, A. and Seehaus, T. and Braun, M. and Maier, A. and Christlein, V.},
 title = {Glacier Calving Front Segmentation Using Attention U-Net},
 url = {10.1109/IGARSS47720.2021.9555067},
 pages = {3483--3486},
 booktitle = {IEEE International Geoscience and Remote Sensing Symposium (IGARSS)},
 year = {2021},
}

@ARTICLE{Davari_Baller.2021,
  author={Davari, A. and Baller, C. and Seehaus, T. and Braun, M. and Maier, A. and Christlein, V.},
  journal={IEEE Transactions on Geoscience and Remote Sensing}, 
  title={Pixelwise Distance Regression for Glacier Calving Front Detection and Segmentation}, 
  year={2022},
  volume={60},
  number={},
  pages={1-10},
  doi={10.1109/TGRS.2022.3158591}}

@article{Davari_Islam.2021,
 author = {Davari et al., A.},
 year = {2021},
 title = {On Mathews Correlation Coefficient and Improved Distance Map Loss for Automatic Glacier Calving Front Segmentation in SAR Imagery},
 volume={60},
 pages = {1--12},
 journal = {IEEE Transactions on Geoscience and Remote Sensing},
 doi = {10.1109/TGRS.2021.3115883 }
}

@article{Baumhoer.2019,
 author = {Baumhoer, C. A. and Dietz, A. J. and Kneisel, C. and Kuenzer, C.},
 year = {2019},
 title = {Automated Extraction of Antarctic Glacier and Ice Shelf Fronts from Sentinel-1 Imagery Using Deep Learning},
 pages = {2529},
 volume = {11},
 number = {21},
 journal = {Remote Sensing},
 doi = {10.3390/rs11212529}
}

@article{Cheng.2021,
 author = {Cheng et al., D.},
 year = {2021},
 title = {Calving Front Machine (CALFIN): glacial termini dataset  and automated deep learning extraction method  for Greenland, 1972--2019},
 url = {https://tc.copernicus.org/articles/15/1663/2021/},
 pages = {1663--1675},
 volume = {15},
 number = {3},
 journal = {The Cryosphere},
 doi = {10.5194/tc-15-1663-2021}
}

@article{Heidler.2021,
 author = {Heidler, K. and Mou, L. and Baumhoer, C. and Dietz, A. and Zhu, X. X.},
 year = {2021},
 title = {HED-UNet: Combined Segmentation and Edge Detection for Monitoring the Antarctic Coastline},
 pages = {1--14},
 volume={60},
 journal = {IEEE Transactions on Geoscience and Remote Sensing},
 doi = {10.1109/TGRS.2021.3064606}
}

@article{Heidler.2023,
 author = {Heidler, K. and Mou, L. and Loebel, E. and Scheinert, M. and Lef{\`e}vre, S. and Zhu, X. X.},
 year = {2023},
 title = {A Deep Active Contour Model for Delineating Glacier Calving Fronts},
 volume={61},
 number={},
 pages={1-12},
 journal = {IEEE Transactions on Geoscience and Remote Sensing}
}

@article{Herrmann.2023,
 author = {Herrmann et al., O.},
 year = {2023},
 title = {Out-of-the-box calving-front detection method using deep learning},
 url = {https://tc.copernicus.org/articles/17/4957/2023/},
 pages = {4957--4977},
 volume = {17},
 number = {11},
 journal = {The Cryosphere},
 doi = {10.5194/tc-17-4957-2023 }
}

@InProceedings{Kirillov.2023,
    author    = {Kirillov et al., A.},
    title     = {Segment Anything},
    booktitle = {Proceedings of the IEEE/CVF International Conference on Computer Vision (ICCV)},
    month     = {October},
    year      = {2023},
    pages     = {4015-4026}
}

@misc{Wang.2023,
 author = {Wang, X. and Zhang, X. and Cao, Y. and Wang, W. and Shen, C. and Huang, T.},
 year = {2023},
 title = {SegGPT: Segmenting Everything In Context},
 url = {https://arxiv.org/abs/2304.03284},
 publisher = {{arXiv preprint}},
 note = {Preprint at https://arxiv.org/abs/2304.03284}
}

@inproceedings{Zou.2023,
 author = {Zou et al., X.},
 booktitle = {Advances in Neural Information Processing Systems},
 editor = {A. Oh and T. Naumann and A. Globerson and K. Saenko and M. Hardt and S. Levine},
 pages = {19769--19782},
 publisher = {Curran Associates, Inc.},
 title = {Segment Everything Everywhere All at Once},
 url = {https://proceedings.neurips.cc/paper_files/paper/2023/file/3ef61f7e4afacf9a2c5b71c726172b86-Paper-Conference.pdf},
 volume = {36},
 year = {2023}
}

@article{Loebel.2022,
 author = {Loebel et al., E.},
 year = {2022},
 title = {Extracting glacier calving fronts by deep learning: the benefit of multi-spectral, topographic and textural input features},
 volume={60},
 number={},
 pages={1-12},
 journal = {IEEE Transactions on Geoscience and Remote Sensing},
 doi = {10.1109/TGRS.2022.3208454 },
}

@article{Mohajerani.2019,
 author = {Mohajerani, Y. and Wood, M. and Velicogna, I. and Rignot, E.},
 year = {2019},
 title = {Detection of Glacier Calving Margins with Convolutional Neural Networks: A Case Study},
 pages = {74},
 volume = {11},
 number = {1},
 journal = {Remote Sensing},
 doi = {10.3390/rs11010074 }
}

@article{Wu.2023_1,
 author = {Wu et al., F.},
 year = {2023},
 title = {AMD-HookNet for Glacier Front Segmentation},
 pages = {1--12},
 volume = {61},
 journal = {IEEE Transactions on Geoscience and Remote Sensing},
 doi = {10.1109/TGRS.2023.3245419  }
}

@ARTICLE{Wu.2023_2,
  author={Wu et al., F.},
  journal={IEEE Transactions on Geoscience and Remote Sensing}, 
  title={Contextual HookFormer for Glacier Calving Front Segmentation}, 
  year={2024},
  volume={62},
  number={},
  pages={1-15},
  doi={10.1109/TGRS.2024.3368215}}

@article{Zhang.2019,
 author = {Zhang, E. and Liu, L. and Huang, L.},
 year = {2019},
 title = {Automatically delineating the calving front of Jakobshavn Isbr{\ae} from multitemporal TerraSAR-X images: a deep learning approach},
 pages = {1729--1741},
 volume = {13},
 number = {6},
 journal = {The Cryosphere},
 doi = {10.5194/tc-13-1729-2019},
}

@article{Zhang.2021                           ,
 author = {Zhang, E. and Liu, L. and Huang, L. and Ng, K. S.},
 year = {2021},
 title = {An automated, generalized, deep-learning-based method for delineating the calving fronts of Greenland glaciers from multi-sensor remote sensing imagery},
 pages = {112265},
 volume = {254},
 issn = {00344257},
 journal = {Remote Sensing of Environment},
 doi = {10.1016/j.rse.2020.112265}
}

@article{Zhang.2023,
 author = {Zhang, E. and Catania, G. and Trugman, D. T.},
 year = {2023},
 title = {AutoTerm: an automated pipeline for glacier terminus extraction using machine learning and a ``big data'' repository of Greenland glacier termini},
 url = {https://tc.copernicus.org/articles/17/3485/2023/},
 pages = {3485--3503},
 volume = {17},
 number = {8},
 journal = {The Cryosphere},
 doi = {10.5194/tc-17-3485-2023},
}

@ARTICLE{Zhu.2023,
  author={Zhu et al., Q.},
  journal={IEEE Transactions on Geoscience and Remote Sensing}, 
  title={GLA-STDeepLab: SAR Enhancing Glacier and Ice Shelf Front Detection Using Swin-TransDeepLab With Global–Local Attention}, 
  year={2023},
  volume={61},
  number={},
  pages={1-13},
  doi={10.1109/TGRS.2023.3324404 }
}

@InProceedings{Liu.2021,
    author    = {Liu et al., Z.},
    title     = {Swin Transformer: Hierarchical Vision Transformer Using Shifted Windows},
    booktitle = {Proceedings of the IEEE/CVF International Conference on Computer Vision (ICCV)},
    month     = {October},
    year      = {2021},
    pages     = {10012-10022}
}

@article{Bondzio.2017,   
author = {Bondzio et al., J. H.},
title = {The mechanisms behind Jakobshavn Isbræ's acceleration and mass loss: A 3-D thermomechanical model study},
journal = {Geophysical Research Letters},
volume = {44},
number = {12},
pages = {6252-6260},
doi = {https://doi.org/10.1002/2017GL073309},   
url = {https://agupubs.onlinelibrary.wiley.com/doi/abs/10.1002/2017GL073309},
year = {2017}
}

@article{Vieli.2011,
 author = {Vieli, A. and Nick, F. M.},
 year = {2011},
 title = {Understanding and Modelling Rapid Dynamic Changes of Tidewater Outlet Glaciers: Issues and Implications},
 pages = {437--458},
 volume = {32},
 number = {4},
 issn = {1573-0956},
 journal = {Surveys in Geophysics},
 doi = {10.1007/s10712-011-9132-4  }
}

@article{Rounce.2023,
 author = {Rounce et al., D. R.},
 year = {2023},
 title = {Global glacier change in the 21st century: Every increase in temperature matters},
 pages = {78--83},
 volume = {379},
 number = {6627},
 issn = {1095-9203},
 journal = {Science},
 doi = {10.1126/science.abo1324 },
}

@Article{Otosaka.2023,
AUTHOR = {Otosaka et al., I. N.},
TITLE = {Mass balance of the Greenland and Antarctic ice sheets from 1992 to 2020},
JOURNAL = {Earth System Science Data},
VOLUME = {15},
YEAR = {2023},
NUMBER = {4},
PAGES = {1597--1616},
URL = {https://essd.copernicus.org/articles/15/1597/2023/},
DOI = {10.5194/essd-15-1597-2023}
}

@article{Kochtitzky.2023, 
author={Kochtitzky et al., W.},
title={Progress toward globally complete frontal ablation estimates of marine-terminating glaciers}, 
volume={63}, 
DOI={10.1017/aog.2023.35}, 
number={87–89}, 
journal={Annals of Glaciology},  
year={2022},
pages={143–152}}

@article{Kochtitzky.2022,
 author = {Kochtitzky et al., W.},
 year = {2022},
 title = {The unquantified mass loss of Northern Hemisphere marine-terminating glaciers from 2000--2020},
 pages = {5835},
 volume = {13},
 number = {1},
 journal = {Nature communications},
 doi = {10.1038/s41467-022-33231-x},
}

@article{Sheperd.2018,
 author = {Sheperd et al., A.},
 year = {2018},
 title = {Mass balance of the Antarctic Ice Sheet from 1992 to 2017},
 pages = {219--222},
 volume = {558},
 number = {7709},
 journal = {Nature},
 doi = {10.1038/s41586-018-0179-y}
}

@article{Hugonnet.2021,
 author = {Hugonnet et al., R.},
 year = {2021},
 title = {Accelerated global glacier mass loss in the early twenty-first century},
 pages = {726--731},
 volume = {592},
 number = {7856},
 journal = {Nature},
 doi = {10.1038/s41586-021-03436-z}
}

@article{Khan.2015,
 author = {Khan, S. A. and Aschwanden, A. and Bj{\o}rk, A. A. and Wahr, J. and Kjeldsen, K. K. and Kj{\ae}r, K. H.},
 year = {2015},
 title = {Greenland ice sheet mass balance: a review},
 pages = {046801},
 volume = {78},
 number = {4},
 journal = {Reports on progress in physics. Physical Society (Great Britain)},
 doi = {10.1088/0034-4885/78/4/046801}
}

@misc{gourmelon2022ccfa,
 author={Nora {Gourmelon} and Thorsten {Seehaus} and Matthias Holger {Braun} and Andreas {Maier} and Vincent {Christlein}},
 title={{CaFFe (CAlving Fronts and where to Find thEm: a benchmark dataset and methodology for automatic glacier calving front extraction from sar imagery)}},
 year={2022},
 doi={10.1594/PANGAEA.940950},
 url={https://doi.org/10.1594/PANGAEA.940950}   ,
 type={data set},
 publisher={PANGAEA},
 note = {{PANGAEA}, https://doi.org/10.1594/PANGAEA.940950} 
}

@phdthesis{Thiel_1994,
  TITLE = {{Les distances de chanfrein en analyse d'images : fondements et applications}},
  AUTHOR = {Thiel, E.},
  URL = {https://theses.hal.science/tel-00005113},
  HAL_LOCAL_REFERENCE = {theses/1994/Thiel.Edouard},
  SCHOOL = {{Universit{\'e} Joseph-Fourier - Grenoble I}},
  YEAR = {1994},
  MONTH = Sep,
  TYPE = {Theses},
  HAL_ID = {tel-00005113},
  HAL_VERSION = {v1},
}

@ARTICLE{Yeghiazaryan_2018,
  title    = "Family of boundary overlap metrics for the evaluation of medical
              image segmentation",
  author   = "Yeghiazaryan, V. and Voiculescu, I.",
  journal  = "Journal of Medical Imaging",
  volume   =  5,
  number   =  1,
  pages    = "015006",
  month    =  feb,
  year     =  2018,
  address  = "United States",
  language = "en"
}

@misc{ESA.2019_dataset,
 author={{ESA Greenland Ice Sheet CCI project team}},
 title={{ESA Greenland Ice Sheet Climate Change Initiative (Greenland\_Ice\_Sheet\_cci): Greenland Calving Front Locations, v3.0}},
 year={2019},
 url={https://catalogue.ceda.ac.uk/uuid/8889dfe3de45406e815bce13ae8a0c92},
 note={{Centre for Environmental Data Analysis}, [Dataset]},
}

@misc{Zhang.2020_dataset_network,
 author={{Zhang}, E. and {Liu}, L. and {Huang}, L. and {Ng}, K. S.},
 title={{Network delineated calving fronts at Jakobshavn Isbr{\ae}, Kangerlussuaq, and Helheim}},
 year={2020},
 doi={10.1594/PANGAEA.923272},
 note={{PANGAEA}, [Dataset]},
}

@misc{Zhang.2019_dataset_network,
 author={{Zhang}, E.},
 title={{The calving fronts delineated by the network in Jakobshavn Isbr{\ae}}},
 year={2019},
 doi={10.1594/PANGAEA.897064},
 note={{PANGAEA}, [Dataset]},
}

@misc{Zhang.2019_dataset,
 author={{Zhang}, E.},
 title={{The ground truth of the calving fronts in Jakobshavn Isbr{\ae}}},
 year={2019},
 doi={10.1594/PANGAEA.897065},
 note={{PANGAEA}, [Dataset]},
}

@misc{Schild.2013_dataset,
 author={Schild, K. M. and Hamilton, G.},
 title={Terminus position time series: Helheim and Kangerdlugssuaq glaciers, Greenland},
 year={2013},
 doi={10.18739/A2W93G},
 note={{Arctic Data Center}, [Dataset]}
}

@misc{Fausto.2019_dataset,
 author={Fausto et al., R. S.},
 title={Programme for monitoring of the Greenland ice sheet (PROMICE): Calving front line, 1999-2018},
 year={2019},
 doi={10.22008/promice/data/calving_front_lines},
 note={{Arctic Data Center}, [Dataset]}
}

@misc{king.2020_dataset,
 author={King, M. and Howat, I.},
 title={Data from: Dynamic ice loss from the Greenland Ice Sheet driven by sustained glacier retreat},
 year={2020},
 doi={10.5061/dryad.qrfj6q5cb},
 note = {{Dryad}, [Dataset]},
}

@misc{lippl.2019_dataset,
 author={{Lippl}, S.},
 title={{Glacier Surface Velocities and Outlet Areas from 2014-2018 on James Ross Island, Northern Antarctic Peninsula}},
 year={2019},
 doi={10.1594/PANGAEA.907062},
 note={{PANGAEA}, [Dataset]},
}

@misc{bommasani.2022,
      title={On the Opportunities and Risks of Foundation Models}, 
      author={R. Bommasani et al.},
      year={2022},
      eprint={2108.07258},
      archivePrefix={arXiv},
      primaryClass={cs.LG},
        note = {Preprint at https://arxiv.org/abs/2108.07258},
}

@misc{Cheng.2020_dataset,
 author={Cheng, Daniel and Hayes, Wayne and Larour, Eric},
 title={CALFIN: Calving front dataset for East/West Greenland, 1972--2019},
 year={2020},
 doi={10.7280/D1FH5D},
 howpublished = {{Dryad}, [Dataset]},
}

@misc{Zhang.2020_dataset,
 author={Enze {Zhang} and Lin {Liu} and Lingcao {Huang} and Ka Shing {Ng}},
 title={{Manually delineated calving fronts at Jakobshavn Isbr{\ae}, Kangerlussuaq, and Helheim}},
 year={2020},
 doi={10.1594/PANGAEA.923270},
 howpublished={{PANGAEA}, [Dataset]},
}

@misc{add.2021_dataset,
 author={Gerrish, L. and Fretwell, P. and Cooper, P.},
 title={{High resolution vector polylines of the Antarctic coastline (7.4)}},
 doi={10.5285/e46be5bc-ef8e-4fd5-967b-92863fbe2835},
 year={2021},
 howpublished={{UK Polar Data Centre, Natural Environment Research Council, UK Research \& Innovation}, [Dataset]},
}

@misc{glims.2018_dataset,
 author={Raup, B.H. and Racoviteanu, A. and Khalsa, S.J.S. and Helm, C. and Armstrong, R. and Arnaud, Y.},
 title={{GLIMS and NSIDC (2005, updated 2018): Global Land Ice Measurements from Space glacier database. Compiled and made available by the international GLIMS community and the National Snow and Ice Data Center, Boulder CO, U.S.A.}},
 doi={10.7265/N5V98602},
 year={2018},
 note={{GLIMS}, [Dataset]},
}

@misc{caffe,
 author={Gourmelon, Nora and Seehaus, Thorsten and Braun, Matthias Holger and Maier, Andreas and Christlein, Vincent},
 title={CaFFe (CAlving Fronts and where to Find thEm: a benchmark dataset and methodology for automatic glacier calving front extraction from sar imagery)},
 doi={10.1594/PANGAEA.940950},
 year={2022},
 howpublished={PANGAEA, [Dataset]},
}

@article{Baumhoer.2023,
 author = {Baumhoer, Celia A. and Dietz, Andreas J. and Heidler, Konrad and Kuenzer, Claudia},
 year = {2023},
 title = {IceLines -- A new data set of Antarctic ice shelf front positions},
 pages = {138},
 volume = {10},
 number = {1},
 journal = {Scientific Data}
}

@article{Li.2024,
 author = {Li, T. and Heidler, K. and Mou, L. and Ign{\'e}czi, {\'A}. and Zhu, X. X. and Bamber, J. L.},
 year = {2024},
 title = {A high-resolution calving front data product for marine-terminating glaciers in Svalbard},
 pages = {919--939},
 volume = {16},
 number = {2},
 journal = {Earth System Science Data}
}

@article{Loebel.2023,
 author = {Loebel et al., E.},
 year = {2023},
 title = {Calving front monitoring at sub-seasonal resolution: a deep learning application to Greenland glaciers},
 pages = {1--21},
 volume = {2023},
 journal = {The Cryosphere Discussions}
}

@article{Loebel.2025,
 author = {Loebel, E. and Baumhoer, C. A. and Dietz, A. and Scheinert, M. and Horwath, M.},
 year = {2025},
 title = {Calving front positions for 42 key glaciers of the Antarctic Peninsula Ice Sheet: a sub-seasonal record from 2013 to 2023 based on deep-learning application to Landsat multi-spectral imagery},
 pages = {65--78},
 volume = {17},
 number = {1},
 journal = {Earth System Science Data}
}

@misc{enveo,
 author={ENVEO},
 title={Greenland Calving Front Dataset v3.0},
 url={http://products.esa-icesheets-cci.org/products/
downloadlist/CFL},
 year={2017},
 note={[Dataset], (last access: 6 February 2025)}
}

@article{Ma.2024,
author={Ma, Jun
and He, Yuting
and Li, Feifei
and Han, Lin
and You, Chenyu
and Wang, Bo},
title={Segment anything in medical images},
journal={Nature Communications},
year={2024},
month={Jan},
day={22},
volume={15},
number={1},
pages={654},
issn={2041-1723},
doi={10.1038/s41467-024-44824-z},
url={https://doi.org/10.1038/s41467-024-44824-z}
}

@article{Wagner.2022,
author = {Wagner et al., Fabian},
title = {Ultralow-parameter denoising: Trainable bilateral filter layers in computed tomography},
journal = {Medical Physics},
volume = {49},
number = {8},
pages = {5107-5120},
doi = {https://doi.org/10.1002/mp.15718},
url = {https://aapm.onlinelibrary.wiley.com/doi/abs/10.1002/mp.15718},
eprint = {https://aapm.onlinelibrary.wiley.com/doi/pdf/10.1002/mp.15718},
year = {2022}
}

@article{Baumhoer.2018,
 author = {Baumhoer, Celia and Dietz, Andreas and Dech, Stefan and Kuenzer, Claudia},
 year = {2018},
 title = {Remote Sensing of Antarctic Glacier and Ice-Shelf Front Dynamics - A Review},
 pages = {1445:1--1445:28},
 volume = {10},
 number = {9},
 journal = {Remote Sensing}
}

@article{Karimi.2020,
 author = {Karimi, Davood and Salcudean, Septimiu E.},
 year = {2020},
 title = {Reducing the Hausdorff Distance in Medical Image Segmentation With Convolutional Neural Networks},
 pages = {499--513},
 volume = {39},
 number = {2},
 journal = {IEEE Transactions on Medical Imaging}
}

@article{Sun.2022,
 author = {Sun, Gang and Yu, Hancheng and Jiang, Xiangtao and Feng, Mingkui},
 year = {2022},
 title = {Adaptive Feature Pyramid Network to Predict Crisp Boundaries via NMS Layer and ODS F-Measure Loss Function},
 volume = {13},
 number = {1},
 journal = {Information}
}

@InProceedings{Heidari_2023_WACV,
    author    = {Heidari et al., Moein},
    title     = {HiFormer: Hierarchical Multi-Scale Representations Using Transformers for Medical Image Segmentation},
    booktitle = {Proceedings of the IEEE/CVF Winter Conference on Applications of Computer Vision (WACV)},
    month     = {January},
    year      = {2023},
    pages     = {6202-6212}
}

@article{Liu.2004,
author = {H. Liu and K. C. Jezek},
title = {Automated extraction of coastline from satellite imagery by integrating Canny edge detection and locally adaptive thresholding methods},
journal = {International Journal of Remote Sensing},
volume = {25},
number = {5},
pages = {937--958},
year = {2004},
publisher = {Taylor \& Francis},
doi = {10.1080/0143116031000139890},
URL = {https://doi.org/10.1080/0143116031000139890},
eprint = {https://doi.org/10.1080/0143116031000139890}
}

@ARTICLE{Mason.1996,

  author={Mason, D.C. and Davenport, I.J.},

  journal={IEEE Transactions on Geoscience and Remote Sensing}, 

  title={Accurate and efficient determination of the shoreline in ERS-1 SAR images}, 

  year={1996},

  volume={34},

  number={5},

  pages={1243-1253},

  doi={10.1109/36.536540}}

@article{Modava.2017,
author = {Mohammad Modava and Gholamreza Akbarizadeh},
title = {Coastline extraction from SAR images using spatial fuzzy clustering and the active contour method},
journal = {International Journal of Remote Sensing},
volume = {38},
number = {2},
pages = {355--370},
year = {2017},
publisher = {Taylor \& Francis},
doi = {10.1080/01431161.2016.1266104},
URL = {https://doi.org/10.1080/01431161.2016.1266104},
eprint = {https://doi.org/10.1080/01431161.2016.1266104}
}

@ARTICLE{Tello.2011,

  author={Tello Alonso, Mariví and Lopez-Martinez, Carlos and Mallorqui, Jordi J. and Salembier, Philippe},

  journal={IEEE Transactions on Geoscience and Remote Sensing}, 

  title={Edge Enhancement Algorithm Based on the Wavelet Transform for Automatic Edge Detection in SAR Images}, 

  year={2011},

  volume={49},

  number={1},

  pages={222-235},

  doi={10.1109/TGRS.2010.2052814}}

@INPROCEEDINGS{Javed.2013,

  author={Javed, Umer and Riaz, Muhammad Mohsin and Ghafoor, Abdul and Cheema, Tanveer Ahmed},

  booktitle={2013 International Symposium on Intelligent Signal Processing and Communication Systems}, 

  title={Fuzzy active contours based SAR image segmentation}, 

  year={2013},

  volume={},

  number={},

  pages={17-21},

  doi={10.1109/ISPACS.2013.6704515}}

@article{Sohn.1999,
author = {Hong-Gyoo Sohn and K. C. Jezek},
title = {Mapping ice sheet margins from ERS-1 SAR and SPOT imagery},
journal = {International Journal of Remote Sensing},
volume = {20},
number = {15-16},
pages = {3201--3216},
year = {1999},
publisher = {Taylor \& Francis},
doi = {10.1080/014311699211705},
URL = {https://doi.org/10.1080/014311699211705},
eprint = {https://doi.org/10.1080/014311699211705}
}

@inproceedings{Krieger.2017,
            year = {2017},
           title = {Automatic calving front delienation on TerraSAR-X and Sentinel-1 SAR imagery},
       booktitle = {2017 IEEE International Geoscience and Remote Sensing Symposium (IGARSS)},
          author = {Krieger, Lukas and Floricioiu, Dana},
             url = {https://elib.dlr.de/114563/}
}

@TECHREPORT{IPCC.2023,
  title     = "Climate Change 2023: Synthesis Report. Contribution of Working Groups {I}, {II} and {III} to the Sixth Assessment Report of the Intergovernmental Panel on Climate Change [core writing team, H. lee and J. romero (eds.)].",
  author    = "{IPCC, 2023}",
  editor    = "Arias et al., Paola",
  publisher = "Intergovernmental Panel on Climate Change (IPCC)",
  month     =  jul,
  year      =  2023,
  institution = {IPCC},
  address = {Geneva, Switzerland},
  doi = {10.59327/IPCC/AR6-9789291691647}
}

@misc{Xiao.2024,
      title={Foundation Models for Remote Sensing and Earth Observation: A Survey}, 
      author={Aoran Xiao and Weihao Xuan and Junjue Wang and Jiaxing Huang and Dacheng Tao and Shijian Lu and Naoto Yokoya},
      year={2024},
      eprint={2410.16602},
      archivePrefix={arXiv},
      primaryClass={cs.CV},
      url={https://arxiv.org/abs/2410.16602}, 
}

\end{document}